\documentclass{article} %
\usepackage{iclr2027_conference,times}

\usepackage[utf8]{inputenc}
\usepackage[T1]{fontenc}
\usepackage{graphicx}
\usepackage[table,dvipsnames]{xcolor}
\usepackage{amsmath}
\usepackage{amssymb}
\usepackage{amsfonts}
\usepackage{nicefrac}
\usepackage{microtype}
\usepackage{xspace}
\usepackage{booktabs}
\usepackage{multirow}
\usepackage{longtable}
\usepackage{array}
\usepackage{wrapfig}
\usepackage{titletoc}
\usepackage[labelfont=bf,font=small]{caption}
\usepackage{subcaption}
\usepackage{algorithm}
\usepackage{algpseudocode}
\usepackage{amsthm}
\newtheorem{lemma}{Lemma}
\usepackage{pgf}
\usepackage{pifont}
\usepackage{pgfplots}
\pgfplotsset{compat=1.17}
\usepackage{pgfplotstable}
\usepackage{tikz}
\usepackage{csvsimple}
\usepackage{xifthen}
\usepackage{arydshln}
\usepackage{adjustbox}
\usepackage{balance}
\usepackage{tcolorbox}
\usepackage{etoolbox}
\usepackage{pifont}
\usepgfplotslibrary{fillbetween}
\usetikzlibrary{patterns}
\usetikzlibrary{shapes,calc,positioning, decorations.text, arrows.meta, shadows.blur, shadings, fit}

\makeatletter
\DeclareRobustCommand\onedot{\futurelet\@let@token\@onedot}
\def\@onedot{\ifx\@let@token.\else.\null\fi\xspace}
\def\eg{\emph{e.g}\onedot}
\def\Eg{\emph{E.g}\onedot}
\def\ie{\emph{i.e}\onedot}
\def\Ie{\emph{I.e}\onedot}
\def\cf{\emph{cf}\onedot}
\def\Cf{\emph{Cf}\onedot}
\def\etc{\emph{etc}\onedot}
\def\vs{\emph{vs}\onedot}
\def\wrt{w.r.t\onedot}
\def\dof{d.o.f\onedot}

\def\etal{\emph{et al}\onedot}
\makeatother

\definecolor{citeblue}{rgb}{0.12,0.49,0.85}
\definecolor{gtgreen}{HTML}{1E8449}
\definecolor{clipblue}{HTML}{2874A6}
\definecolor{sgapgreen}{HTML}{239B56}
\definecolor{rowclip}{HTML}{EBF5FB}
\definecolor{rowsgap}{HTML}{EAFAF1}
\definecolor{querybg}{HTML}{F8F9FA}
\definecolor{siglipviolet}{HTML}{6C3483}
\definecolor{rowsiglip}{HTML}{F5EEF8}

\usepackage[pagebackref,breaklinks,colorlinks,citecolor=citeblue]{hyperref}
\usepackage{url}
\usepackage[capitalize]{cleveref}
\crefname{section}{Sec.}{Secs.}
\Crefname{section}{Section}{Sections}
\crefname{table}{Tab.}{Tabs.}
\Crefname{table}{Table}{Tables}

\newlength{\imw}
\title{Global Average Precision\\for Representation Learning}

\author{
    \textbf{Bill Psomas}$^{1}$\thanks{Equal technical contribution.} \quad 
    \textbf{Mohammad Mahdi}$^2$  \quad 
    \textbf{Michalis Thomas}  \\
    \textbf{Danda Pani Paudel}$^2$  \quad 
    \textbf{Giorgos Tolias}$^1$ \quad
    \textbf{Giorgos Kordopatis-Zilos}$^{1\hspace{0.5pt}*}$ \\[2pt]
    {\small $^1$VRG, FEE, Czech Technical University in Prague \quad $^2$INSAIT, Sofia University ``St. Kliment Ohridski''}
}

\iclrfinalcopy
\makeatletter
\def\@maketitle{\vbox{\hsize\textwidth
\lhead{}
{\centering\LARGE\sc \@title\par}
\hbox to \hsize{\hfil\begin{tabular}[t]{c}\bf\rule{\z@}{24pt}\@author\end{tabular}\hfil}%
\vskip 0.3in minus 0.1in}}
\makeatother

\begin{document}

\maketitle

\def\Lqtrain{\ensuremath{L_q^{\text{\tiny \raisebox{2pt}{train}}}}\xspace}
\def\Lxtrain{\ensuremath{L_x^{\text{\tiny \raisebox{2pt}{train}}}}\xspace}
\def\Lqtest{\ensuremath{L_q^{\text{\tiny \raisebox{2pt}{test}}}}\xspace}
\def\Lxtest{\ensuremath{L_x^{\text{\tiny \raisebox{2pt}{test}}}}\xspace}
\def\Lxasmk{\ensuremath{L^{\text{\tiny \raisebox{2pt}{asmk}}}_x}\xspace}

\newcommand{\ione}{i\hspace{-.05em}+\hspace{-.07em}1}

\newcommand{\mypartight}[1]{\noindent {\bf #1}}
\newcommand{\myparagraph}[1]{\vspace{3pt}\noindent\textbf{#1}\xspace}

\newcommand{\optional}[1]{{#1}}
\newcommand{\alert}[1]{{\color{red}{#1}}}
\newcommand{\gt}[1]{{\color{purple}{GT: #1}}}
\newcommand{\gtt}[1]{{\color{purple}{#1}}}
\newcommand{\gtr}[2]{{\color{purple}\st{#1} {#2}}}

\newcommand{\gkz}[1]{{\color{cyan}{GKZ: #1}}}
\newcommand{\gkzt}[1]{{\color{cyan}{#1}}}
\newcommand{\gkzr}[2]{{\color{cyan}\st{#1} {#2}}}

\newcommand{\bp}[1]{{\color{blue}{BP: #1}}}

\newcommand{\am}[1]{{\color{orange}{AM: #1}}}
\newcommand{\amt}[1]{{\color{orange}{#1}}}
\newcommand{\amr}[2]{{\color{orange}\st{#1} {#2}}}

\newcommand{\gray}[1]{{\color{gray}{#1}}}

\definecolor{appleblue}{RGB}{0,122,255}

\newcolumntype{Y}{>{\centering\arraybackslash}p{4em}}

\def\roxf{$\mathcal{R}$Oxford\xspace}
\def\rox{$\mathcal{R}$Oxf\xspace}
\def\ro{$\mathcal{R}$O\xspace}
\def\rpar{$\mathcal{R}$Paris\xspace}
\def\rpa{$\mathcal{R}$Par\xspace}
\def\rp{$\mathcal{R}$P\xspace}
\def\rdis{$\mathcal{R}$1M\xspace}

\newcommand\resnet[3]{\ensuremath{\prescript{#2}{}{\mathtt{R}}{#1}_{\scriptscriptstyle #3}}\xspace}

\newcommand{\uAP}{\text{gAP}\xspace} %
\newcommand{\ours}{\text{gSAP}\xspace} %

\newcommand{\stddev}[1]{\textsubscript{$\pm$#1}}

\newcommand{\diffup}[1]{{\color{OliveGreen}{($\uparrow$ #1)}}}
\newcommand{\diffdown}[1]{{\color{BrickRed}{($\downarrow$ #1)}}}

\newcolumntype{C}{>{\raggedleft\arraybackslash}p{18pt}}

\def\nmsp{\hspace{-6pt}}
\def\nssp{\hspace{-3pt}}
\def\nxssp{\hspace{-1pt}}
\def\zsp{\hspace{0pt}}
\def\xssp{\hspace{1pt}}
\def\ssp{\hspace{3pt}}
\def\msp{\hspace{6pt}}
\def\mlsp{\hspace{9pt}}
\def\lsp{\hspace{12pt}}
\def\xlsp{\hspace{20pt}}

\newcommand{\head}[1]{{\smallskip\noindent\bf #1}}
\newcommand{\equ}[1]{(\ref{equ:#1})\xspace}

\newcommand{\nn}[1]{\ensuremath{\text{NN}_{#1}}\xspace}
\def\l1{\ensuremath{\ell_1}\xspace}
\def\l2{\ensuremath{\ell_2}\xspace}

\newcommand{\tran}{^\top}
\newcommand{\mtran}{^{-\top}}
\newcommand{\zcol}{\mathbf{0}}
\newcommand{\zrow}{\zcol\tran}

\newcommand{\ind}{\mathds{1}}
\newcommand{\expect}{\mathbb{E}}
\newcommand{\nat}{\mathbb{N}}
\newcommand{\zahl}{\mathbb{Z}}
\newcommand{\real}{\mathbb{R}}
\newcommand{\proj}{\mathbb{P}}
\newcommand{\prob}{\mathbf{Pr}}

\newcommand{\mif}{\textrm{if }}
\newcommand{\other}{\textrm{otherwise}}
\newcommand{\minimize}{\textrm{minimize }}
\newcommand{\maximize}{\textrm{maximize }}

\newcommand{\id}{\operatorname{id}}
\newcommand{\const}{\operatorname{const}}
\newcommand{\sgn}{\operatorname{sgn}}
\newcommand{\erf}{\operatorname{erf}}
\newcommand{\var}{\operatorname{Var}}
\newcommand{\mean}{\operatorname{mean}}
\newcommand{\trace}{\operatorname{tr}}
\newcommand{\diag}{\operatorname{diag}}
\newcommand{\vect}{\operatorname{vec}}
\newcommand{\cov}{\operatorname{cov}}

\newcommand{\softmax}{\operatorname{softmax}}
\newcommand{\clip}{\operatorname{clip}}

\newcommand{\defn}{\mathrel{:=}}
\newcommand{\peq}{\mathrel{+\!=}}
\newcommand{\meq}{\mathrel{-\!=}}

\newcommand{\floor}[1]{\left\lfloor{#1}\right\rfloor}
\newcommand{\ceil}[1]{\left\lceil{#1}\right\rceil}
\newcommand{\inner}[1]{\left\langle{#1}\right\rangle}
\newcommand{\norm}[1]{\left\|{#1}\right\|}
\newcommand{\frob}[1]{\norm{#1}_F}
\newcommand{\card}[1]{\left|{#1}\right|\xspace}
\newcommand{\diff}{\mathrm{d}}
\newcommand{\der}[3][]{\frac{d^{#1}#2}{d#3^{#1}}}
\newcommand{\pder}[3][]{\frac{\partial^{#1}{#2}}{\partial{#3^{#1}}}}
\newcommand{\ipder}[3][]{\partial^{#1}{#2}/\partial{#3^{#1}}}
\newcommand{\dder}[3]{\frac{\partial^2{#1}}{\partial{#2}\partial{#3}}}

\newcommand{\wb}[1]{\overline{#1}}
\newcommand{\wt}[1]{\widetilde{#1}}

\newcommand{\cA}{\mathcal{A}}
\newcommand{\cB}{\mathcal{B}}
\newcommand{\cC}{\mathcal{C}}
\newcommand{\cD}{\mathcal{D}}
\newcommand{\cE}{\mathcal{E}}
\newcommand{\cF}{\mathcal{F}}
\newcommand{\cG}{\mathcal{G}}
\newcommand{\cH}{\mathcal{H}}
\newcommand{\cI}{\mathcal{I}}
\newcommand{\cJ}{\mathcal{J}}
\newcommand{\cK}{\mathcal{K}}
\newcommand{\cL}{\mathcal{L}}
\newcommand{\cM}{\mathcal{M}}
\newcommand{\cN}{\mathcal{N}}
\newcommand{\cO}{\mathcal{O}}
\newcommand{\cP}{\mathcal{P}}
\newcommand{\cQ}{\mathcal{Q}}
\newcommand{\cR}{\mathcal{R}}
\newcommand{\cS}{\mathcal{S}}
\newcommand{\cT}{\mathcal{T}}
\newcommand{\cU}{\mathcal{U}}
\newcommand{\cV}{\mathcal{V}}
\newcommand{\cW}{\mathcal{W}}
\newcommand{\cX}{\mathcal{X}}
\newcommand{\cY}{\mathcal{Y}}
\newcommand{\cZ}{\mathcal{Z}}

\newcommand{\vA}{\mathbf{A}}
\newcommand{\vB}{\mathbf{B}}
\newcommand{\vC}{\mathbf{C}}
\newcommand{\vD}{\mathbf{D}}
\newcommand{\vE}{\mathbf{E}}
\newcommand{\vF}{\mathbf{F}}
\newcommand{\vG}{\mathbf{G}}
\newcommand{\vH}{\mathbf{H}}
\newcommand{\vI}{\mathbf{I}}
\newcommand{\vJ}{\mathbf{J}}
\newcommand{\vK}{\mathbf{K}}
\newcommand{\vL}{\mathbf{L}}
\newcommand{\vM}{\mathbf{M}}
\newcommand{\vN}{\mathbf{N}}
\newcommand{\vO}{\mathbf{O}}
\newcommand{\vP}{\mathbf{P}}
\newcommand{\vQ}{\mathbf{Q}}
\newcommand{\vR}{\mathbf{R}}
\newcommand{\vS}{\mathbf{S}}
\newcommand{\vT}{\mathbf{T}}
\newcommand{\vU}{\mathbf{U}}
\newcommand{\vV}{\mathbf{V}}
\newcommand{\vW}{\mathbf{W}}
\newcommand{\vX}{\mathbf{X}}
\newcommand{\vY}{\mathbf{Y}}
\newcommand{\vZ}{\mathbf{Z}}

\newcommand{\va}{\mathbf{a}}
\newcommand{\vb}{\mathbf{b}}
\newcommand{\vc}{\mathbf{c}}
\newcommand{\vd}{\mathbf{d}}
\newcommand{\ve}{\mathbf{e}}
\newcommand{\vf}{\mathbf{f}}
\newcommand{\vg}{\mathbf{g}}
\newcommand{\vh}{\mathbf{h}}
\newcommand{\vi}{\mathbf{i}}
\newcommand{\vj}{\mathbf{j}}
\newcommand{\vk}{\mathbf{k}}
\newcommand{\vl}{\mathbf{l}}
\newcommand{\vm}{\mathbf{m}}
\newcommand{\vn}{\mathbf{n}}
\newcommand{\vo}{\mathbf{o}}
\newcommand{\vp}{\mathbf{p}}
\newcommand{\vq}{\mathbf{q}}
\newcommand{\vr}{\mathbf{r}}
\newcommand{\Vs}{\mathbf{s}}
\newcommand{\vt}{\mathbf{t}}
\newcommand{\vu}{\mathbf{u}}
\newcommand{\vv}{\mathbf{v}}
\newcommand{\vw}{\mathbf{w}}
\newcommand{\vx}{\mathbf{x}}
\newcommand{\vy}{\mathbf{y}}
\newcommand{\vz}{\mathbf{z}}

\newcommand{\vone}{\mathbf{1}}
\newcommand{\vzero}{\mathbf{0}}

\newcommand{\valpha}{{\boldsymbol{\alpha}}}
\newcommand{\vbeta}{{\boldsymbol{\beta}}}
\newcommand{\vgamma}{{\boldsymbol{\gamma}}}
\newcommand{\vdelta}{{\boldsymbol{\delta}}}
\newcommand{\vepsilon}{{\boldsymbol{\epsilon}}}
\newcommand{\vzeta}{{\boldsymbol{\zeta}}}
\newcommand{\veta}{{\boldsymbol{\eta}}}
\newcommand{\vtheta}{{\boldsymbol{\theta}}}
\newcommand{\viota}{{\boldsymbol{\iota}}}
\newcommand{\vkappa}{{\boldsymbol{\kappa}}}
\newcommand{\vlambda}{{\boldsymbol{\lambda}}}
\newcommand{\vmu}{{\boldsymbol{\mu}}}
\newcommand{\vnu}{{\boldsymbol{\nu}}}
\newcommand{\vxi}{{\boldsymbol{\xi}}}
\newcommand{\vomikron}{{\boldsymbol{\omikron}}}
\newcommand{\vpi}{{\boldsymbol{\pi}}}
\newcommand{\vrho}{{\boldsymbol{\rho}}}
\newcommand{\vsigma}{{\boldsymbol{\sigma}}}
\newcommand{\vtau}{{\boldsymbol{\tau}}}
\newcommand{\vupsilon}{{\boldsymbol{\upsilon}}}
\newcommand{\vphi}{{\boldsymbol{\phi}}}
\newcommand{\vchi}{{\boldsymbol{\chi}}}
\newcommand{\vpsi}{{\boldsymbol{\psi}}}
\newcommand{\vomega}{{\boldsymbol{\omega}}}

\newcommand{\rLambda}{\mathrm{\Lambda}}
\newcommand{\rSigma}{\mathrm{\Sigma}}

\makeatletter
\DeclareRobustCommand\onedot{\futurelet\@let@token\@onedot}
\def\@onedot{\ifx\@let@token.\else.\null\fi\xspace}
\def\eg{\emph{e.g}\onedot} \def\Eg{\emph{E.g}\onedot}
\def\ie{\emph{i.e}\onedot} \def\Ie{\emph{I.e}\onedot}
\def\vs{\emph{vs\onedot}}
\def\cf{\emph{cf}\onedot} \def\Cf{\emph{C.f}\onedot}
\def\etc{\emph{etc}\onedot} \def\vs{\emph{vs}\onedot}
\def\wrt{w.r.t\onedot} \def\dof{d.o.f\onedot}
\def\etal{\emph{et al}\onedot}
\makeatother

\newcommand\rurl[1]{%
  \href{https://#1}{\nolinkurl{#1}}%
}

\newcommand{\bentarrow}[1][]{%
  \begin{tikzpicture}[#1]%
    \draw (0,0.7ex) -- (0,0) -- (0.75em,0);
    \draw (0.55em,0.2em) -- (0.75em,0) -- (0.55em,-0.2em);
  \end{tikzpicture}%
}

\definecolor{higha}{HTML}{009b10} %
\definecolor{lowa}{HTML}{ec462e}  %
\definecolor{mediuma}{HTML}{FFFFFF} %

\newcommand*{\opacitya}{50} %
\newcommand*{\minvalcolora}{2.5} %
\newcommand*{\midvalcolora}{11.6} %
\newcommand*{\maxvalcolora}{37.3} %
\newcommand{\grca}[1]{
    \ifdim #1pt < \midvalcolora pt
        \pgfmathparse{(#1-\minvalcolora)/(\midvalcolora-\minvalcolora)}
        \let\normalizedval\pgfmathresult
    
        \pgfmathparse{100*(\normalizedval)^(2.0)} 
        \xdef\tempa{\pgfmathresult}
        \pgfmathparse{min(100,max(0,\tempa))}
        \xdef\tempa{\pgfmathresult}
    
        \cellcolor{mediuma!\tempa!lowa!\opacitya} #1
    \else
        \pgfmathparse{(#1-\midvalcolora)/(\maxvalcolora-\midvalcolora)}
        \let\normalizedval\pgfmathresult
    
        \pgfmathparse{100*(\normalizedval)^(2.0)}
        \xdef\tempa{\pgfmathresult}
        \pgfmathparse{min(100,max(0,\tempa))}
        \xdef\tempa{\pgfmathresult}
    
        \cellcolor{higha!\tempa!mediuma!\opacitya} #1
    \fi
}

\definecolor{highb}{HTML}{009b10} %
\definecolor{lowb}{HTML}{ec462e}  %
\definecolor{mediumb}{HTML}{FFFFFF} %

\newcommand*{\opacityb}{50} %
\newcommand*{\minvalcolorb}{1.8} %
\newcommand*{\midvalcolorb}{8.6} %
\newcommand*{\maxvalcolorb}{31.3} %
\newcommand{\grcb}[1]{
    \ifdim #1pt < \midvalcolorb pt
        \pgfmathparse{(#1-\minvalcolorb)/(\midvalcolorb-\minvalcolorb)}
        \let\normalizedval\pgfmathresult
    
        \pgfmathparse{100*(\normalizedval)^(2.0)} 
        \xdef\tempa{\pgfmathresult}
        \pgfmathparse{min(100,max(0,\tempa))}
        \xdef\tempa{\pgfmathresult}
    
        \cellcolor{mediumb!\tempa!lowb!\opacityb} #1
    \else
        \pgfmathparse{(#1-\midvalcolorb)/(\maxvalcolorb-\midvalcolorb)}
        \let\normalizedval\pgfmathresult
    
        \pgfmathparse{100*(\normalizedval)^(2.0)}
        \xdef\tempa{\pgfmathresult}
        \pgfmathparse{min(100,max(0,\tempa))}
        \xdef\tempa{\pgfmathresult}
    
        \cellcolor{highb!\tempa!mediumb!\opacityb} #1
    \fi
}

\definecolor{highc}{HTML}{009b10} %
\definecolor{lowc}{HTML}{ec462e}  %
\definecolor{mediumc}{HTML}{FFFFFF} %

\newcommand*{\opacityc}{50} %
\newcommand*{\minvalcolorc}{1.7} %
\newcommand*{\midvalcolorc}{5.9} %
\newcommand*{\maxvalcolorc}{20.8} %
\newcommand{\grcc}[1]{
    \ifdim #1pt < \midvalcolorc pt
        \pgfmathparse{(#1-\minvalcolorc)/(\midvalcolorc-\minvalcolorc)}
        \let\normalizedval\pgfmathresult
    
        \pgfmathparse{100*(\normalizedval)^(2.0)} 
        \xdef\tempa{\pgfmathresult}
        \pgfmathparse{min(100,max(0,\tempa))}
        \xdef\tempa{\pgfmathresult}
    
        \cellcolor{mediumc!\tempa!lowc!\opacityc} #1
    \else
        \pgfmathparse{(#1-\midvalcolorc)/(\maxvalcolorc-\midvalcolorc)}
        \let\normalizedval\pgfmathresult
    
        \pgfmathparse{100*(\normalizedval)^(2.0)}
        \xdef\tempa{\pgfmathresult}
        \pgfmathparse{min(100,max(0,\tempa))}
        \xdef\tempa{\pgfmathresult}
    
        \cellcolor{highc!\tempa!mediumc!\opacityc} #1
    \fi
}

\definecolor{highd}{HTML}{009b10} %
\definecolor{lowd}{HTML}{ec462e}  %
\definecolor{mediumd}{HTML}{FFFFFF} %

\newcommand*{\opacityd}{50} %
\newcommand*{\minvalcolord}{1.5} %
\newcommand*{\midvalcolord}{7.5} %
\newcommand*{\maxvalcolord}{19.8} %
\newcommand{\grcd}[1]{
    \ifdim #1pt < \midvalcolord pt
        \pgfmathparse{(#1-\minvalcolord)/(\midvalcolord-\minvalcolord)}
        \let\normalizedval\pgfmathresult
    
        \pgfmathparse{100*(\normalizedval)^(2.0)} 
        \xdef\tempa{\pgfmathresult}
        \pgfmathparse{min(100,max(0,\tempa))}
        \xdef\tempa{\pgfmathresult}
    
        \cellcolor{mediumd!\tempa!lowd!\opacityd} #1
    \else
        \pgfmathparse{(#1-\midvalcolord)/(\maxvalcolord-\midvalcolord)}
        \let\normalizedval\pgfmathresult
    
        \pgfmathparse{100*(\normalizedval)^(2.0)}
        \xdef\tempa{\pgfmathresult}
        \pgfmathparse{min(100,max(0,\tempa))}
        \xdef\tempa{\pgfmathresult}
    
        \cellcolor{highd!\tempa!mediumd!\opacityd} #1
    \fi
}

\definecolor{highe}{HTML}{009b10} %
\definecolor{lowe}{HTML}{ec462e}  %
\definecolor{mediume}{HTML}{FFFFFF} %

\newcommand*{\opacitye}{50} %
\newcommand*{\minvalcolore}{2.3} %
\newcommand*{\midvalcolore}{10.6} %
\newcommand*{\maxvalcolore}{24.7} %
\newcommand{\grce}[1]{
    \ifdim #1pt < \midvalcolore pt
        \pgfmathparse{(#1-\minvalcolore)/(\midvalcolore-\minvalcolore)}
        \let\normalizedval\pgfmathresult
    
        \pgfmathparse{100*(\normalizedval)^(2.0)} 
        \xdef\tempa{\pgfmathresult}
        \pgfmathparse{min(100,max(0,\tempa))}
        \xdef\tempa{\pgfmathresult}
    
        \cellcolor{mediume!\tempa!lowe!\opacitye} #1
    \else
        \pgfmathparse{(#1-\midvalcolore)/(\maxvalcolore-\midvalcolore)}
        \let\normalizedval\pgfmathresult
    
        \pgfmathparse{100*(\normalizedval)^(2.0)}
        \xdef\tempa{\pgfmathresult}
        \pgfmathparse{min(100,max(0,\tempa))}
        \xdef\tempa{\pgfmathresult}
    
        \cellcolor{highe!\tempa!mediume!\opacitye} #1
    \fi
}

\newcommand{\amescol}{cyan}
\newcommand{\sglipcol}{OliveGreen}
\newcommand{\dinocol}{BrickRed}
\newcommand{\globalcol}{BrickRed}

\definecolor{lightred}{RGB}{231, 76, 60}   
\definecolor{lightblue}{RGB}{54, 69, 79}
\definecolor{lightgreen}{RGB}{50, 205, 50} 

\definecolor{appleblue}{RGB}{80,122,255}  %
\definecolor{applepink}{RGB}{217,127,174}  %
\definecolor{appleteal}{RGB}{85,163,152}  %
\definecolor{applepurple}{RGB}{138,107,221}  %
\definecolor{applemustard}{RGB}{234,179,77}  %
\definecolor{appleorange}{RGB}{255,127,80}  %
\definecolor{applegreen}{RGB}{52,199,89}
\definecolor{appleyellow}{RGB}{255,204,0}
\definecolor{applered}{RGB}{255,59,48}  %

\definecolor{cvprblue}{rgb}{0.21,0.49,0.74}
\definecolor{materialPurple}{rgb}{0.615, 0.275, 1} %
\definecolor{plotlyBlue}{rgb}{0, 0.482, 1} %
\definecolor{plotlyGreen}{rgb}{0.157, 0.655, 0.271} %
\definecolor{plotlyRed}{rgb}{0.863, 0.208, 0.271} %
\definecolor{plotlyOrange}{rgb}{1, 0.757, 0.027} %
\definecolor{plotlyYellow}{rgb}{1, 0.843, 0} %
\definecolor{plotlyCyan}{rgb}{0.09, 0.745, 0.812} %
\definecolor{plotlyMagenta}{rgb}{0.917, 0.722, 0.894} %
\definecolor{plotlyTeal}{rgb}{0, 0.545, 0.455} %
\definecolor{plotlyNavy}{rgb}{0.4, 0.063, 0.949} %
\definecolor{plotlyPurple}{rgb}{0.58, 0.404, 0.741} %
\definecolor{plotlyBrown}{rgb}{0.82, 0.604, 0.416} %
\definecolor{plotlyPink}{rgb}{0.89, 0.467, 0.761} %
\definecolor{plotlyGray}{rgb}{0.498, 0.498, 0.498} %
\definecolor{plotlyLightBlue}{rgb}{0.122, 0.467, 0.706} %
\definecolor{plotlyLightOrange}{rgb}{1, 0.733, 0.471} %
\definecolor{plotlyLightGreen}{rgb}{0.596, 0.875, 0.541} %

\definecolor{calendarPurple}{rgb}{0.635, 0.349, 1.000} %
\definecolor{calendarRed}{rgb}{1.000, 0.420, 0.420} %
\definecolor{calendarPinkRed}{rgb}{1.000, 0.396, 0.518} %
\definecolor{calendarLightPink}{rgb}{1.000, 0.663, 0.663} %
\definecolor{calendarLightPurple}{rgb}{0.847, 0.627, 1.000} %
\definecolor{calendarDarkBlue}{rgb}{0.424, 0.435, 1.000} %
\definecolor{calendarBlue}{rgb}{0.000, 0.620, 1.000} %
\definecolor{calendarGreen}{rgb}{0.000, 0.714, 0.427} %
\definecolor{calendarLightMint}{rgb}{0.678, 0.910, 0.827} %
\definecolor{calendarYellow}{rgb}{1.000, 0.922, 0.231} %
\definecolor{calendarAmber}{rgb}{1.000, 0.757, 0.027} %
\definecolor{calendarOrange}{rgb}{1.000, 0.596, 0.000} %
\definecolor{calendarBrown}{rgb}{0.490, 0.353, 0.314} %
\definecolor{calendarGray}{rgb}{0.710, 0.710, 0.710} %
\definecolor{calendarGrayBlue}{rgb}{0.482, 0.604, 0.631} %
\definecolor{calendarLightBlue}{rgb}{0.647, 0.847, 0.867} %
\definecolor{calendarDarkPurple}{rgb}{0.416, 0.051, 0.678} %

\pgfplotsset{
  colormap/magma/.style={%
    /pgfplots/colormap={magma}{%
      rgb=(0.001462, 0.000466, 0.013866)
      rgb=(0.035520, 0.028397, 0.125209)
      rgb=(0.102815, 0.063010, 0.257854)
      rgb=(0.191460, 0.064818, 0.396152)
      rgb=(0.291366, 0.064553, 0.475462)
      rgb=(0.384299, 0.097855, 0.501002)
      rgb=(0.475780, 0.134577, 0.507921)
      rgb=(0.569172, 0.167454, 0.504105)
      rgb=(0.664915, 0.198075, 0.488836)
      rgb=(0.761077, 0.231214, 0.460162)
      rgb=(0.852126, 0.276106, 0.418573)
      rgb=(0.925937, 0.346844, 0.374959)
      rgb=(0.969680, 0.446936, 0.360311)
      rgb=(0.989363, 0.557873, 0.391671)
      rgb=(0.996580, 0.668256, 0.456192)
      rgb=(0.996727, 0.776795, 0.541039)
      rgb=(0.992440, 0.884330, 0.640099)
      rgb=(0.987053, 0.991438, 0.749504)
    },
  },
  colormap/inferno/.style={%
    /pgfplots/colormap={inferno}{%
      rgb=(0.001462, 0.000466, 0.013866)
      rgb=(0.037668, 0.025921, 0.132232)
      rgb=(0.116656, 0.047574, 0.272321)
      rgb=(0.217949, 0.036615, 0.383522)
      rgb=(0.316282, 0.053490, 0.425116)
      rgb=(0.410113, 0.087896, 0.433098)
      rgb=(0.503493, 0.121575, 0.423356)
      rgb=(0.596940, 0.154848, 0.398125)
      rgb=(0.688653, 0.192239, 0.357603)
      rgb=(0.775059, 0.239667, 0.303526)
      rgb=(0.851384, 0.302260, 0.239636)
      rgb=(0.912966, 0.381636, 0.169755)
      rgb=(0.956852, 0.475356, 0.094695)
      rgb=(0.981895, 0.579392, 0.026250)
      rgb=(0.987464, 0.690366, 0.079990)
      rgb=(0.973088, 0.805409, 0.216877)
      rgb=(0.947594, 0.917399, 0.410665)
      rgb=(0.988362, 0.998364, 0.644924)
    },
  },
  colormap/plasma/.style={%
    /pgfplots/colormap={plasma}{%
      rgb=(0.050383, 0.029803, 0.527975)
      rgb=(0.186213, 0.018803, 0.587228)
      rgb=(0.287076, 0.010855, 0.627295)
      rgb=(0.381047, 0.001814, 0.653068)
      rgb=(0.471457, 0.005678, 0.659897)
      rgb=(0.557243, 0.047331, 0.643443)
      rgb=(0.636008, 0.112092, 0.605205)
      rgb=(0.706178, 0.178437, 0.553657)
      rgb=(0.768090, 0.244817, 0.498465)
      rgb=(0.823132, 0.311261, 0.444806)
      rgb=(0.872303, 0.378774, 0.393355)
      rgb=(0.915471, 0.448807, 0.342890)
      rgb=(0.951344, 0.522850, 0.292275)
      rgb=(0.977856, 0.602051, 0.241387)
      rgb=(0.992541, 0.687030, 0.192170)
      rgb=(0.992505, 0.777967, 0.152855)
      rgb=(0.974443, 0.874622, 0.144061)
      rgb=(0.940015, 0.975158, 0.131326)
    },
  },
}

\begin{abstract}
Standard information retrieval metrics, such as mean Average Precision (mAP), assess performance one query at a time, based on how the similarities between a query and its positives compare against those with its negatives. The same holds for common representation learning losses, such as InfoNCE and per-query AP surrogates. None of them considers whether similarities are comparable \emph{across} queries, which any system with a single decision threshold relies on. Global Average Precision (\uAP) does, by ranking all query--candidate pairs in one list and computing a single AP. We introduce \ours, a differentiable surrogate of \uAP. It needs only a similarity matrix and a binary matrix marking the positive pairs, the same input as existing losses, so it is a drop-in replacement for them and \emph{agnostic} to the encoder, the modality, and the source of supervision. Since it considers all possible pairwise comparisons in the batch jointly, it also remains trainable at low temperatures, a regime where per-query surrogates run out of gradient.
Swapping it into established recipes improves supervised metric learning, cross-modal alignment, and self-supervised pretraining, where, to our knowledge, it is the first ranking loss to replace the community standard InfoNCE in the latter two. Its similarities are more consistent across queries, which drives the gains under a universal threshold. \ours retrieves up to four times as many positive pairs as the strongest AP surrogate at the same precision, and it degrades the least when queries with no positives in the database are added. Beyond thresholding, models trained with \ours also learn better representations, with higher transfer, $k$NN and zero-shot classification accuracy.
\end{abstract}

\section{Introduction}
\label{sec:intro}

\begin{figure}[t]
    \centering
    \captionsetup{skip=3pt}
    \vspace{-20pt}
    \hspace{7.5pt}%
    \begin{subfigure}[b]{0.6493\linewidth}
        \centering
        \includegraphics[width=\linewidth,trim=0 0 13.5bp 0,clip]{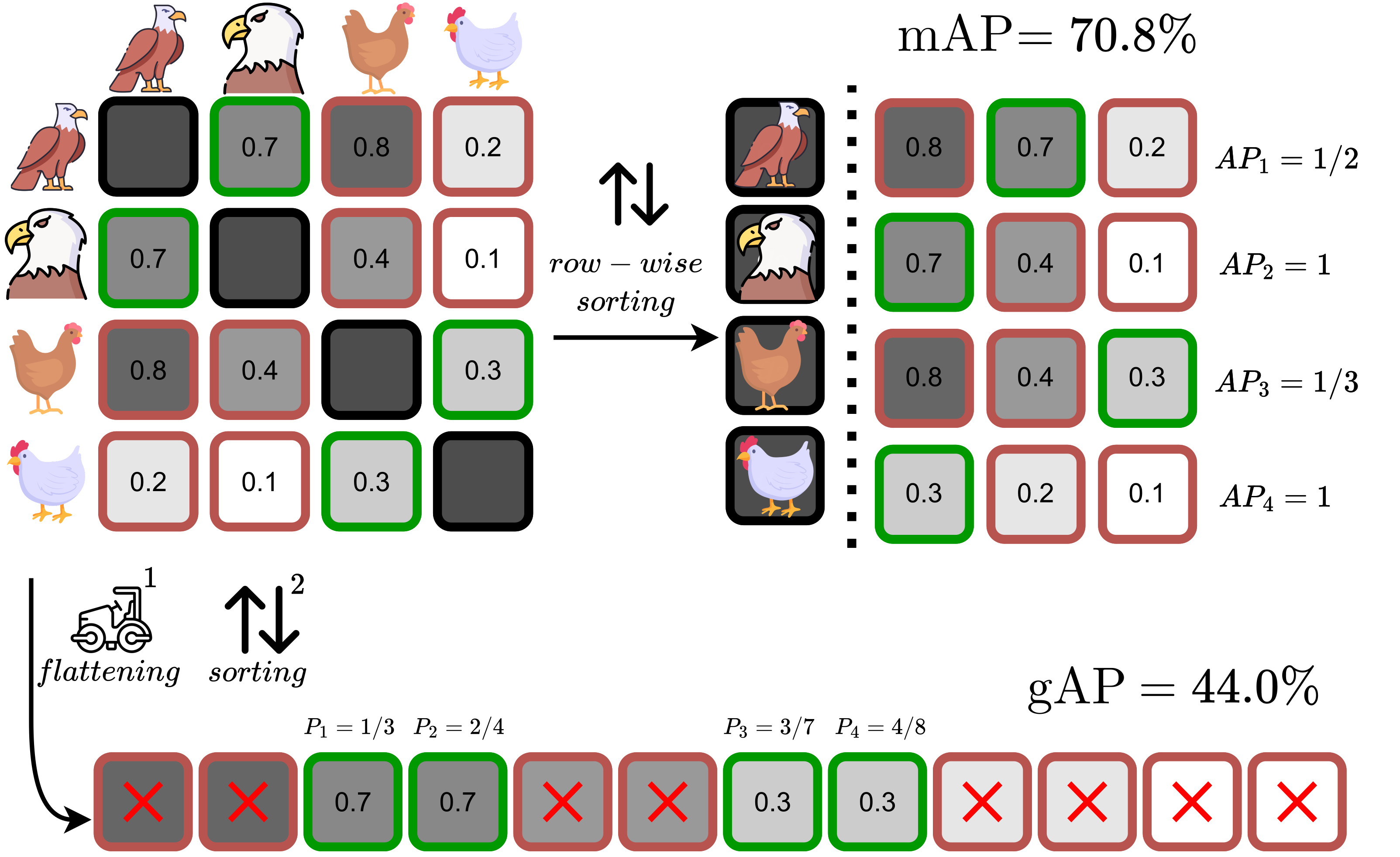}
        \vspace{-14pt}
        \caption{}
    \end{subfigure}%
    \hspace{0.0297\linewidth}%
    \begin{subfigure}[b]{0.291\linewidth}
        \centering
        \pgfplotsset{every tick label/.append style={font=\tiny}}

\input{data/sim_dist_data}

\begin{tabular}{c}
\hspace{0pt}
\begin{tikzpicture}
    \begin{axis}[
        width=1.25\linewidth,
        height=0.915\linewidth,
        ylabel style={yshift=-5pt},
        label style={font=\scriptsize},
        title style={yshift=-5pt},
        xmin=0.45, xmax=0.9,
        ymin=0, ymax=120,
        ytick=\empty,
        bar width=3pt,
        xtick={0.2, 0.4, 0.6, 0.8, 1.0}, 
        xmajorgrids=true,
        enlargelimits=0.0,
        title={mSAP},
        title style={yshift=-2pt},
    ]
        \addplot[ybar, fill=appleblue, smooth,tension=0.3,draw=none, opacity=0.8] table[x=sim, y=dist,each nth point=1] {\sappos};
        \addplot[ybar, fill=appleorange, smooth,tension=0.3,draw=none, opacity=0.8] table[x=sim, y=dist,each nth point=1] {\sapneg};

    \end{axis}
    \begin{axis}[
        width=1.25\linewidth,
        height=0.915\linewidth,
        ylabel style={yshift=-5pt},
        label style={font=\scriptsize},
        title style={yshift=-5pt},
        xmin=0.45, xmax=0.9,
        ymin=0, ymax=120,
        ytick=\empty,
        xmajorgrids=true,
        enlargelimits=0.0,
        no markers,
        smooth, tension=1.0,
        legend pos=north east,
        xticklabel=\empty,
        legend style={
            font=\scriptsize, 
            at={(.58,.96)},
            anchor=north west, 
            opacity=1.,
        }
    ]
        \addplot[appleblue!70!black,smooth,tension=0.3,thick, opacity=1, line width=1.2pt] table[x=sim, y=dist,each nth point=1] {\sappos};
        \addlegendentry{pos}
        \addplot[appleorange!70!black,smooth,tension=0.3,thick, opacity=1, line width=1.2pt] table[x=sim, y=dist,each nth point=1] {\sapneg};
        \addlegendentry{neg}

    \end{axis}
\end{tikzpicture}

\vspace{-6pt}

\\ 

\begin{tikzpicture}
    \begin{axis}[
        width=1.25\linewidth,
        height=0.915\linewidth,
        ylabel style={yshift=-5pt},
        label style={font=\scriptsize},
        title style={yshift=-5pt},
        xmin=0.45, xmax=0.9,
        ymin=0, ymax=120,
        ytick=\empty,
        bar width=3pt,
        xtick={0.2, 0.4, 0.6, 0.8, 1.0}, 
        xmajorgrids=true,
        enlargelimits=0.0,
        title={\ours},
        title style={yshift=-2pt},
    ]
        \addplot[ybar, fill=appleblue, smooth,tension=0.3,draw=none, opacity=0.8] table[x=sim, y=dist,each nth point=1] {\sgappos};
        \addplot[ybar, fill=appleorange, smooth,tension=0.3,draw=none, opacity=0.8] table[x=sim, y=dist,each nth point=1] {\sgapneg};

    \end{axis}+
    \begin{axis}[
        width=1.25\linewidth,
        height=0.915\linewidth,
        ylabel style={yshift=-5pt},
        label style={font=\scriptsize},
        title style={yshift=-5pt},
        xmin=0.45, xmax=0.9,
        ymin=0, ymax=120,
        ytick=\empty,
        xmajorgrids=true,
        enlargelimits=0.0,
        no markers,
        smooth, tension=1.0,
        xticklabel=\empty,
    ]
        \addplot[appleblue!70!black,smooth,tension=0.3,thick, opacity=1, line width=1.2pt] table[x=sim, y=dist,each nth point=1] {\sgappos};
        \addplot[appleorange!70!black,smooth,tension=0.3,thick, opacity=1, line width=1.2pt] table[x=sim, y=dist,each nth point=1] {\sgapneg};

    \end{axis}
\end{tikzpicture}

\end{tabular}
        \vspace{-20.6pt}
        \caption{}
    \end{subfigure}
    \caption{\textbf{Overview of mAP and \uAP and the effect of optimizing their surrogates.} (a) Illustration of the metric computation for a batch of images. 
For mAP, each row of the similarity matrix is treated as an independent query: AP is computed per row after ranking the similarities, and the final score is obtained by averaging across queries. For \uAP, all pairwise similarities are first aggregated into a single list, and a single AP value is computed globally. 
Consequently, mAP evaluates ranking quality per query and is insensitive to how similarities compare across queries, whereas \uAP penalizes inconsistencies in similarity distributions across queries. (b) Distribution of positive and top-negative similarities after overfitting a single batch using mSAP~\citep{brown2020smooth} and the proposed \ours loss. 
The proposed loss produces better separation between positive and negative similarities, allowing them to be distinguished by a single global threshold.
\label{fig:teaser}
\vspace{-14pt}
}
\end{figure}

Average Precision (AP)~\citep{schutze2008introduction} is the standard metric for ranking quality and is reported as mean AP (mAP) over queries across retrieval and recognition tasks~\citep{radenovic2018fine,li2014deepreid,rolinek2020optimizing}. mAP treats each query \emph{separately} (\Cref{fig:teaser}), so the distribution of similarities may vary from query to query.
Global Average Precision (\uAP)~\citep{perronnin2009family}, in contrast, ranks all query--database pairs in a single global list and computes a single AP, \ie it evaluates all queries \emph{jointly}.
It therefore captures the consistency of similarities across queries as well as ranking quality, and bridges retrieval and detection through a threshold that separates relevant from irrelevant pairs, which is why several benchmarks adopt it~\citep{weyand2020google,ypsilantis2021met,douze2021image,pizzi20242023,kordopatis2023self}. \uAP is thus the right metric for similarity models and systems operating under a single decision threshold. In \Cref{fig:teaser}(b), a per-query surrogate leaves positive and top-negative similarities overlapping across queries, whereas the proposed loss separates them.

Despite its desirable properties, \uAP is not used as a training objective, mainly because the ranking at its core is discrete and because it is computed over every pair of the similarity matrix, which is costly when the matrix is large.
In this work, we turn \uAP into a practical loss for representation learning via pruning negative similarities that do not significantly impact the loss. We derive a smooth surrogate, \ours, that directly optimizes this global metric. Per-query losses, such as the widely used InfoNCE~\citep{oord2018representation} and the existing AP surrogates~\citep{qin2010general,revaud2019learning,brown2020smooth,ramzi2021robust}, optimize the ranking of each query in isolation and therefore leave the similarity scales of different queries unconstrained. \ours ranks the whole similarity matrix at once and replaces the loss of any model whose training objective is based on a similarity matrix.

Ranking the whole matrix also eases the trade-off between the fidelity of the surrogate and the flow of its gradient. A small temperature makes the sigmoid follow the step function closely~\citep{qin2010general,brown2020smooth}, but saturates most comparisons, so their gradients vanish~\citep{ramzi2021robust}. Since \ours involves many more comparisons than per-query surrogates, as each positive is compared with every similarity in the batch rather than with its own row only, a sufficient fraction of them remains unsaturated. This allows low temperatures, and thus an accurate AP approximation, without vanishing gradients, which is not possible for the per-query counterparts.

The formulation needs only a similarity matrix and a binary relevance matrix, and is therefore \emph{agnostic} to the encoders, the modality, and the source of supervision. Relevance marks same-class pairs in supervised metric learning, matching pairs in image--text, video--text, or audio--text alignment, and augmented views of the same sample in self-supervised learning. To our knowledge, \ours is the first ranking loss to replace InfoNCE in CLIP-style cross-modal and self-supervised training.

We evaluate \ours by replacing only the loss in established recipes for supervised metric learning, CLIP and CLAP cross-modal alignment, MoCo~v3 self-supervised pretraining, and tasks with distractor queries. Across all of them, \ours improves \uAP while remaining competitive in mAP and Recall@$k$, and outperforms InfoNCE and SigLIP, the two standard cross-modal losses, on every retrieval benchmark. It also yields better representations, as the encoders trained with it reach higher transfer, $k$NN and zero-shot classification accuracy. The gains come from the consistency across queries, which translates into more matches under a single shared threshold and the smallest loss of \uAP when distractor queries are added.

In summary, our contributions are outlined below.
\vspace{-6pt}
\begin{itemize}
    \setlength\itemsep{0pt}
    \item We advocate \uAP as a metric for representation learning, since it captures both ranking quality and the consistency of similarities across queries, which per-query metrics ignore.
    \item We propose \ours, a smooth surrogate of \uAP and a drop-in replacement for existing losses that keeps gradients alive at temperatures where per-query surrogates saturate, and negative dropping, which scales it to very large batches.
    \item On supervised, cross-modal (image, text, video, audio), self-supervised, and distractor-query tasks, where \ours is the first ranking loss to replace InfoNCE, it attains the best \uAP on every retrieval benchmark, matches or exceeds the best per-query metrics in nearly all cases, and gains by far the most when one threshold serves all queries.
\end{itemize}

\section{Related work}
\label{sec:related}

Representation learning losses differ in how much of the similarity matrix they couple. Pairwise losses score each entry on its own; the contrastive loss~\citep{chopra2005learning} with an absolute margin and SigLIP~\citep{zhai2023sigmoid} with a bias shared by all pairs impose one global threshold, but at a single operating point. Most losses act per query. Triplet~\citep{schroff2015facenet} and Multi-Similarity~\citep{wang2019multi} weight pairs relative to each anchor, and proxy losses~\citep{deng2019arcface,kim2020proxy} compare each sample with class proxies. InfoNCE~\citep{oord2018representation}, standard in vision--language~\citep{radford2021learning} and self-supervised~\citep{chen2021empirical} learning, normalizes each row and column with a softmax; RankCLIP~\citep{zhang2025rankclip} adds listwise ordinal-consistency terms across and within modalities. Ranking surrogates directly optimize per-query AP~\citep{rolinek2020optimizing,revaud2019learning,brown2020smooth,ramzi2021robust} or Recall@$k$~\citep{patel2022recall}; GlobalAP~\citep{liu2023globalap}, despite its name, extends each query's list with a memory bank and still averages AP over queries. These surrogates depend only on within-query orderings and are unchanged by any per-query increasing transformation of the similarities, and each directional InfoNCE term is unchanged by a per-query offset in its own direction. None of these terms constrains how similarities compare \emph{across queries}.
Since per-query losses leave this unconstrained, comparable similarities are typically obtained with auxiliary terms~\citep{ramzi2021robust,zhang2023threshold,liu2022oneface} or post hoc through per-query score normalization~\citep{jegou2007contextual,conneau2017word,pizzi2022self}. Batch-wise ranking across groups has also been explored in object detection~\citep{chen2019towards}, where AP is optimized over all anchors and classes of a batch via gradient approximation. By contrast, our approach trains models by directly optimizing a single surrogate loss that ranks all samples of the batch by their similarities, with every sample acting as a query, and requires neither additional regularization nor post-hoc operations. An extended discussion of related work is provided in \Cref{sec:supp_related}. 

\section{Method}
\label{sec:method}

\subsection{Average Precision} 
\label{sec:ap}

Average Precision (AP) is a standard metric for evaluating the quality of ranking in information retrieval tasks~\citep{schutze2008introduction}. It summarizes the Precision--Recall (PR) curve as the area under the curve, providing a single scalar value.

Let $\Vs = [s_1,\dots,s_N]$ denote an unsorted list of scalar similarities between items, and $\vy = [y_1,\dots,y_N]$, with $y_i\in\{0, 1\}$, the corresponding ground-truth labels, where $y_i=1$ indicates a positive (\ie same class/semantics) item, and $y_i=0$ a negative item. We define the index sets of the positive items and of all items as:
\begin{equation}
\cP(\vy)=\{i \ | \ y_i=1\} \quad  \text{and} \quad \Omega = \{1,\dots,N\}.
\end{equation}
AP requires ranking the values in $\Vs$ in descending order and is defined as the average of precision values computed at the rank of each positive item as
\begin{equation}
\mathrm{AP}(\Vs, \vy) = \frac{1}{\card{\cP(\vy)}}\sum_{i\in \cP(\vy)} \mathrm{P}_i(\Vs, \vy)
\end{equation}

Computing precision $\mathrm{P}_i$ related to item $i$ requires counting the number of positives and the number of all items that are ranked before it~\citep{qin2010general}.
This counting process is performed by comparing $s_i$ with the similarities of all positives and all other items. 
Precision of  item $i$ is given by 
\begin{equation}
\mathrm{P}_i(\Vs, \vy) = \frac{\sum_{j\in \cP(\vy)} h(s_j - s_i)}{\sum_{j \in \Omega} h(s_j - s_i)}
\label{eq:precision}
\end{equation}
where $h$ is the Heaviside step function:
\begin{equation}
h(\Delta s) = 
    \begin{cases}
    1 & \Delta s \geq 0 \\
    0 & \mathrm{otherwise}
    \end{cases}
    \label{eq:heaviside}
\end{equation}
AP is optimal, equal to 1, when all positives are ranked before all negatives. 

\subsection{Evaluation metrics}
\label{sec:eval_metrics}

Consider a typical retrieval setting with $M$ queries and $N$ database items. The pairwise similarity matrix, comparing each query and database item, and the corresponding ground-truth label matrix are denoted by 
\begin{equation}
S =
    \begin{bmatrix}
    s_{1,1} & \cdots & s_{1,N} \\
    \vdots & \ddots & \vdots \\
    s_{M,1} & \cdots & s_{M,N}
    \end{bmatrix}
    \in \real^{M \times N} \quad 
Y =
    \begin{bmatrix}
    y_{1,1} & \cdots & y_{1,N} \\
    \vdots & \ddots & \vdots \\
    y_{M,1} & \cdots & y_{M,N}
    \end{bmatrix}
    \in \{0,1\}^{M \times N}.
\end{equation}
Each row of these matrices corresponds to the similarities and labels of a query, with $S_{i}$ and $Y_{i}$ denoting the elements of the $i$-th row related to the $i$-th query.

\paragraph{mean Average Precision (mAP).} mAP evaluates each query independently, in a row-wise fashion, and averages the AP across all queries:
\begin{equation}
\mathrm{mAP}(S, Y) = \frac{1}{M} \sum_{i=1}^{M} \mathrm{AP}(S_i, Y_i).
\label{eq:map}
\end{equation}
This metric evaluates the ranking quality separately per query, treating each query equally. 
There is no interaction between different queries, hence similarity distributions can vary among rows.
This metric can be optimal even when similarities are not comparable across queries, \eg the similarities of all positives of one row are smaller than all negatives of another row, despite AP being equal to 1 for both rows.
Therefore, mAP does not reflect the consistency of similarities across queries.

\paragraph{global Average Precision (\uAP).} In the case of  \uAP, the AP is computed only once, considering similarities across all queries jointly. It is defined as:
\begin{equation}
\text{\uAP}(S, Y) = \mathrm{AP}(\vect(S), \vect(Y)),
\label{eq:uap}
\end{equation}
where $\vect: \real^{M \times N} \rightarrow \real^{MN}$ is the row-wise flattening operation. This metric evaluates the global ranking quality across all queries jointly. 
Optimal \uAP requires all positives, no matter the query, to have larger similarity than all negatives, no matter the query. 
Therefore, this metric is sensitive to variations in the similarity distribution across queries.
It reflects both the ranking quality per query and the consistency of similarities across queries.

\subsection{Surrogate losses}
\label{sec:surrogates}

Standard AP relies on the discrete Heaviside step function (\cref{eq:heaviside}) and is, therefore, non-differentiable, which prevents direct gradient-based optimization. Prior works address this by smoothing the AP metric. Following \citet{brown2020smooth}, we define the smooth surrogate of the binary indicator function as
\begin{equation}
    \tilde{h}(\Delta s)= \sigma(\Delta s/\tau) = \frac{1}{1 + e^{-\Delta s / \tau}},
\label{eq:surrogate}
\end{equation}
where $\sigma(\cdot)$ is the sigmoid function, and $\tau$ is a temperature parameter controlling smoothness.
By using this approximation, the AP loss becomes differentiable and is optimized in an end-to-end manner. It is then straightforward to derive from~\cref{eq:map} the surrogate mAP loss as
\begin{equation}
\cL_{\mathrm{mSAP}} = 1 - \tilde{\mathrm{mAP}}(S, Y) = 1 - \frac{1}{M} \sum_{i=1}^{M} \tilde{\mathrm{AP}}(S_{i}, Y_{i}),
\end{equation}
where $\tilde{\mathrm{AP}}(\cdot, \cdot)$ is the AP function implemented with the smooth surrogate from~\cref{eq:surrogate}. Similarly, we can derive from~\cref{eq:uap} the smooth variant of \uAP, called \ours,
\begin{equation}
\cL_{\text{\ours}} = 1 - \tilde{\text{\uAP}}(S, Y) = 1 - \tilde{\text{AP}}(\vect(S), \vect(Y)).
\end{equation}
In that way, we obtain differentiable approximations of the two metrics that can be used to train deep similarity models.
Smaller (larger) $\tau$ yields a steeper (smoother) sigmoid function and a better (worse) approximation of the exact metric.
At the same time, a steep (smooth) sigmoid reaches saturation faster and has a larger (smaller) range where gradients are vanishing, and training is hindered.  

\subsection{Training with Global Average Precision}
\label{sec:training}

\paragraph{Representation learning.}
Encoders map inputs to a feature space where similarities between samples are estimated via cosine similarity. In the unimodal setting, an image encoder $f_\theta: \cX \rightarrow \real^d$ maps an image $x_i$ to an $\ell_2$-normalized embedding $\vx_i = f_\theta(x_i)$, where $\theta$ are the learnable parameters. The similarity between two images $x_i, x_j$ with embeddings $\vx_i, \vx_j \in \real^d$ is computed as the dot product $s(\vx_i, \vx_j) = \vx_i^\top \vx_j$, denoted $s_{ij}$. In the multimodal setting, separate encoders map inputs from different modalities into a shared embedding space. Given an image $x_i$ and a text description $t_j$, an image encoder $f_\theta: \cX \rightarrow \real^d$ and a text encoder $g_\phi: \cT \rightarrow \real^d$ are used to produce $\ell_2$-normalized embeddings $\vx_i = f_\theta(x_i)$ and $\vt_j = g_\phi(t_j)$. Cross-modal similarity is computed as $s(\vx_i, \vt_j) = \vx_i^\top \vt_j$.

Given a batch $\cB$ of size $B$, we compute the pairwise similarity matrix $S^{\cB} \in \real^{B \times B}$ and the corresponding binary label matrix $Y^{\cB} \in \{0,1\}^{B \times B}$ indicating whether pairs match. In the unimodal case, $\cB = [x_1,\dots,x_B]$ is a batch of images and $Y^{\cB}$ encodes class matches between image pairs. In the multimodal case, $\cB = [(x_1,t_1),\dots,(x_B,t_B)]$ consists of image--text pairs and $Y^{\cB}$ is typically the identity matrix indicating aligned image–text pairs. In the unimodal case, the diagonal entries are removed to avoid trivial self-matches.
The models are then trained by optimizing either the mSAP or \ours loss using $S^{\cB}$ and $Y^{\cB}$.

\begin{figure}[t]
    \centering
    \includegraphics[scale=0.9]{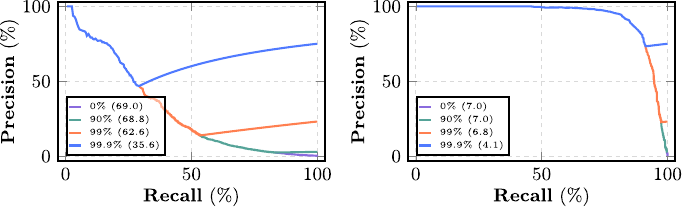}  
    \vspace{-8pt}
    \caption{\textbf{Analysis of negative dropping.} Precision--Recall curve computed on a single training batch using DINOv3 before (left) and after (right) training with varying percentages of lowest negative similarity dropping. The legend shows the drop rate $\alpha$ and the corresponding \ours loss (in parentheses).
    \label{fig:pr_curves}
    \vspace{-12pt}
    }
\end{figure}

\paragraph{Negative dropping.}
The cost of \ours is linear in the number of pairwise similarities in the batch, and nearly all of them are negatives.
The smallest negative similarities are already well separated from the positives and barely affect \uAP. Hence, to reduce computational and memory requirements in \ours, we drop the fraction $\alpha \in [0\%,100\%]$ of the smallest negative similarities. Dropping is implemented by removing these similarities, along with their corresponding labels, from the flattened similarity and label vectors used for the loss.
All positives are kept, and the optimization focuses on the negatives that still violate the ranking.
The rate $\alpha$ trades memory for fidelity to the full loss.
\Cref{fig:pr_curves} shows its effect on the PR curve of a training batch.
Removing up to $90\%$ of the negatives only alters the low-precision tail of the curve, and since the loss is the \emph{area above} the PR curve, a larger $\alpha$ simplifies the optimization and drives the loss towards zero.
In practice, large values of $\alpha$ cost minimal performance and under a fixed GPU memory budget they allow larger batches.

\begin{figure}[t]
    \centering
    \input{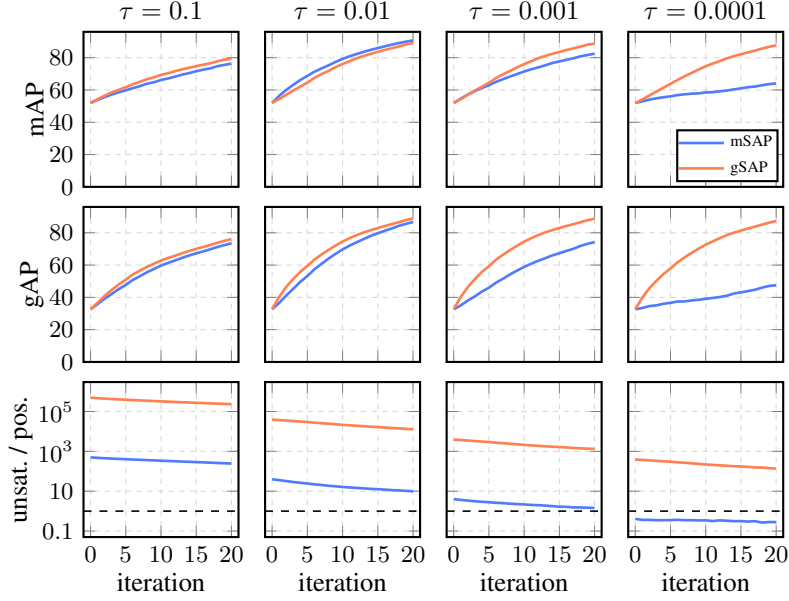}
    \vspace{-12pt}
    \caption{\textbf{Loss behavior analysis via single batch overfitting.}
    One SOP training batch ($250$ classes $\times$ $4$ images), DINOv3, $20$ updates, the same batch and initialization in every panel; columns are temperatures $\tau$, color is the loss ({\color{appleblue} mSAP}, {\color{appleorange} \ours}).
    Top and middle: exact mAP and \uAP on the batch, in \%.
    Bottom: average number of unsaturated comparisons per positive ($|\Delta s|/\tau<4$), \ie their total count divided by the number of positives, log scale. The dashed line indicates one unsaturated comparison per positive.
    \label{fig:loss_comparison}
    \vspace{-15pt}
    }
\end{figure}

\subsection{Analysis by single batch overfitting}
\label{sec:supp_overfit}

To better understand the behavior of the surrogate losses during optimization, we conduct a toy experiment in which the model is deliberately overfitted on a single fixed batch, following the protocol of \Cref{fig:teaser}(b). For each loss, we optimize the model for $20$ iterations at four temperatures $\tau$, from the same initialization and on the same batch. \Cref{fig:loss_comparison} tracks the mAP and \uAP on the batch and the average number of unsaturated comparisons per positive. More precisely, a \emph{comparison} is one term $\tilde{h}(s_j - s_i)$ of the surrogate of \cref{eq:precision}, between a positive $i\in\cP(\vy)$ and an item $j\in\Omega$ of the list the loss ranks, \ie a row of $S$ for mSAP and $\vect(S)$ for \ours. A comparison is \emph{unsaturated} when $|s_j - s_i| < 4\tau$, \ie when it still carries non-zero gradient during back-propagation\footnote{Each comparison contributes to the loss through $\sigma(\Delta s/\tau)$, so its gradient with respect to $\Delta s$ is proportional to $\sigma'(\Delta s/\tau)$, which peaks at $\sigma'(0)=0.25$ and decays to $\sigma'(4)\approx0.018$, \ie $7\%$ of the peak. The cut at $4$ is a convention, as comparisons beyond it carry almost no gradient~\citep{ramzi2021robust}.}.

We draw the following conclusions: 
(i) \textbf{\ours improves both metrics, mSAP lags behind in \uAP.} For $\tau\le0.01$, where the surrogate approximates its metric, mAP and \uAP move together under \ours, whereas under mSAP \uAP trails mAP by a considerable margin, increasingly so as $\tau$ decreases. mSAP is insensitive to the cross-query consistency that \uAP depends on, so it improves \uAP only as a side effect of better per-query rankings. \ours, on the other hand, ranks the whole batch and moves both, at no cost in mAP, which is clearly above that of mSAP at three of the four temperatures.
\noindent (ii) \textbf{\ours is robust to $\tau$ because its positives have more unsaturated comparisons.} A small $\tau$ makes the surrogate exact but saturates the sigmoid comparisons~\citep{qin2010general,brown2020smooth}. Empirically, the \emph{share} of unsaturated comparisons is nearly the same for both losses, about $50\%$, $4\%$, $0.4\%$ and $0.04\%$ at the four temperatures at initialization, roughly proportional to $\tau$, and it stays within a factor of two between the losses during training. What differs, by three orders of magnitude, is the \emph{number} of comparisons per positive that enter the loss. mSAP scores each positive against the $B-2\approx10^{3}$ other similarities of its row, \ours against the $B(B-1)\approx10^{6}$ similarities of the whole batch. The bottom row of the figure shows the product of the two, the average number of unsaturated comparisons per positive. At $\tau=10^{-4}$ an mSAP positive has fewer than one, \ie most positives receive no gradient, and the run stalls well below the other settings. By contrast, a \ours positive still has more than $150$ unsaturated comparisons, every positive keeps a gradient, and the run reaches the same mAP and \uAP as at the other values of $\tau$ (validated in \Cref{fig:ablation}). This is why \ours can operate in the regime of near-exact approximation that mSAP cannot.
\noindent (iii) \textbf{At $\tau=0.1$ both surrogates are poor proxies of their metric.} Both runs end well below the others. Gradients are plentiful, but the approximation is loose and the optimization noisy. Even in this regime, \ours reaches slightly higher values of both metrics.
\noindent In summary, mSAP suffers more from the trade-off between an accurate approximation of the metric and the difficulty of optimizing it.

\section{Experiments}
\label{sec:experiments}

\paragraph{Cross-modal learning.}
We study vision--language representation learning by fine-tuning the released OpenAI CLIP ViT-B/32~\citep{radford2021learning}. Starting from the pretrained model, we train for 5 additional epochs on the Conceptual 12M (CC12M)~\citep{changpinyo2021conceptual} dataset using three alternative losses: (i) the original CLIP InfoNCE-based loss~\citep{oord2018representation}, (ii) the SigLIP sigmoid loss~\citep{zhai2023sigmoid}, and (iii) our \ours loss. We evaluate these models in zero-shot retrieval and classification. We follow the CLIP benchmark suite~\citep{cherti_clip_bench}, where applicable. Additional experiments with other vision--language models, further evaluation tasks and analyses, as well as experiments for audio--language alignment, are provided in \Cref{sec:supp_vlm}. 

\begin{table}[t]
    \centering
    \caption{\textbf{Zero-shot retrieval on image datasets.} \uAP, mAP, and Recall@5 comparison of \ours with CLIP and SigLIP losses. All models fine-tune the pretrained OpenAI CLIP ViT-B/32 for 5 epochs on CC12M.}
    \vspace{-8pt}
    \label{tab:sota_vlm_image_retrieval_vitb32}
    \hspace{-10pt}
    \scalebox{0.9}{
    \newcolumntype{C}{>{\centering\arraybackslash}p{21pt}}
\scriptsize
\setlength{\tabcolsep}{0.8pt}
\begin{tabular}{lCCCCCCCCCCCCCCC}
\toprule
\multirow{2}{*}{\textbf{method}} &
\multicolumn{5}{c}{\textbf{Flickr8k}} &
\multicolumn{5}{c}{\textbf{Flickr30k}} &
\multicolumn{5}{c}{\textbf{MS COCO}} \\
\cmidrule(lr){2-6}\cmidrule(lr){7-11}\cmidrule(lr){12-16}
& \multirow{2}{*}{\uAP} & mAP t$\rightarrow$i & mAP i$\rightarrow$t & R@5 t$\rightarrow$i & R@5 i$\rightarrow$t
& \multirow{2}{*}{\uAP} & mAP t$\rightarrow$i & mAP i$\rightarrow$t & R@5 t$\rightarrow$i & R@5 i$\rightarrow$t
& \multirow{2}{*}{\uAP} & mAP t$\rightarrow$i & mAP i$\rightarrow$t & R@5 t$\rightarrow$i & R@5 i$\rightarrow$t \\
\midrule
\textbf{pretrained} &
31.0 & 65.0 & 57.7 & 79.2 & 92.0 &
45.3 & 68.7 & 62.7 & 82.9 & 94.0 &
13.1 & 42.1 & 36.7 & 55.0 & 73.4 \\
\midrule

\textbf{$+$ CLIP} &
41.3 & 71.9 & 63.5 & 85.5 & 92.4 &
56.1 & 75.1 & 68.3 & 87.6 & 94.7 &
18.7 & 49.6 & 41.1 & 63.6 & 76.9 \\

\textbf{$+$ SigLIP} &
43.2 & 72.5 & 64.6 & 86.0 & 92.7 &
58.5 & 76.3 & 69.5 & 88.5 & 95.7 &
19.4 & 50.6 & 41.8 & 64.7 & 77.5 \\

\rowcolor{appleteal!60}
\textbf{$+$ \ours} &
\textbf{45.1} & \textbf{73.2} & \textbf{65.6} & \textbf{86.7} & \textbf{93.3} &
\textbf{59.7} & \textbf{76.6} & \textbf{70.5} & \textbf{88.9} & \textbf{96.0} &
\textbf{21.4} & \textbf{51.3} & \textbf{43.1} & \textbf{65.3} & \textbf{79.0} \\

\bottomrule
\end{tabular}

    }
    \vspace{-7pt}
\end{table}

\begin{table}[t]
    \centering
    \caption{\textbf{Zero-shot retrieval on video datasets.} \uAP and Recall@$k$ comparison. Same setting as \Cref{tab:sota_vlm_image_retrieval_vitb32}.}
    \vspace{-8pt}
    \label{tab:sota_vlm_video_retrieval_vitb32}
    \scalebox{0.9}{
    \newcolumntype{C}{>{\centering\arraybackslash}p{17pt}}
\scriptsize
\setlength{\tabcolsep}{0.8pt}
\begin{tabular}{lCCCCCCCCCCCCCCCCCCCCC}
\toprule
\multirow{3}{*}{\textbf{method}} &
\multicolumn{7}{c}{\textbf{ActivityNet}} &
\multicolumn{7}{c}{\textbf{MSR-VTT}} &
\multicolumn{7}{c}{\textbf{MSVD}} \\
\cmidrule(lr){2-8}\cmidrule(lr){9-15}\cmidrule(lr){16-22}
& \multirow{2}{*}{\uAP} & \multicolumn{3}{c}{text$\to$video} & \multicolumn{3}{c}{video$\to$text}
& \multirow{2}{*}{\uAP} & \multicolumn{3}{c}{text$\to$video} & \multicolumn{3}{c}{video$\to$text}
& \multirow{2}{*}{\uAP} & \multicolumn{3}{c}{text$\to$video} & \multicolumn{3}{c}{video$\to$text} \\
\cmidrule(lr){3-5}\cmidrule(lr){6-8}\cmidrule(lr){10-12}\cmidrule(lr){13-15}\cmidrule(lr){17-19}\cmidrule(lr){20-22}
& & R@1 & R@5 & R@10 & R@1 & R@5 & R@10
& & R@1 & R@5 & R@10 & R@1 & R@5 & R@10
& & R@1 & R@5 & R@10 & R@1 & R@5 & R@10 \\
\midrule
\textbf{pretrained} &
10.0 & 22.4 & 49.8 & 64.1 & 21.3 & 49.4 & 61.2 &
4.1 & 14.3 & 30.4 & 38.9 & 29.4 & 51.9 & 61.1 &
14.5 & 32.8 & 60.5 & 69.9 & 51.2 & 71.6 & 77.8 \\
\midrule
\textbf{$+$ CLIP} &
13.6 & 27.7 & 61.0 & 72.8 & 26.7 & 57.6 & 70.3 &
4.1 & 16.2 & 33.8 & 42.8 & 28.0 & 49.2 & 59.5 &
22.0 & 38.9 & 67.4 & 76.9 & 50.4 & 73.7 & 81.8 \\
\textbf{$+$ SigLIP} &
13.0 & 28.8 & 62.0 & 74.2 & 25.9 & 56.0 & 69.1 &
4.2 & 16.6 & 34.5 & 43.5 & 28.6 & 50.9 & 61.1 &
21.4 & 39.8 & 67.8 & 77.1 & 52.3 & 73.6 & 81.1 \\
\rowcolor{appleteal!60}
\textbf{$+$ \ours} &
\textbf{14.9} & \textbf{28.9} & \textbf{62.2} & \textbf{74.6} & \textbf{27.5} & \textbf{57.8} & \textbf{70.5} &
\textbf{4.7} & \textbf{17.0} & \textbf{35.0} & \textbf{44.0} & \textbf{30.2} & \textbf{52.3} & \textbf{62.0} &
\textbf{23.2} & \textbf{39.9} & \textbf{68.4} & \textbf{77.4} & \textbf{52.7} & \textbf{73.9} & \textbf{82.1} \\
\bottomrule
\end{tabular}

    }
    \vspace{-7pt}
\end{table}

\begin{table}[h!]
    \centering
    \caption{\textbf{Zero-shot classification on image and video datasets.} Top-1, top-5, and mean per-class recall (mR) comparison of \ours with CLIP (InfoNCE) and SigLIP losses. Same setting as \Cref{tab:sota_vlm_image_retrieval_vitb32}.}
    \label{tab:sota_vlm_classification_vitb32}
    \vspace{-8pt}
    \scalebox{0.9}{
    \newcolumntype{C}{>{\centering\arraybackslash}p{17pt}}
\scriptsize
\setlength{\tabcolsep}{1.5pt}
\begin{tabular}{lCCCCCCCCCCCCCCCCCC}
\toprule
\multirow{3}{*}{\textbf{method}} &
\multicolumn{12}{c}{\textbf{image datasets}} &
\multicolumn{6}{c}{\textbf{video datasets}} \\
\cmidrule(lr){2-13}\cmidrule(lr){14-19}
& \multicolumn{3}{c}{\textbf{IN-1k}} &
\multicolumn{3}{c}{\textbf{IN-V2}} &
\multicolumn{3}{c}{\textbf{IN-R}} &
\multicolumn{3}{c}{\textbf{IN-Sketch}} &
\multicolumn{2}{c}{\textbf{K-400}} &
\multicolumn{2}{c}{\textbf{K-600}} &
\multicolumn{2}{c}{\textbf{K-700}} \\
\cmidrule(lr){2-4}\cmidrule(lr){5-7}\cmidrule(lr){8-10}\cmidrule(lr){11-13}
\cmidrule(lr){14-15}\cmidrule(lr){16-17}\cmidrule(lr){18-19}
& top-1 & top-5 & mR
& top-1 & top-5 & mR
& top-1 & top-5 & mR
& top-1 & top-5 & mR
& top-1 & top-5
& top-1 & top-5
& top-1 & top-5 \\
\midrule
\textbf{pretrained} &
59.6 & 86.4 & 59.6 &
52.9 & 80.8 & 52.9 &
67.0 & 87.2 & 64.9 &
38.9 & 66.9 & 39.0 &
38.4 & 64.3 &
17.8 & 33.1 &
27.9 & 51.1 \\
\midrule
\textbf{$+$ CLIP} &
60.7 & 87.4 & 60.7 &
53.6 & 82.0 & 53.6 &
68.3 & 87.9 & 66.8 &
42.3 & 70.4 & 42.3 &
39.4 & 65.8 &
18.6 & 34.3 &
29.1 & 53.5 \\
\textbf{$+$ SigLIP} &
61.3 & 87.8 & 61.3 &
54.1 & 82.1 & 54.2 &
69.1 & 88.2 & 67.8 &
43.2 & 71.6 & 43.3 &
39.8 & 66.3 &
19.0 & 34.8 &
29.5 & 54.1 \\
\rowcolor{appleteal!60}
\textbf{$+$ \ours} &
\textbf{61.8} & \textbf{87.9} & \textbf{61.8} &
\textbf{54.3} & \textbf{82.5} & \textbf{54.3} &
\textbf{69.4} & \textbf{88.3} & \textbf{68.1} &
\textbf{43.7} & \textbf{71.7} & \textbf{43.7} &
\textbf{40.2} & \textbf{66.5} &
\textbf{19.2} & \textbf{34.9} &
\textbf{29.8} & \textbf{54.4} \\
\bottomrule
\end{tabular}

    }
    \vspace{-14pt}
\end{table}

\noindent\textit{Zero-shot retrieval.}
We evaluate bidirectional retrieval, \ie, text$\to$image/video and image/video$\to$text, on three image datasets, Flickr8k~\citep{flickr}, Flickr30k~\citep{flickr30k}, and MS COCO Captions~\citep{cococaptions} (\Cref{tab:sota_vlm_image_retrieval_vitb32}), and three video datasets, ActivityNet~\citep{activitynet}, MSR-VTT~\citep{msrvtt}, and MSVD~\citep{msvd} (\Cref{tab:sota_vlm_video_retrieval_vitb32}). For image datasets, we report \uAP, mAP, and Recall@5; for video datasets, we report \uAP and Recall@$k$. Across all six datasets, \ours delivers the strongest retrieval performance, with consistent gains over both InfoNCE and SigLIP in every metric and direction.

\noindent\textit{Zero-shot classification.}
We evaluate on four image datasets, ImageNet-1k~\citep{deng2009imagenet}, ImageNet-V2~\citep{recht2019imagenetv2}, ImageNet-R~\citep{hendrycks2021many}, and ImageNet-Sketch~\citep{wang2019learning}; and three video datasets, Kinetics-400~\citep{kay2017kinetics}, -600~\citep{carreira2018short}, and -700~\citep{carreira2019short} (\Cref{tab:sota_vlm_classification_vitb32}), using the original prompt templates for each dataset. Across all seven datasets, \ours again provides the most consistent gains, outperforming both InfoNCE and SigLIP in every metric. Beyond these seven datasets, we also evaluate on the full suite of 40 datasets of the benchmark~\citep{cherti_clip_bench} (37 classification and 3 retrieval), spanning natural, fine-grained, satellite, medical, and synthetic imagery. Aggregating all 108 scores (\Cref{tab:full_benchmark_vitb32}), \ours improves on average by +0.49 over InfoNCE and +0.64 over SigLIP, indicating that the benefits of optimizing \uAP also translate to improvements in recognition.

\begin{wraptable}{r}{0.52\linewidth}
    \centering
    \vspace{-12pt}
    \caption{\textbf{Image classification with self-supervised pretraining.} Top-1 accuracy of models trained with MoCo~v3 using InfoNCE or \ours. Performance of the frozen backbone with linear probing (LP) and $k$-nearest neighbors ($k$NN, $k{=}20$) on in-distribution (ID, ImageNet-1k), out-of-distribution (OOD, mean of ImageNet-V2/-R/-Sketch), transfer learning (ten datasets), and robustness under corruption (ImageNet-C, 75 sets).}
    \vspace{-8pt}
    \label{tab:sota_vision_ssl}
    \scalebox{0.94}{\newcolumntype{C}{>{\centering\arraybackslash}p{12pt}}
\scriptsize
\setlength{\tabcolsep}{3pt}
\begin{tabular}{lcCCCCCCCC}
    \toprule
    \multirow{2}{*}{\textbf{method}} & \multirow{2}{*}{\textbf{model}} & \multicolumn{2}{c}{\textbf{ID}}  & \multicolumn{2}{c}{\textbf{OOD}} & \multicolumn{2}{c}{\textbf{transfer}} & \multicolumn{2}{c}{\textbf{robustness}} \\ \cmidrule(lr){3-4} \cmidrule(lr){5-6}  \cmidrule(lr){7-8}  \cmidrule(lr){9-10}
    & & LP & $k$NN & LP & $k$NN & LP & $k$NN & LP & $k$NN \\ \midrule
    \textbf{InfoNCE}      & ViT-S & 73.2 & 68.4 & 38.2 & 31.5 & 69.8 & 61.0 & 44.6 & 39.4 \\
    \rowcolor{appleteal!60}
    \textbf{\ours}        & ViT-S & \textbf{73.5} & \textbf{70.9} & \textbf{39.7} & \textbf{35.4} & \textbf{73.9} & \textbf{65.5} & \textbf{46.3} & \textbf{44.7} \\\midrule
    \textbf{InfoNCE}      & ViT-B & \textbf{76.7} & 71.4 & \textbf{42.0} & 36.6 & 74.8 & 62.5 & 49.6 & 47.0 \\
    \rowcolor{appleteal!60}
    \textbf{\ours}        & ViT-B & 75.6 & \textbf{72.8} & \textbf{42.0} & \textbf{38.1} & \textbf{77.8} & \textbf{66.8} & \textbf{49.7} & \textbf{48.6} \\ 
    \bottomrule
\end{tabular}
}
    \vspace{-12pt}
\end{wraptable}

\paragraph{Self-supervised learning.}
In addition, we test whether the benefit survives without labels by following the MoCo~v3~\citep{chen2021empirical} recipe and replacing InfoNCE by \ours, with everything else unchanged. \Cref{tab:sota_vision_ssl} reports top-1 accuracy of frozen backbones under linear probing (LP) and $k$-nearest neighbors ($k$NN). Per-dataset results and settings are in \Cref{sec:supp_ssl,sec:ssl_settings}.
Three observations follow: (i) The gain carries over to the label-free setting, where \ours trails only on in-distribution LP at ViT-B. (ii) The gain is consistently larger under $k$NN, which operates on the learned similarities directly, whereas LP fits a new linear map that can absorb a similarity scale that differs from query to query. (iii) The transfer gain is considerable and concentrated on the six fine-grained datasets (\Cref{tab:ssl_transfer}), where hard negatives crowd the positives, which is what a single shared threshold penalizes. Furthermore, the same substitution improves music tagging in audio contrastive learning, so \ours is not tied to a particular modality (\Cref{sec:supp_ssl}).

\paragraph{Metric learning.}
We compare \ours with competing losses on two standard benchmarks, Stanford Online Products (SOP)~\citep{oh2016deep} and iNaturalist~\citep{van2018inaturalist}, following the splits and setup of prior work~\citep{brown2020smooth,patel2022recall}. Evaluation settings and implementation details are provided in \Cref{sec:metric_learning_impl}.

\begin{wraptable}{r}{0.54\linewidth}
    \centering
    \vspace{-12pt}
    \caption{\textbf{Metric learning losses comparison.} Recall@1 (R@1), mAP and \uAP on two datasets. \ours is compared with ranking, pairwise, and proxy losses. Fine-tuning of DINOv3. No negative dropping is applied.}
    \vspace{-5pt}
    \scalebox{1.2}{
        \label{tab:sota}
    }
    \scriptsize
\setlength{\tabcolsep}{2.5pt}
\begin{tabular}{l ccc ccc}
    \toprule
    \multirow{2}{*}{\textbf{method}} & \multicolumn{3}{c}{\textbf{SOP}} & \multicolumn{3}{c}{\textbf{iNaturalist}} \\ \cmidrule(lr){2-4} \cmidrule(lr){5-7}
    & R@1 & mAP & \uAP & R@1 & mAP & \uAP \\ \midrule
    \textbf{Contrastive} & 85.3 & 68.4 & 53.7 & 80.6 & 42.1 & 28.0 \\
    \textbf{Triplet} & 84.6 & 69.1 & 50.0 & 77.8 & 38.6 & 18.2 \\
    \textbf{ArcFace} & 82.7 & 64.8 & 50.3 & 80.2 & 42.4 & 23.0 \\
    \textbf{Proxy Anchor} & 84.7 & 67.7 & 51.9 & 78.7 & 40.9 & 21.8 \\
    \textbf{Multi-Similarity} & 86.6 & 72.0 & 57.3 & \textbf{81.6} & 45.2 & 29.9 \\
    \textbf{RS@$k$} & 86.7 & 72.5 & 56.7 & 81.0 & 44.5 & 26.6 \\
    \textbf{mSAP} & \textbf{86.9} & \textbf{72.9} & 57.0 & 81.3 & 45.4 & 27.8 \\
    \rowcolor{appleteal!60} \textbf{\ours} & 86.8 & \textbf{72.9} & \textbf{58.9} & 81.5 & \textbf{45.5} & \textbf{31.6} \\
    \bottomrule
\end{tabular}

\end{wraptable}

\noindent\textit{Comparison with metric learning losses.}
\Cref{tab:sota} compares \ours with pairwise (Contrastive~\citep{chopra2005learning}, Triplet~\citep{schroff2015facenet}, Multi-Similarity~\citep{wang2019multi}), proxy (ArcFace~\citep{deng2019arcface}, Proxy Anchor~\citep{kim2020proxy}) and ranking losses (the Recall@$k$ surrogate RS@$k$~\citep{patel2022recall}, mSAP~\citep{brown2020smooth}) under fine-tuning of DINOv3~\citep{simeoni2025dinov3}. \ours achieves the best \uAP on both datasets by a clear margin, while its mAP and R@1 remain close to the best. The ranking losses match \ours on mAP and R@1, so their lower \uAP stems not from worse per-query ranking but from similarity distributions that are less consistent across queries, a property that \uAP captures and mAP and R@1 do not.
We delve into a theoretical analysis of the losses in \Cref{sec:analysis}, in terms of their invariance to per-query similarity and the theoretical best \uAP that can be achieved given the per-query ranking.

\begin{figure}[t]
    \centering
    \vspace{-4pt}
    \pgfplotsset{every tick label/.append style={font=\footnotesize}}
\pgfplotsset{
    recallh/.style={color=applemustard, line width=1.0pt, const plot mark mid, no marks},
    saph/.style={color=applepink, line width=1.0pt, const plot mark mid, no marks},
    sgaph/.style={color=appleteal!90!black, line width=1.2pt, const plot mark mid, no marks},
}
\input{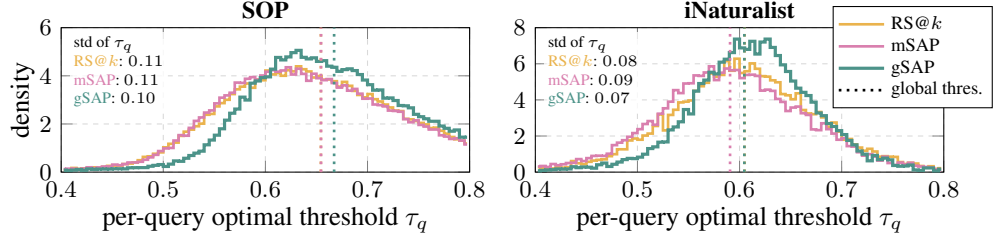}

\begin{tabular}{@{}c@{\hspace{4pt}}c@{}}
\begin{tikzpicture}
    \begin{axis}[%
      width=0.5\linewidth, height=0.25\linewidth,
      ylabel={density}, xlabel={per-query optimal threshold $\tau_q$},
      grid=both, grid style={color=lightgray!60, dash pattern=on 2pt off 2pt},
      xlabel style={yshift=3pt}, ylabel style={yshift=-5pt},
      xmin=0.40, xmax=0.80, ymin=0, ymax=6,
      xtick={0.4,0.5,0.6,0.7,0.8},
      title={\textbf{SOP}}, title style={yshift=-6pt, font=\footnotesize},
      ]
    \addplot[recallh] table[x=tau,y=rsk] {\tauqsop};
    \addplot[saph] table[x=tau,y=msap] {\tauqsop};
    \addplot[sgaph] table[x=tau,y=gsap] {\tauqsop};
    \node[anchor=north west, font=\tiny, align=left, inner sep=1.5pt, fill=white, fill opacity=0.8, text opacity=1] at (rel axis cs:0.02,0.97) {std of $\tau_q$\\{\color{applemustard}RS@$k$}: $0.11$\\{\color{applepink}mSAP}: $0.11$\\{\color{appleteal!90!black}\ours}: $0.10$};
    \draw[dotted, color=applemustard, line width=1.0pt] (axis cs:0.6543,0) -- (axis cs:0.6543,6);
    \draw[dotted, color=applepink, line width=1.0pt] (axis cs:0.6545,0) -- (axis cs:0.6545,6);
    \draw[dotted, color=appleteal!90!black, line width=1.0pt] (axis cs:0.6673,0) -- (axis cs:0.6673,6);
    \end{axis}
\end{tikzpicture}
&
\begin{tikzpicture}
    \begin{axis}[%
      width=0.5\linewidth, height=0.25\linewidth,
      xlabel={per-query optimal threshold $\tau_q$},
      grid=both, grid style={color=lightgray!60, dash pattern=on 2pt off 2pt},
      xlabel style={yshift=3pt}, ylabel style={yshift=-4pt},
      xmin=0.40, xmax=0.80, ymin=0, ymax=8,
      xtick={0.4,0.5,0.6,0.7,0.8},
      title={\textbf{iNaturalist}}, title style={yshift=-6pt, font=\footnotesize},
      legend style={at={(1.12,1.15)}, anchor=north east, draw=black, cells={anchor=west}, font=\scriptsize, inner sep=1.5pt, row sep=-2pt},
      ]
    \addplot[recallh] table[x=tau,y=rsk] {\tauqinat};
    \addlegendentry{RS@$k$}
    \addplot[saph] table[x=tau,y=msap] {\tauqinat};
    \addlegendentry{mSAP}
    \addplot[sgaph] table[x=tau,y=gsap] {\tauqinat};
    \node[anchor=north west, font=\tiny, align=left, inner sep=1.5pt, fill=white, fill opacity=0.8, text opacity=1] at (rel axis cs:0.02,0.97) {std of $\tau_q$\\{\color{applemustard}RS@$k$}: $0.08$\\{\color{applepink}mSAP}: $0.09$\\{\color{appleteal!90!black}\ours}: $0.07$};
    \addlegendentry{\ours}
    \addlegendimage{dotted, thick, color=black, line width=1.0pt}
    \addlegendentry{global thres.}
    \draw[dotted, color=applemustard, line width=1.0pt] (axis cs:0.6050,0) -- (axis cs:0.6050,8);
    \draw[dotted, color=applepink, line width=1.0pt] (axis cs:0.5907,0) -- (axis cs:0.5907,8);
    \draw[dotted, color=appleteal!90!black, line width=1.0pt] (axis cs:0.6047,0) -- (axis cs:0.6047,8);
    \end{axis}
\end{tikzpicture}
\end{tabular}
    \vspace{-12pt}
    \caption{\textbf{Per-query optimal thresholds.} Distribution of the F1-optimal threshold of each evaluation query. Dotted lines indicate the shared threshold selected on the fitting classes. Standard deviation of $\tau_q$ per loss in the top-left of each panel.}
    \label{fig:tauq}
    \vspace{-10pt}
\end{figure}

\begin{wraptable}{l}{0.5\linewidth}
    \vspace{-12pt}
    \centering
    \caption{\textbf{One threshold for all queries.} F1$_q$ is the mean per-query F1 when every query uses its own optimal threshold. F1$_g$ is the same mean when all queries share one threshold, \ie it is applied globally, selected on the fitting classes. R@P90 is the recall of that shared threshold set for precision $0.90$. 
    \label{tab:threshold}
    \vspace{-7pt}
    }
    \scriptsize
\setlength{\tabcolsep}{3.5pt}
\begin{tabular}{l ccc ccc}
    \toprule
    \multirow{2}{*}{\textbf{method}} & \multicolumn{3}{c}{\textbf{SOP}} & \multicolumn{3}{c}{\textbf{iNaturalist}} \\ \cmidrule(lr){2-4} \cmidrule(lr){5-7}
    & F1$_q$ & F1$_g$ & R@P90 & F1$_q$ & F1$_g$ & R@P90 \\ \midrule
    \textbf{RS@$k$} & 75.8 & 57.5 & 18.3 & 47.7 & 35.8 & 0.6 \\
    \textbf{mSAP} & 76.2 & 57.6 & 18.7 & 48.5 & 36.6 & 0.7 \\
    \rowcolor{appleteal!60} \textbf{\ours} & \textbf{76.2} & \textbf{59.3} & \textbf{21.0} & \textbf{48.7} & \textbf{38.3} & \textbf{3.0} \\
    \bottomrule
\end{tabular}

    \vspace{-10pt}
\end{wraptable}

\noindent\textit{Consistency and thresholding.}
Consistency matters wherever one threshold has to serve every query. \Cref{tab:threshold} shows this for the three ranking losses, with the test classes split into fitting classes, on which a threshold is selected, and evaluation classes, on which it is applied (\Cref{sec:metric_learning_impl}). When each query is allowed its own optimal threshold, the three losses perform on par on both datasets, in line with mAP and R@1, since the decision is then made within each query's ranking. With a single threshold, selected on the fitting classes and shared by all evaluation queries, \ours outperforms both other losses on both datasets. Notably, at a precision of $0.90$, it retrieves $12\%$ more matches than mSAP, the second best, on SOP and $4\times$ as many on iNaturalist. \Cref{fig:tauq} indicates the reason for its effectiveness. The per-query optimal thresholds of \ours are more concentrated around the shared one, so a single value fits more queries. The gap between F1$_q$ and F1$_g$ is the cost of inconsistent similarities, and \ours reduces it without changing what happens inside each query.

\begin{table}[t]
    \vspace{-2pt}
    \centering
    \caption{\textbf{Comparison with the metric learning state-of-the-art.} Recall@$k$ (R@$k$) of the vanilla losses as reported in the original papers on two datasets and on two backbones. R50 stands for ResNet50. Dimensionality of the output representations is indicated in the superscript. \textbf{Best} in bold, second best \underline{underlined}. 4{,}000 batch size and $\alpha$=95\% drop rate are used.
    \label{tab:sota_rep}
    \vspace{-8pt}
    }
    \scalebox{0.87}{
    \newcolumntype{C}{>{\centering\arraybackslash}p{25pt}}
\scriptsize
\setlength{\tabcolsep}{4pt}
\begin{tabular}{lcCCCCCC}
    \toprule
    \multirow{2}{*}{\textbf{method}} & \multirow{2}{*}{\textbf{arch.\textsuperscript{dim}}} & \multicolumn{3}{c}{\textbf{SOP}} & \multicolumn{3}{c}{\textbf{iNaturalist}} \\ \cmidrule(lr){3-5} \cmidrule(lr){6-8}
    & & R@1 & R@10 & R@100 & R@1 & R@4 & R@16 \\ \midrule
    \textbf{Triplet SH}~\citep{wu2017sampling} & R50\textsuperscript{512} & 72.7 & 86.2 & 93.8 & 58.1 & 75.5 & 86.8 \\ 
    \textbf{NormSoftmax}~\citep{zhai2018classification} & R50\textsuperscript{512} & 78.2 & 90.6 & 96.2 & -- & -- & -- \\ 
    \textbf{FastAP}~\citep{cakir2019deep} & R50\textsuperscript{512}      & 76.4 & 89.0 & 95.1 & 60.6 & 77.0 & 87.2 \\
    \textbf{Blackbox}~\citep{rolinek2020optimizing} & R50\textsuperscript{512} & 78.6 & 90.5 & 96.0 & 62.9 & 79.0 & 88.9 \\
    \textbf{SoftBin}~\citep{revaud2019learning} & R50\textsuperscript{512} & 80.6 & 91.3 & 96.1 & 64.2 & 77.1 & 82.7 \\
    \textbf{ProxyNCA}~\citep{movshovitz2017no} & R50\textsuperscript{512} & 73.7 & -- & -- & 61.6 & 77.4 & 87.0 \\
    \textbf{ProxyNCA++}~\citep{teh2020proxynca++}  & R50\textsuperscript{512} & 80.7 & 92.5 & 96.7 & -- & -- & -- \\
    \textbf{SAP}~\citep{brown2020smooth}    & R50\textsuperscript{512} & 80.1 & 91.5 & 96.6 & 67.2 & 81.8 & 90.3   \\
    \textbf{ROADMAP}~\citep{ramzi2021robust}    & R50\textsuperscript{512} & \underline{83.1} & 92.7 & 96.3 & 69.1 & 83.1 & \textbf{91.3}  \\
    \textbf{RS@$k$}~\citep{patel2022recall} & R50\textsuperscript{512} & 82.8 & \underline{92.9} & \textbf{97.0} & \underline{71.2} & \underline{84.0} & \textbf{91.3}    \\
    \textbf{HIER}~\citep{kim2023hier} & R50\textsuperscript{512} & 80.2 & 91.5 & 96.6 & -- & -- & -- \\
    \textbf{Potential Field}~\citep{bhatnagar2025potential} & R50\textsuperscript{512} &  82.9 &  92.5 & 96.8 & -- & -- & -- \\
    \rowcolor{appleteal!60}
    \textbf{\ours}                  & R50\textsuperscript{512} &  \textbf{83.3} &  \textbf{93.2} &  \textbf{97.0} &  \textbf{71.3} &  \textbf{84.1} &  \textbf{91.3}  \\ \midrule
    \textbf{SAP}~\citep{brown2020smooth}    & ViT-B/16\textsuperscript{512} & 86.6 & 95.4 & 98.4 & 79.1 & 89.0 & 94.2  \\
    \textbf{RS@$k$}~\citep{patel2022recall} & ViT-B/16\textsuperscript{512} & \underline{88.0} & \underline{96.1} & \underline{98.6} & \underline{83.9} & \underline{92.1} & \underline{95.9}  \\
    \rowcolor{appleteal!60}
    \textbf{\ours}                  & ViT-B/16\textsuperscript{512} & \textbf{88.4} & \textbf{96.3} & \textbf{98.7} & \textbf{84.9} & \textbf{92.8} & \textbf{96.5} \\
    \bottomrule
\end{tabular}

    }
    \vspace{-7pt}
\end{table}

\noindent\textit{Comparison with state-of-the-art methods.} \Cref{tab:sota_rep} compares \ours with the state of the art on SOP and iNaturalist using ResNet50~\citep{he2016deep} and ViT-B/16~\citep{dosovitskiy2020image} backbones. For these experiments, we leverage the proposed negative dropping strategy with 95\% drop rate, to scale the batch size to 4{,}000 on a single GPU. \ours ranks among the state of the art across both datasets and architectures, showing that optimizing the \uAP surrogate also yields strong performance on the standard retrieval settings.\vspace{-1pt}

\begin{table}[t]
    \centering
    \caption{\textbf{Tasks with distractor queries.} \uAP: all queries. $\uAP^{-}$: distractor queries removed. {\color{gray!65!black} drop}: relative drop (\%); lower is better. SN: after score normalization~\citep{pizzi2022self}. \textbf{Best} in bold, second best \underline{underlined}.}
    \label{tab:rebuttal_tasks}
    \vspace{-8pt}
    \scalebox{0.87}{
    \newcolumntype{C}{>{\centering\arraybackslash}p{18pt}}
\newcolumntype{D}{>{\leavevmode\color{gray!65!black}\centering\arraybackslash}p{18pt}}
\scriptsize
\setlength{\tabcolsep}{3pt}
\begin{tabular}{l CCD CCD CCD CCD CCD}
    \toprule
    \multirow{2}{*}{\textbf{method}} & \multicolumn{6}{c}{\textbf{DISC21}} & \multicolumn{3}{c}{\textbf{Met}} & \multicolumn{3}{c}{\textbf{ILIAS}} & \multicolumn{3}{c}{\textbf{WildlifeReID-10k}} \\ \cmidrule(lr){2-7} \cmidrule(lr){8-10} \cmidrule(lr){11-13} \cmidrule(lr){14-16}
    & \uAP & $\uAP^{-}$ & \multicolumn{1}{c}{drop$\downarrow$} & $\uAP_{\text{SN}}$ & $\uAP^{-}_{\text{SN}}$ & \multicolumn{1}{c}{drop$\downarrow$} & \uAP & $\uAP^{-}$ & \multicolumn{1}{c}{drop$\downarrow$} & \uAP & $\uAP^{-}$ & \multicolumn{1}{c}{drop$\downarrow$} & \uAP & $\uAP^{-}$ & \multicolumn{1}{c}{drop$\downarrow$} \\ \midrule
    \textbf{baseline} & 51.6 & 64.6 & \underline{20.1} & 65.5 & 70.1 & \textbf{6.6} & 30.4 & 52.9 & 42.5 & 3.5 & 17.1 & 79.5 & 69.5 & 73.9 & 6.0 \\
    \textbf{InfoNCE} & 50.2 & 66.0 & 23.9 & 67.5 & \underline{73.3} & 7.9 & 56.2 & 72.2 & 22.2 & \underline{7.4} & \underline{22.0} & \underline{66.4} & 77.1 & 80.0 & 3.6 \\
    \textbf{mSAP} & \underline{54.5} & \underline{68.2} & \underline{20.1} & \underline{68.0} & 73.2 & \underline{7.1} & \underline{58.3} & \textbf{73.0} & \underline{20.1} & 5.6 & 19.6 & 71.4 & \underline{80.2} & \underline{82.8} & \underline{3.1} \\
    \rowcolor{appleteal!60}
    \textbf{\ours} & \textbf{55.9} & \textbf{69.0} & \textbf{19.0} & \textbf{68.5} & \textbf{73.7} & \underline{7.1} & \textbf{62.8} & \textbf{73.0} & \textbf{14.0} & \textbf{8.7} & \textbf{22.7} & \textbf{61.7} & \textbf{80.9} & \textbf{83.2} & \textbf{2.8} \\
    \bottomrule
\end{tabular}

    }
    \vspace{-13pt}
\end{table}

\vspace{-4pt}
\paragraph{Tasks with distractor queries.}
We finally evaluate on four tasks where a large share of queries are \emph{distractor queries}, \ie queries with no relevant item in the database: copy detection (DISC21~\citep{douze2021image}), instance-level recognition (Met~\citep{ypsilantis2021met}) and retrieval (ILIAS~\citep{kordopatis2025ilias}), and animal re-identification (WildlifeReID-10k~\citep{adam2024wildlifereid}). These are the settings the proposed loss targets, as they traditionally employ \uAP for evaluation. The system must decide with one threshold, shared across queries, whether a match exists at all, so a distractor query is handled correctly only if its negatives fall below the positives of other queries. Per-query metrics ignore distractor queries, since a query without positives has no AP, whereas \uAP considers all their negatives. Additional results and settings are in \Cref{sec:supp_distractors,sec:distractor_settings}. \ours attains the best \uAP on all four tasks and the smallest drop before score normalization. The gain is mostly cross-query. On Met, mSAP and \ours are identical without distractors and \ours is $4.5$ \uAP ahead with them. SN brings the drops of all losses close together but does not close the gap between mSAP and \ours on DISC21 for \uAP, so a post-hoc correction recovers only part of what training for \uAP provides. Finally, the second-best method varies across datasets and settings.
\vspace{-1pt}

\section{Conclusions}
\label{sec:conclusions}

\vspace{-1pt}
We revisit global Average Precision (\uAP), which evaluates ranking quality together with the consistency of similarities across queries, and propose \ours, a differentiable surrogate of it. \ours operates on the same input as the standard losses of established training recipes, a similarity matrix and its binary relevance labels, and swapping it in, and nothing else, improves supervised metric learning, cross-modal alignment, and self-supervised pretraining, surpassing the well-established InfoNCE where it is the first ranking loss to replace it. Its similarities are more consistent across queries, which pays off whenever one threshold serves all queries. Its cost is quadratic in the batch size, which negative dropping mitigates, and part of the achievable consistency remains unrealized, leaving room for complementary approaches, \eg learnable post-hoc score normalization~\citep{pizzi2022self}.

\bibliographystyle{iclr2027_conference}
\bibliography{bib}

@STRING{jan  = {January}}

@STRING{jun  = {June}}

@STRING{jan  = {Jan.}}

@STRING{jun  = {Jun.}}

@string{cvpr      = {Conference on Computer Vision and Pattern Recognition}}

@string{iccv      = {International Conference on Computer Vision}}

@string{eccv      = {European Conference on Computer Vision}}

@string{bmvc      = {British Machine Vision Conference}}

@string{neurips      = {Advances in Neural Information Processing Systems}}

@string{iclr      = {International Conference on Learning Representations}}

@string{acmmm     = {ACM International conference on Multimedia}}

@string{icassp    = {International Conference on Acoustics, Speech, and Signal Processing}}

@STRING{pami      = {{\sc IEEE} Transactions on Pattern Analysis and Machine Intelligence}}

@string{ijcv      = {International Journal of Computer Vision}}

@string{cviu      = {Computer Vision and Image Understanding}}

@STRING{tim       = {{\sc IEEE} Transactions on Image Processing}}

@string{cvpr      = {CVPR}}

@string{iccv      = {ICCV}}

@string{eccv      = {ECCV}}

@string{bmvc      = {BMVC}}

@string{neurips   = {NeurIPS}}

@string{iclr      = {ICLR}}

@string{icml      = {ICML}}

@string{acmmm     = {ACM Multimedia}}

@string{icassp    = {ICASSP}}

@string{aistats   = {AISTATS}}

@string{arxiv     = {arXiv}}

@string{wacv      = {WACV}}

@string{cvprw     = {CVPRW}}

@string{neuripsw  = {NeurIPSW}}

@STRING{pami      = {TPAMI}}

@string{ijcv      = {IJCV}}

@string{cviu      = {CVIU}}

@string{tmlr      = {TMLR}}

@article{ramzi2025optimization,
  title={Optimization of rank losses for image retrieval},
  author={Ramzi, Elias and Audebert, Nicolas and Rambour, Cl{\'e}ment and Araujo, Andr{\'e} and Bitot, Xavier and Thome, Nicolas},
  journal=pami,
  year={2025},
}

@inproceedings{revaud2019learning,
  title={Learning with average precision: Training image retrieval with a listwise loss},
  author={Revaud, Jerome and Almaz{\'a}n, Jon and Rezende, Rafael S and Souza, Cesar Roberto de},
  booktitle=iccv,
  year={2019}
}

@inproceedings{brown2020smooth,
  title={Smooth-ap: Smoothing the path towards large-scale image retrieval},
  author={Brown, Andrew and Xie, Weidi and Kalogeiton, Vicky and Zisserman, Andrew},
  booktitle=eccv,
  year={2020}
}

@inproceedings{patel2022recall,
title={{Recall@$k$} surrogate loss with large batches and similarity mixup},
  author={Patel, Yash and Tolias, Giorgos and Matas, Ji{\v{r}}{\'\i}},
  booktitle=cvpr,
  year={2022}
}

@inproceedings{zhang2023threshold,
  title={Threshold-consistent margin loss for open-world deep metric learning},
  author={Zhang, Qin and Xu, Linghan and Tang, Qingming and Fang, Jun and Wu, Ying Nian and Tighe, Joe and Xing, Yifan},
  booktitle=iclr,
  year={2024}
}

@inproceedings{ramzi2021robust,
  title={Robust and decomposable average precision for image retrieval},
  author={Ramzi, Elias and Thome, Nicolas and Rambour, Cl{\'e}ment and Audebert, Nicolas and Bitot, Xavier},
  booktitle=neurips,
  year={2021}
}

@inproceedings{ramzi2022hierarchical,
  title={Hierarchical average precision training for pertinent image retrieval},
  author={Ramzi, Elias and Audebert, Nicolas and Thome, Nicolas and Rambour, Cl{\'e}ment and Bitot, Xavier},
  booktitle=eccv,
  year={2022},
}

@inproceedings{he2016deep,
  title={Deep residual learning for image recognition},
  author={He, Kaiming and Zhang, Xiangyu and Ren, Shaoqing and Sun, Jian},
  booktitle=cvpr,
  year={2016}
}

@inproceedings{dosovitskiy2020image,
  title={An image is worth 16x16 words: Transformers for image recognition at scale},
  author={Dosovitskiy, Alexey and Beyer, Lucas and Kolesnikov, Alexander and Weissenborn, Dirk and Zhai, Xiaohua and Unterthiner, Thomas and Dehghani, Mostafa and Minderer, Matthias and Heigold, Georg and Gelly, Sylvain and others},
  booktitle=iclr,
  year={2021}
}

@inproceedings{zhai2023sigmoid,
  title={Sigmoid loss for language image pre-training},
  author={Zhai, Xiaohua and Mustafa, Basil and Kolesnikov, Alexander and Beyer, Lucas},
  booktitle=iccv,
  year={2023}
}

@article{oquab2023dinov2,
title={{DINOv2}: Learning robust visual features without supervision},
  author={Oquab, Maxime and Darcet, Timoth{\'e}e and Moutakanni, Th{\'e}o and Vo, Huy and Szafraniec, Marc and Khalidov, Vasil and Fernandez, Pierre and Haziza, Daniel and Massa, Francisco and El-Nouby, Alaaeldin and others},
  journal=tmlr,
  year={2024}
}

@book{schutze2008introduction,
  title={Introduction to information retrieval},
  author={Sch{\"u}tze, Hinrich and Manning, Christopher D and Raghavan, Prabhakar},
  year={2008},
  publisher={Cambridge University Press Cambridge}
}

@article{zhu2004recall,
  title={Recall, precision and average precision},
  author={Zhu, Mu},
  journal={Department of Statistics and Actuarial Science, University of Waterloo, Waterloo},
  year={2004}
}

@inproceedings{oh2016deep,
  title={Deep metric learning via lifted structured feature embedding},
  author={Oh Song, Hyun and Xiang, Yu and Jegelka, Stefanie and Savarese, Silvio},
  booktitle=cvpr,
  year={2016}
}

@inproceedings{van2018inaturalist,
title={The {iNaturalist} species classification and detection dataset},
  author={Van Horn, Grant and Mac Aodha, Oisin and Song, Yang and Cui, Yin and Sun, Chen and Shepard, Alex and Adam, Hartwig and Perona, Pietro and Belongie, Serge},
  booktitle=cvpr,
  year={2018}
}

@inproceedings{krause20133d,
title={{3D} object representations for fine-grained categorization},
  author={Krause, Jonathan and Stark, Michael and Deng, Jia and Fei-Fei, Li},
  booktitle=cvprw,
  year={2013}
}

@inproceedings{perronnin2009family,
  title={A family of contextual measures of similarity between distributions with application to image retrieval},
  author={Perronnin, Florent and Liu, Yan and Renders, Jean-Michel},
  booktitle=cvpr,
  year={2009},
}

@inproceedings{sablayrolles2018spreading,
  title={Spreading vectors for similarity search},
  author={Sablayrolles, Alexandre and Douze, Matthijs and Schmid, Cordelia and J{\'e}gou, Herv{\'e}},
  booktitle=iclr,
  year={2019}
}

@inproceedings{zhang2025rankclip,
title={{RankCLIP}: Ranking-consistent language-image pretraining},
  author={Zhang, Yiming and Zhao, Zhuokai and Chen, Zhaorun and Feng, Zhili and Ding, Zenghui and Sun, Yining},
  booktitle=iccv,
  year={2025}
}

@inproceedings{radford2021learning,
  title={Learning transferable visual models from natural language supervision},
  author={Radford, Alec and Kim, Jong Wook and Hallacy, Chris and Ramesh, Aditya and Goh, Gabriel and Agarwal, Sandhini and Sastry, Girish and Askell, Amanda and Mishkin, Pamela and Clark, Jack and others},
  booktitle=icml,
  year={2021},
}

@inproceedings{tschannen2025siglip,
  title={{SigLIP 2}: Multilingual vision-language encoders with improved semantic understanding, localization, and dense features},
  author={Tschannen, Michael and Gritsenko, Alexey and Wang, Xiao and Naeem, Muhammad Ferjad and Alabdulmohsin, Ibrahim and Parthasarathy, Nikhil and Evans, Talfan and Beyer, Lucas and Xia, Ye and Mustafa, Basil and others},
  booktitle=arxiv,
  year={2025}
}

@inproceedings{changpinyo2021conceptual,
  title={{Conceptual 12M}: Pushing web-scale image-text pre-training to recognize long-tail visual concepts},
  author={Changpinyo, Soravit and Sharma, Piyush and Ding, Nan and Soricut, Radu},
  booktitle=cvpr,
  year={2021}
}

@misc{cherti_clip_bench,
  author       = {Cherti, Mehdi and
                  Beaumont, Romain},
  title        = {{CLIP} benchmark},
  year         = 2022
}

@inproceedings{chen2024make,
  title={How to make cross encoder a good teacher for efficient image-text retrieval?},
  author={Chen, Yuxin and Ma, Zongyang and Zhang, Ziqi and Qi, Zhongang and Yuan, Chunfeng and Li, Bing and Pu, Junfu and Shan, Ying and Qi, Xiaojuan and Hu, Weiming},
  booktitle=cvpr,
  year={2024}
}

@inproceedings{oord2018representation,
  title={Representation learning with contrastive predictive coding},
  author={Oord, Aaron van den and Li, Yazhe and Vinyals, Oriol},
  booktitle=arxiv,
  year={2018}
}

@inproceedings{chopra2005learning,
  title={Learning a similarity metric discriminatively, with application to face verification},
  author={Chopra, Sumit and Hadsell, Raia and LeCun, Yann},
  booktitle=cvpr,
  year={2005},
}

@inproceedings{simeoni2025dinov3,
  title={{DINOv3}},
  author={Sim{\'e}oni, Oriane and Vo, Huy V and Seitzer, Maximilian and Baldassarre, Federico and Oquab, Maxime and Jose, Cijo and Khalidov, Vasil and Szafraniec, Marc and Yi, Seungeun and Ramamonjisoa, Micha{\"e}l and others},
  booktitle=arxiv,
  year={2025}
}

@inproceedings{schroff2015facenet,
  title={{FaceNet}: A unified embedding for face recognition and clustering},
  author={Schroff, Florian and Kalenichenko, Dmitry and Philbin, James},
  booktitle=cvpr,
  year={2015}
}

@article{radenovic2018fine,
  title={Fine-tuning CNN image retrieval with no human annotation},
  author={Radenovi{\'c}, Filip and Tolias, Giorgos and Chum, Ond{\v{r}}ej},
  journal=pami,
  year={2018},
}

@inproceedings{weyand2020google,
  title={{Google Landmarks Dataset v2} -- A large-scale benchmark for instance-level recognition and retrieval},
  author={Weyand, Tobias and Araujo, Andre and Cao, Bingyi and Sim, Jack},
  booktitle=cvpr,
  year={2020}
}

@inproceedings{ypsilantis2021met,
  title={The {Met} dataset: Instance-level recognition for artworks},
  author={Ypsilantis, Nikolaos-Antonios and Garcia, Noa and Han, Guangxing and Ibrahimi, Sarah and Van Noord, Nanne and Tolias, Giorgos},
  booktitle=neurips,
  year={2021}
}

@article{pizzi20242023,
  title={The 2023 video similarity dataset and challenge},
  author={Pizzi, Ed and Kordopatis-Zilos, Giorgos and Patel, Hiral and Postelnicu, Gheorghe and Ravindra, Sugosh Nagavara and Gupta, Akshay and Papadopoulos, Symeon and Tolias, Giorgos and Douze, Matthijs},
  journal=cviu,
  year={2024},
}

@inproceedings{pizzi2022self,
  title={A self-supervised descriptor for image copy detection},
  author={Pizzi, Ed and Roy, Sreya Dutta and Ravindra, Sugosh Nagavara and Goyal, Priya and Douze, Matthijs},
  booktitle=cvpr,
  year={2022}
}

@inproceedings{rolinek2020optimizing,
  title={Optimizing rank-based metrics with blackbox differentiation},
  author={Rol{\'\i}nek, Michal and Musil, V{\'\i}t and Paulus, Anselm and Vlastelica, Marin and Michaelis, Claudio and Martius, Georg},
  booktitle=cvpr,
  year={2020}
}

@inproceedings{cakir2019deep,
  title={Deep metric learning to rank},
  author={Cakir, Fatih and He, Kun and Xia, Xide and Kulis, Brian and Sclaroff, Stan},
  booktitle=cvpr,
  year={2019}
}

@inproceedings{triantafillou2017few,
  title={Few-shot learning through an information retrieval lens},
  author={Triantafillou, Eleni and Zemel, Richard and Urtasun, Raquel},
  booktitle=neurips,
  year={2017}
}

@inproceedings{deng2019arcface,
  title={{ArcFace}: Additive angular margin loss for deep face recognition},
  author={Deng, Jiankang and Guo, Jia and Xue, Niannan and Zafeiriou, Stefanos},
  booktitle=cvpr,
  year={2019}
}

@inproceedings{li2014deepreid,
  title={{DeepReID}: Deep filter pairing neural network for person re-identification},
  author={Li, Wei and Zhao, Rui and Xiao, Tong and Wang, Xiaogang},
  booktitle=cvpr,
  year={2014}
}

@inproceedings{sivic2003video,
  title={Video {Google}: A text retrieval approach to object matching in videos},
  author={Sivic, Josef and Zisserman, Andrew},
  booktitle=iccv,
  year={2003},
}

@inproceedings{voorhees1999trec,
  title={The {TREC-8} Question Answering Track Evaluation.},
  author={Voorhees, Ellen M and Tice, Dawn M and others},
  booktitle={TREC},
  year={1999}
}

@inproceedings{kim2023hier,
  title={{HIER}: Metric learning beyond class labels via hierarchical regularization},
  author={Kim, Sungyeon and Jeong, Boseung and Kwak, Suha},
  booktitle=cvpr,
  year={2023}
}

@article{jegou2010product,
  title={Product quantization for nearest neighbor search},
  author={Jegou, Herve and Douze, Matthijs and Schmid, Cordelia},
  journal=pami,
  year={2010},
}

@article{jarvelin2002cumulated,
  title={Cumulated gain-based evaluation of {IR} techniques},
  author={J{\"a}rvelin, Kalervo and Kek{\"a}l{\"a}inen, Jaana},
  journal={TOIS},
  year={2002},
}

@inproceedings{akiba2019optuna,
  title={Optuna: A next-generation hyperparameter optimization framework},
  author={Akiba, Takuya and Sano, Shotaro and Yanase, Toshihiko and Ohta, Takeru and Koyama, Masanori},
  booktitle={SIGKDD},
  year={2019}
}

@inproceedings{veit2020improving,
  title={Improving calibration in deep metric learning with cross-example softmax},
  author={Veit, Andreas and Wilber, Kimberly},
  booktitle=arxiv,
  year={2020}
}

@inproceedings{liu2022oneface,
  title={{OneFace}: One threshold for all},
  author={Liu, Jiaheng and Yu, Zhipeng and Qin, Haoyu and Wu, Yichao and Liang, Ding and Zhao, Gangming and Xu, Ke},
  booktitle=eccv,
  year={2022},
}

@inproceedings{zhang2024learning,
  title={Learning for transductive threshold calibration in open-world recognition},
  author={Zhang, Qin and An, Dongsheng and Xiao, Tianjun and He, Tong and Tang, Qingming and Wu, Ying Nian and Tighe, Joseph and Xing, Yifan},
  booktitle=cvpr,
  year={2024}
}

@inproceedings{guo2017calibration,
  title={On calibration of modern neural networks},
  author={Guo, Chuan and Pleiss, Geoff and Sun, Yu and Weinberger, Kilian Q},
  booktitle=icml,
  year={2017},
}

@article{qin2010general,
  title={A general approximation framework for direct optimization of information retrieval measures},
  author={Qin, Tao and Liu, Tie-Yan and Li, Hang},
  journal={Information retrieval},
  year={2010},
}

@inproceedings{ermolov2022hyperbolic,
  title={Hyperbolic vision transformers: Combining improvements in metric learning},
  author={Ermolov, Aleksandr and Mirvakhabova, Leyla and Khrulkov, Valentin and Sebe, Nicu and Oseledets, Ivan},
  booktitle=cvpr,
  year={2022}
}

@inproceedings{movshovitz2017no,
  title={No fuss distance metric learning using proxies},
  author={Movshovitz-Attias, Yair and Toshev, Alexander and Leung, Thomas K and Ioffe, Sergey and Singh, Saurabh},
  booktitle=iccv,
  year={2017}
}

@inproceedings{qian2019softtriple,
  title={{SoftTriple} loss: Deep metric learning without triplet sampling},
  author={Qian, Qi and Shang, Lei and Sun, Baigui and Hu, Juhua and Li, Hao and Jin, Rong},
  booktitle=iccv,
  year={2019}
}

@inproceedings{liu2017sphereface,
  title={{SphereFace}: Deep hypersphere embedding for face recognition},
  author={Liu, Weiyang and Wen, Yandong and Yu, Zhiding and Li, Ming and Raj, Bhiksha and Song, Le},
  booktitle=cvpr,
  year={2017}
}

@inproceedings{sohn2016improved,
  title={Improved deep metric learning with multi-class {N}-pair loss objective},
  author={Sohn, Kihyuk},
  booktitle=neurips,
  year={2016}
}

@inproceedings{taylor2008softrank,
  title={{SoftRank}: Optimizing non-smooth rank metrics},
  author={Taylor, Michael and Guiver, John and Robertson, Stephen and Minka, Tom},
  booktitle={WSDM},
  year={2008}
}

@inproceedings{burges2005learning,
  title={Learning to rank using gradient descent},
  author={Burges, Chris and Shaked, Tal and Renshaw, Erin and Lazier, Ari and Deeds, Matt and Hamilton, Nicole and Hullender, Greg},
  booktitle=icml,
  year={2005}
}

@inproceedings{loshchilov2017decoupled,
  title={Decoupled weight decay regularization},
  author={Loshchilov, Ilya and Hutter, Frank},
  booktitle=iclr,
  year={2019}
}

@inproceedings{kim2020proxy,
  title={Proxy anchor loss for deep metric learning},
  author={Kim, Sungyeon and Kim, Dongwon and Cho, Minsu and Kwak, Suha},
  booktitle=cvpr,
  year={2020}
}

@inproceedings{zhai2018classification,
  title={Classification is a strong baseline for deep metric learning},
  author={Zhai, Andrew and Wu, Hao-Yu},
  booktitle=bmvc,
  year={2019}
}

@inproceedings{bhatnagar2025potential,
  title={Potential Field Based Deep Metric Learning},
  author={Bhatnagar, Shubhang and Ahuja, Narendra},
  booktitle=cvpr,
  year={2025}
}

@inproceedings{teh2020proxynca++,
  title={{ProxyNCA++}: Revisiting and revitalizing proxy neighborhood component analysis},
  author={Teh, Eu Wern and DeVries, Terrance and Taylor, Graham W},
  booktitle=eccv,
  year={2020},
}

@inproceedings{wu2017sampling,
  title={Sampling matters in deep embedding learning},
  author={Wu, Chao-Yuan and Manmatha, Raghavan and Smola, Alexander J and Krahenbuhl, Philipp},
  booktitle=iccv,
  year={2017}
}

@inproceedings{wang2019multi,
  title={Multi-similarity loss with general pair weighting for deep metric learning},
  author={Wang, Xun and Han, Xintong and Huang, Weilin and Dong, Dengke and Scott, Matthew R},
  booktitle=cvpr,
  year={2019}
}

@inproceedings{kordopatis2023self,
  title={Self-supervised video similarity learning},
  author={Kordopatis-Zilos, Giorgos and Tolias, Giorgos and Tzelepis, Christos and Kompatsiaris, Ioannis and Patras, Ioannis and Papadopoulos, Symeon},
  booktitle=cvprw,
  year={2023}
}

@inproceedings{paszke2019pytorch,
  title={Pytorch: An imperative style, high-performance deep learning library},
  author={Paszke, Adam and Gross, Sam and Massa, Francisco and Lerer, Adam and Bradbury, James and Chanan, Gregory and Killeen, Trevor and Lin, Zeming and Gimelshein, Natalia and Antiga, Luca and others},
  booktitle=neurips,
  year={2019}
}

@misc{ilharco_gabriel_2021_5143773,
  author       = {Ilharco, Gabriel and
                  Wortsman, Mitchell and
                  Wightman, Ross and
                  Gordon, Cade and
                  Carlini, Nicholas and
                  Taori, Rohan and
                  Dave, Achal and
                  Shankar, Vaishaal and
                  Namkoong, Hongseok and
                  Miller, John and
                  Hajishirzi, Hannaneh and
                  Farhadi, Ali and
                  Schmidt, Ludwig},
  title        = {{OpenCLIP}},
  year         = 2021,
}

@inproceedings{cherti2023reproducible,
  title={Reproducible scaling laws for contrastive language-image learning},
  author={Cherti, Mehdi and Beaumont, Romain and Wightman, Ross and Wortsman, Mitchell and Ilharco, Gabriel and Gordon, Cade and Schuhmann, Christoph and Schmidt, Ludwig and Jitsev, Jenia},
  booktitle=cvpr,
  year={2023}
}

@inproceedings{deng2009imagenet,
  title     = {{ImageNet}: A Large-Scale Hierarchical Image Database},
  author    = {Deng, Jia and Dong, Wei and Socher, Richard and Li, Li-Jia and Li, Kai and Fei-Fei, Li},
  booktitle = cvpr,
  year      = {2009}
}

@inproceedings{recht2019imagenetv2,
  title     = {Do {ImageNet} Classifiers Generalize to {ImageNet}?},
  author    = {Recht, Benjamin and Roelofs, Rebecca and Schmidt, Ludwig and Shankar, Vaishaal},
  booktitle = icml,
  year      = {2019}
}

@inproceedings{wang2019learning,
  title     = {Learning Robust Global Representations by Penalizing Local Predictive Power},
  author    = {Wang, Haohan and Ge, Songwei and Lipton, Zachary C. and Xing, Eric P.},
  booktitle = neurips,
  year      = {2019}
}

@inproceedings{hendrycks2021natural,
  title     = {Natural Adversarial Examples},
  author    = {Hendrycks, Dan and Zhao, Kevin and Basart, Steven and Steinhardt, Jacob and Song, Dawn},
  booktitle = cvpr,
  year      = {2021}
}

@inproceedings{hendrycks2021many,
  title     = {The Many Faces of Robustness: A Critical Analysis of Out-of-Distribution Generalization},
  author    = {Hendrycks, Dan and Basart, Steven and Mu, Norman and Kadavath, Saurav and Wang, Frank and Dorundo, Evan and Desai, Rahul and Zhu, Tyler and Parajuli, Samyak and Guo, Mike and Song, Dawn and Steinhardt, Jacob and Gilmer, Justin},
  booktitle = iccv,
  year      = {2021}
}

@inproceedings{fei2004learning,
  title     = {Learning Generative Visual Models from Few Training Examples: An Incremental {B}ayesian Approach Tested on 101 Object Categories},
  author    = {Fei-Fei, Li and Fergus, Rob and Perona, Pietro},
  booktitle = cvprw,
  year      = {2004}
}

@misc{krizhevsky2009learning,
  title        = {Learning Multiple Layers of Features from Tiny Images},
  author       = {Krizhevsky, Alex},
  howpublished = {Technical Report, University of Toronto},
  year         = {2009}
}

@inproceedings{johnson2017clevr,
  title     = {{CLEVR}: A Diagnostic Dataset for Compositional Language and Elementary Visual Reasoning},
  author    = {Johnson, Justin and Hariharan, Bharath and van der Maaten, Laurens and Fei-Fei, Li and Zitnick, C. Lawrence and Girshick, Ross},
  booktitle = cvpr,
  year      = {2017}
}

@misc{country211_openai,
  title        = {The {Country211} Dataset},
  author       = {{OpenAI}},
  year         = {2021}
}

@misc{kaggle2015dr,
  title        = {Diabetic Retinopathy Detection},
  author       = {{Kaggle} and {EyePACS}},
  year         = {2015}
}

@inproceedings{beattie2016deepmind,
  title     = {{DeepMind Lab}},
  author    = {Beattie, Charles and Leibo, Joel Z. and Teplyashin, Denis and Ward, Tom and Wainwright, Marcus and K{\"u}ttler, Heinrich and Lefrancq, Andrew and Green, Simon and Vald{\'e}s, Victor and Sadik, Amir and Schrittwieser, Julian and Anderson, Keith and York, Sarah and Cant, Michael and Cain, Adam and Bolton, Adrian and Gaffney, Stephen and King, Helen and Hassabis, Demis and Legg, Shane and Petersen, Stig},
  booktitle = {arXiv},
  year      = {2016}
}

@misc{matthey2017dsprites,
  title        = {{dSprites}: Disentanglement Testing Sprites Dataset},
  author       = {Matthey, Loic and Higgins, Irina and Hassabis, Demis and Lerchner, Alexander},
  year         = {2017}
}

@inproceedings{cimpoi2014describing,
  title     = {Describing Textures in the Wild},
  author    = {Cimpoi, Mircea and Maji, Subhransu and Kokkinos, Iasonas and Mohamed, Sammy and Vedaldi, Andrea},
  booktitle = cvpr,
  year      = {2014}
}

@article{helber2019eurosat,
  title   = {{EuroSAT}: A Novel Dataset and Deep Learning Benchmark for Land Use and Land Cover Classification},
  author  = {Helber, Patrick and Bischke, Benjamin and Dengel, Andreas and Borth, Damian},
  journal = {J-STARS},
  year    = {2019}
}

@inproceedings{goodfellow2013challenges,
  title     = {Challenges in Representation Learning: A Report on Three Machine Learning Contests},
  author    = {Goodfellow, Ian J. and Erhan, Dumitru and Carrier, Pierre Luc and Courville, Aaron and Mirza, Mehdi and Hamner, Ben and Cukierski, Will and Tang, Yichuan and Thaler, David and Lee, Dong-Hyun and Zhou, Yingbo and Ramaiah, Chetan and Feng, Fangxiang and Li, Ruifan and Wang, Xiaojie and Athanasakis, Dimitris and Shawe-Taylor, John and Milakov, Maxim and Park, John and Ionescu, Radu and Popescu, Marius and Grozea, Cristian and Bergstra, James and Xie, Jingjing and Romaszko, Lukasz and Xu, Bing and Chuang, Zhang and Bengio, Yoshua},
  booktitle = {ICONIP},
  year      = {2013}
}

@inproceedings{maji2013fine,
  title     = {Fine-Grained Visual Classification of Aircraft},
  author    = {Maji, Subhransu and Rahtu, Esa and Kannala, Juho and Blaschko, Matthew and Vedaldi, Andrea},
  booktitle = {arXiv},
  year      = {2013}
}

@inproceedings{nilsback2008automated,
  title     = {Automated Flower Classification over a Large Number of Classes},
  author    = {Nilsback, Maria-Elena and Zisserman, Andrew},
  booktitle = {ICVGIP},
  year      = {2008}
}

@inproceedings{stallkamp2011gtsrb,
  title     = {The German Traffic Sign Recognition Benchmark: A Multi-Class Classification Competition},
  author    = {Stallkamp, Johannes and Schlipsing, Marc and Salmen, Jan and Igel, Christian},
  booktitle = {IJCNN},
  year      = {2011}
}

@inproceedings{netzer2011reading,
  title   = {Reading Digits in Natural Images with Unsupervised Feature Learning},
  author  = {Netzer, Yuval and Wang, Tao and Coates, Adam and Bissacco, Alessandro and Wu, Bo and Ng, Andrew Y.},
  booktitle = neuripsw,
  year    = {2011}
}

@article{everingham2010pascal,
  title   = {The Pascal Visual Object Classes ({VOC}) Challenge},
  author  = {Everingham, Mark and Van Gool, Luc and Williams, Christopher K. I. and Winn, John and Zisserman, Andrew},
  journal = ijcv,
  year    = {2010}
}

@inproceedings{barbu2019objectnet,
  title     = {{ObjectNet}: A Large-Scale Bias-Controlled Dataset for Pushing the Limits of Object Recognition Models},
  author    = {Barbu, Andrei and Mayo, David and Alverio, Julian and Luo, William and Wang, Christopher and Gutfreund, Dan and Tenenbaum, Joshua and Katz, Boris},
  booktitle = neurips,
  year      = {2019}
}

@article{cheng2017remote,
  title   = {Remote Sensing Image Scene Classification: Benchmark and State of the Art},
  author  = {Cheng, Gong and Han, Junwei and Lu, Xiaoqiang},
  journal = {Proceedings of the IEEE},
  year    = {2017}
}

@inproceedings{veeling2018rotation,
  title     = {Rotation Equivariant {CNNs} for Digital Pathology},
  author    = {Veeling, Bastiaan S. and Linmans, Jasper and Winkens, Jim and Cohen, Taco and Welling, Max},
  booktitle = {MICCAI},
  year      = {2018}
}

@inproceedings{parkhi2012cats,
  title     = {Cats and Dogs},
  author    = {Parkhi, Omkar M. and Vedaldi, Andrea and Zisserman, Andrew and Jawahar, C. V.},
  booktitle = cvpr,
  year      = {2012}
}

@inproceedings{geiger2012we,
  title     = {Are We Ready for Autonomous Driving? The {KITTI} Vision Benchmark Suite},
  author    = {Geiger, Andreas and Lenz, Philip and Urtasun, Raquel},
  booktitle = cvpr,
  year      = {2012}
}

@article{lecun1998gradient,
  title   = {Gradient-Based Learning Applied to Document Recognition},
  author  = {LeCun, Yann and Bottou, L{\'e}on and Bengio, Yoshua and Haffner, Patrick},
  journal = {Proceedings of the IEEE},
  year    = {1998}
}

@inproceedings{socher2013recursive,
  title     = {Recursive Deep Models for Semantic Compositionality over a Sentiment Treebank},
  author    = {Socher, Richard and Perelygin, Alex and Wu, Jean Y. and Chuang, Jason and Manning, Christopher D. and Ng, Andrew Y. and Potts, Christopher},
  booktitle = {EMNLP},
  year      = {2013}
}

@inproceedings{lecun2004learning,
  title     = {Learning Methods for Generic Object Recognition with Invariance to Pose and Lighting},
  author    = {LeCun, Yann and Huang, Fu Jie and Bottou, L{\'e}on},
  booktitle = cvpr,
  year      = {2004}
}

@inproceedings{coates2011analysis,
  title     = {An Analysis of Single-Layer Networks in Unsupervised Feature Learning},
  author    = {Coates, Adam and Ng, Andrew Y. and Lee, Honglak},
  booktitle = {AISTATS},
  year      = {2011}
}

@inproceedings{xiao2010sun,
  title     = {{SUN} Database: Large-Scale Scene Recognition from Abbey to Zoo},
  author    = {Xiao, Jianxiong and Hays, James and Ehinger, Krista A. and Oliva, Aude and Torralba, Antonio},
  booktitle = cvpr,
  year      = {2010}
}

@InProceedings{conconchi,
    author    = {Rosasco, Andrea and Berti, Stefano and Pasquale, Giulia and Malafronte, Damiano and Sato, Shogo and Segawa, Hiroyuki and Inada, Tetsugo and Natale, Lorenzo},
    title     = {ConCon-Chi: Concept-Context Chimera Benchmark for Personalized Vision-Language Tasks},
    booktitle = cvpr,
    year      = {2024},
}

@article{flickr,
  title={Framing image description as a ranking task: Data, models and evaluation metrics},
  author={Hodosh, Micah and Young, Peter and Hockenmaier, Julia},
  journal={JAIR},
  year={2013}
}

@article{flickr30k,
  title={From image descriptions to visual denotations: New similarity metrics for semantic inference over event descriptions},
  author={Young, Peter and Lai, Alice and Hodosh, Micah and Hockenmaier, Julia},
  journal={TACL},
  year={2014}
}

@inproceedings{cococaptions,
  title={Microsoft {COCO} captions: Data collection and evaluation server},
  author={Chen, Xinlei and Fang, Hao and Lin, Tsung-Yi and Vedantam, Ramakrishna and Gupta, Saurabh and Doll{\'a}r, Piotr and Zitnick, C Lawrence},
  booktitle=arxiv,
  year={2015}
}

@inproceedings{activitynet,
  title={{ActivityNet}: A large-scale video benchmark for human activity understanding},
  author={Caba Heilbron, Fabian and Escorcia, Victor and Ghanem, Bernard and Carlos Niebles, Juan},
  booktitle=cvpr,
  year={2015}
}

@inproceedings{msrvtt,
  title={{MSR-VTT}: A large video description dataset for bridging video and language},
  author={Xu, Jun and Mei, Tao and Yao, Ting and Rui, Yong},
  booktitle=cvpr,
  year={2016}
}

@inproceedings{msvd,
  title={Collecting highly parallel data for paraphrase evaluation},
  author={Chen, David and Dolan, William B},
  booktitle={ACL},
  year={2011}
}

@inproceedings{kay2017kinetics,
  title={The {Kinetics} human action video dataset},
  author={Kay, Will and Carreira, Joao and Simonyan, Karen and Zhang, Brian and Hillier, Chloe and Vijayanarasimhan, Sudheendra and Viola, Fabio and Green, Tim and Back, Trevor and Natsev, Paul and others},
  booktitle=arxiv,
  year={2017}
}

@inproceedings{carreira2018short,
  title={A short note about {Kinetics-600}},
  author={Carreira, Joao and Noland, Eric and Banki-Horvath, Andras and Hillier, Chloe and Zisserman, Andrew},
  booktitle=arxiv,
  year={2018}
}

@inproceedings{carreira2019short,
  title={A short note on the {Kinetics-700} human action dataset},
  author={Carreira, Joao and Noland, Eric and Hillier, Chloe and Zisserman, Andrew},
  booktitle=arxiv,
  year={2019}
}

@inproceedings{wu2023large,
  title={Large-scale contrastive language-audio pretraining with feature fusion and keyword-to-caption augmentation},
  author={Wu, Yusong and Chen, Ke and Zhang, Tianyu and Hui, Yuchen and Berg-Kirkpatrick, Taylor and Dubnov, Shlomo},
  booktitle=icassp,
  year={2023},
}

@inproceedings{kim2019audiocaps,
    title = "{A}udio{C}aps: Generating Captions for Audios in The Wild",
    author = "Kim, Chris Dongjoo  and
      Kim, Byeongchang  and
      Lee, Hyunmin  and
      Kim, Gunhee",
    booktitle = "NAACL",
    year = "2019",
}

@inproceedings{gemmeke2017audioset,
title     = {Audio Set: An ontology and human-labeled dataset for audio events},
author    = {Gemmeke, Jort F. and Ellis, Daniel P. W. and Freedman, Dylan and Jansen, Aren and Lawrence, Wade and Moore, R. Channing and Plakal, Manoj and Ritter, Marvin},
booktitle = icassp,
year      = {2017},
}

@inproceedings{chen2022htsat,
title     = {{HTS-AT}: A hierarchical token-semantic audio transformer for sound classification and detection},
author    = {Chen, Ke and Du, Xingjian and Zhu, Bilei and Ma, Zejun and Berg-Kirkpatrick, Taylor and Dubnov, Shlomo},
booktitle = icassp,
year      = {2022},
}

@inproceedings{drossos2020clotho,
  title     = {Clotho: An Audio Captioning Dataset},
  author    = {Drossos, Konstantinos and Lipping, Samuel and Virtanen, Tuomas},
  booktitle = icassp,
  year      = {2020},
}

@inproceedings{piczak2015esc,
  title     = {{ESC}: Dataset for Environmental Sound Classification},
  author    = {Piczak, Karol J.},
  booktitle = acmmm,
  year      = {2015},
}

@inproceedings{salamon2014urban,
  title     = {A Dataset and Taxonomy for Urban Sound Research},
  author    = {Salamon, Justin and Jacoby, Christopher and Bello, Juan Pablo},
  booktitle = acmmm,
  year      = {2014},
}

@inproceedings{chen2019towards,
  title={Towards Accurate One-Stage Object Detection with {AP}-Loss},
  author={Chen, Kean and Li, Jianguo and Lin, Weiyao and See, John and Wang, Ji and Duan, Lingyu and Chen, Zhibo and He, Changwei and Zou, Junni},
  booktitle=cvpr,
  year={2019}
}

@inproceedings{chen2021empirical,
  title     = {An Empirical Study of Training Self-Supervised Vision Transformers},
  author    = {Chen, Xinlei and Xie, Saining and He, Kaiming},
  booktitle = {ICCV},
  year      = {2021}
}

@inproceedings{venkataramanan2022takes,
  title={It Takes Two to Tango: Mixup for Deep Metric Learning},
  author={Venkataramanan, Shashanka and Psomas, Bill and Kijak, Ewa and Amsaleg, Laurent and Karantzalos, Konstantinos and Avrithis, Yannis},
  booktitle=iclr,
  year={2022}
}

@inproceedings{kakogeorgiou2022hide,
  title={What to Hide from Your Students: Attention-Guided Masked Image Modeling},
  author={Kakogeorgiou, Ioannis and Gidaris, Spyros and Psomas, Bill and Avrithis, Yannis and Bursuc, Andrei and Karantzalos, Konstantinos and Komodakis, Nikos},
  booktitle=eccv,
  year={2022}
}

@inproceedings{zhou2022ibot,
  title={{iBOT}: Image {BERT} Pre-Training with Online Tokenizer},
  author={Zhou, Jinghao and Wei, Chen and Wang, Huiyu and Shen, Wei and Xie, Cihang and Yuille, Alan and Kong, Tao},
  booktitle=iclr,
  year={2022}
}

@inproceedings{he2022masked,
  title={Masked Autoencoders Are Scalable Vision Learners},
  author={He, Kaiming and Chen, Xinlei and Xie, Saining and Li, Yanghao and Doll{\'a}r, Piotr and Girshick, Ross},
  booktitle=cvpr,
  year={2022}
}

@inproceedings{schuhmann2021laion,
  title   = {{LAION-400M}: Open Dataset of {CLIP}-Filtered 400 Million Image-Text Pairs},
  author  = {Schuhmann, Christoph and Vencu, Richard and Beaumont, Romain and Kaczmarczyk, Robert and Mullis, Clayton and Katta, Aarush and Coombes, Theo and Jitsev, Jenia and Komatsuzaki, Aran},
  booktitle=arxiv,
  year    = {2021}
}

@inproceedings{bao2022beit,
  title={{BEiT}: {BERT} Pre-Training of Image Transformers},
  author={Bao, Hangbo and Dong, Li and Piao, Songhao and Wei, Furu},
  booktitle=iclr,
  year={2022}
}

@inproceedings{chen2020simple,
  title={A Simple Framework for Contrastive Learning of Visual Representations},
  author={Chen, Ting and Kornblith, Simon and Norouzi, Mohammad and Hinton, Geoffrey},
  booktitle=icml,
  year={2020}
}

@inproceedings{he2020momentum,
  title={Momentum Contrast for Unsupervised Visual Representation Learning},
  author={He, Kaiming and Fan, Haoqi and Wu, Yuxin and Xie, Saining and Girshick, Ross},
  booktitle=cvpr,
  year={2020}
}

@inproceedings{grill2020bootstrap,
  title={Bootstrap Your Own Latent: A New Approach to Self-Supervised Learning},
  author={Grill, Jean-Bastien and Strub, Florian and Altch{\'e}, Florent and Tallec, Corentin and Richemond, Pierre H. and Buchatskaya, Elena and Doersch, Carl and Pires, Bernardo Avila and Guo, Zhaohan Daniel and Azar, Mohammad Gheshlaghi and Piot, Bilal and Kavukcuoglu, Koray and Munos, R{\'e}mi and Valko, Michal},
  booktitle=neurips,
  year={2020}
}

@inproceedings{caron2021emerging,
  title={Emerging Properties in Self-Supervised Vision Transformers},
  author={Caron, Mathilde and Touvron, Hugo and Misra, Ishan and J{\'e}gou, Herv{\'e} and Mairal, Julien and Bojanowski, Piotr and Joulin, Armand},
  booktitle=iccv,
  year={2021}
}

@inproceedings{hendrycks2019benchmarking,
  title     = {Benchmarking Neural Network Robustness to Common Corruptions and Perturbations},
  author    = {Hendrycks, Dan and Dietterich, Thomas},
  booktitle = {ICLR},
  year      = {2019}
}

@techreport{wah2011caltech,
  title       = {The {Caltech-UCSD} Birds-200-2011 Dataset},
  author      = {Wah, Catherine and Branson, Steve and Welinder, Peter and Perona, Pietro and Belongie, Serge},
  institution = {California Institute of Technology},
  number      = {CNS-TR-2011-001},
  year        = {2011}
}

@inproceedings{bossard2014food,
  title     = {Food-101 -- Mining Discriminative Components with Random Forests},
  author    = {Bossard, Lukas and Guillaumin, Matthieu and Van Gool, Luc},
  booktitle = eccv,
  year      = {2014}
}

@inproceedings{law2009evaluation,
  title={Evaluation of algorithms using games: The case of music tagging},
  author={Law, Edith and West, Kris and Mandel, Michael I and Bay, Mert and Downie, J Stephen},
  booktitle={ISMIR},
  year={2009}
}

@inproceedings{douze2021image,
  title={The 2021 image similarity dataset and challenge},
  author={Douze, Matthijs and Tolias, Giorgos and Pizzi, Ed and Papakipos, Zo{\"e} and Chanussot, Lowik and Radenovic, Filip and Jenicek, Tomas and Maximov, Maxim and Leal-Taix{\'e}, Laura and Elezi, Ismail and others},
  booktitle=arxiv,
  year={2021}
}

@inproceedings{kordopatis2025ilias,
  title={{ILIAS}: Instance-level image retrieval at scale},
  author={Kordopatis-Zilos, Giorgos and Stojni{\'c}, Vladan and Manko, Anna and {\v{S}}uma, Pavel and Ypsilantis, Nikolaos-Antonios and Efthymiadis, Nikos and Laskar, Zakaria and Matas, Ji{\v{r}}{\'i} and Chum, Ond{\v{r}}ej and Tolias, Giorgos},
  booktitle=cvpr,
  year={2025}
}

@inproceedings{liu2019roberta,
  title   = {{RoBERTa}: A Robustly Optimized {BERT} Pretraining Approach},
  author  = {Liu, Yinhan and Ott, Myle and Goyal, Naman and Du, Jingfei and Joshi, Mandar and Chen, Danqi and Levy, Omer and Lewis, Mike and Zettlemoyer, Luke and Stoyanov, Veselin},
  booktitle = arXiv,
  year    = {2019}
}

@inproceedings{adam2024wildlifereid,
  title={{WildlifeReID-10k}: Wildlife re-identification dataset with 10k individual animals},
  author={Adam, Lukas and Cermak, Vojtech and Papafitsoros, Kostas and Picek, Lukas},
  booktitle=cvprw,
  year={2025}
}

@article{sidney1975decomposition,
  title={Decomposition algorithms for single-machine sequencing with precedence relations and deferral costs},
  author={Sidney, Jeffrey B},
  journal={Operations Research},
  year={1975}
}

@article{fawcett2007pav,
  title={{PAV} and the {ROC} convex hull},
  author={Fawcett, Tom and Niculescu-Mizil, Alexandru},
  journal={Machine Learning},
  year={2007}
}

@article{provost2001robust,
  title={Robust classification for imprecise environments},
  author={Provost, Foster and Fawcett, Tom},
  journal={Machine Learning},
  year={2001}
}

@article{robertson1977probability,
  title={The probability ranking principle in {IR}},
  author={Robertson, Stephen E},
  journal={Journal of Documentation},
  year={1977}
}

@article{auckenthaler2000score,
  title={Score normalization for text-independent speaker verification systems},
  author={Auckenthaler, Roland and Carey, Michael and Lloyd-Thomas, Harvey},
  journal={Digital Signal Processing},
  year={2000}
}

@inproceedings{jegou2007contextual,
  title={A contextual dissimilarity measure for accurate and efficient image search},
  author={J{\'e}gou, Herv{\'e} and Harzallah, Hedi and Schmid, Cordelia},
  booktitle=cvpr,
  year={2007}
}

@inproceedings{conneau2017word,
  title={Word translation without parallel data},
  author={Conneau, Alexis and Lample, Guillaume and Ranzato, Marc'Aurelio and Denoyer, Ludovic and J{\'e}gou, Herv{\'e}},
  booktitle=iclr,
  year={2017}
}

@article{liu2023globalap,
title = {GlobalAP: Global average precision optimization for person re-identification},
journal = {Pattern Recognition},
author = {Yifei Liu and Yaling Liang and Pengfei Wang and Ziheng Chen and Changxing Ding},
year = {2023},
}

@inproceedings{spijkervet2021contrastive,
  title={Contrastive Learning of Musical Representations},
  author={Spijkervet, Janne and Burgoyne, John Ashley},
  booktitle={ISMIR},
  year={2021}
}

@inproceedings{lee2017sample,
  title={Sample-level Deep Convolutional Neural Networks for Music Auto-tagging Using Raw Waveforms},
  author={Lee, Jongpil and Park, Jiyoung and Kim, Keunhyoung Luke and Nam, Juhan},
  booktitle={SMC},
  year={2017}
}

@inproceedings{bolya2025perception,
  title={Perception Encoder: The best visual embeddings are not at the output of the network},
  author={Bolya, Daniel and Huang, Po-Yao and Sun, Peize and Cho, Jang Hyun and Madotto, Andrea and Wei, Chen and Ma, Tengyu and Zhi, Jiale and Rajasegaran, Jathushan and Rasheed, Hanoona and others},
  booktitle=neurips,
  year={2025}
}

@inproceedings{cermak2024wildlifedatasets,
  title={{WildlifeDatasets}: An open-source toolkit for animal re-identification},
  author={{\v{C}}erm{\'a}k, Vojt{\v{e}}ch and Picek, Lukas and Adam, Luk{\'a}{\v{s}} and Papafitsoros, Kostas},
  booktitle=wacv,
  year={2024}
}

@inproceedings{otarashvili2024multispecies,
  title={Multispecies Animal Re-ID Using a Large Community-Curated Dataset},
  author={Otarashvili, Lasha and Subramanian, Tamilselvan and Holmberg, Jason and Levenson, J.J. and Stewart, Charles V.},
  booktitle=arxiv,
  year={2024}
}

@inproceedings{ypsilantis2023towards,
  title={Towards Universal Image Embeddings: A Large-Scale Dataset and Challenge for Generic Image Representations},
  author={Ypsilantis, Nikolaos-Antonios and Chen, Kaifeng and Cao, Bingyi and Lipovsk{\'y}, M{\'a}rio and Dogan-Sch{\"o}nberger, Pelin and Makosa, Grzegorz and Bluntschli, Boris and Seyedhosseini, Mojtaba and Chum, Ond{\v{r}}ej and Araujo, Andr{\'e}},
  booktitle=iccv,
  year={2023}
}

@inproceedings{kornblith2019better,
  title     = {Do Better {ImageNet} Models Transfer Better?},
  author    = {Kornblith, Simon and Shlens, Jonathon and Le, Quoc V.},
  booktitle = cvpr,
  year      = {2019}
}

@inproceedings{ericsson2021how,
  title     = {How Well Do Self-Supervised Models Transfer?},
  author    = {Ericsson, Linus and Gouk, Henry and Hospedales, Timothy M.},
  booktitle = cvpr,
  year      = {2021}
}

\clearpage
\appendix

\crefalias{section}{appendix}
\crefalias{subsection}{appendix}
\crefalias{subsubsection}{appendix}
\crefname{appendix}{Appendix}{Appendices}
\Crefname{appendix}{Appendix}{Appendices}

\setcounter{figure}{0}\renewcommand{\thefigure}{\Alph{figure}}
\setcounter{table}{0}\renewcommand{\thetable}{\Alph{table}}

\phantomsection
\addcontentsline{toc}{section}{Appendix}
\begin{center}
  {\Large\sc Appendix}
\end{center}

\startcontents[appendix]
{\hypersetup{linkcolor=black}\printcontents[appendix]{}{1}{\setcounter{tocdepth}{2}}}

\section{Extended related work}
\label{sec:supp_related}

\paragraph{Metric learning.} Early work employs pairwise and triplet losses, \eg, the contrastive loss~\citep{chopra2005learning} and the triplet loss~\citep{schroff2015facenet}, which pull positives together and push negatives beyond a margin. To incorporate more context, listwise~\citep{oh2016deep} and N-pair~\citep{sohn2016improved} losses sample multiple positives and negatives per anchor, and Multi-Similarity~\citep{wang2019multi} weights all pairs of the batch relative to each anchor. Proxy-based methods reduce complexity by learning class proxies~\citep{movshovitz2017no,teh2020proxynca++,kim2020proxy,zhai2018classification} or multiple proxies per class~\citep{qian2019softtriple}, while angular-margin losses~\citep{liu2017sphereface,deng2019arcface} enforce margin-based separation on the hypersphere. Hierarchical methods~\citep{ramzi2022hierarchical,ramzi2025optimization} incorporate error severity into the ranking, while other works train with hyperbolic losses~\citep{ermolov2022hyperbolic,kim2023hier} or potential fields~\citep{bhatnagar2025potential}. Orthogonal to the choice of loss, Metrix~\citep{venkataramanan2022takes} brings mixup to metric learning by interpolating both examples and target labels within a generalized formulation of these losses. All these losses sum terms defined per pair, per anchor, or per proxy, whereas \ours couples all pairs of the batch through a single ranking.

\paragraph{Vision--language alignment.}
Vision--language pretraining is commonly instantiated with CLIP-style contrastive learning~\citep{radford2021learning}, where image--text alignment is optimized via a softmax-normalized loss over in-batch negatives; the same loss underlies audio--language models such as CLAP~\citep{wu2023large}. Recent work revisits this design by introducing new losses. RankCLIP~\citep{zhang2025rankclip} extends pairwise contrastive learning to a listwise formulation and enforces both in-modal and cross-modal ordinal ranking consistency, aiming to capture many-to-many relationships between images and texts. SigLIP~\citep{zhai2023sigmoid,tschannen2025siglip} adopts a pairwise sigmoid loss that removes the need for row-wise normalization; its bias, shared by all pairs, sets one global threshold, but each pair is scored independently at that single operating point. Beyond pretraining, CPRD~\citep{chen2024make} distills cross-encoder knowledge into vision--language encoders by matching the partial order of hard negatives, demonstrating that relative ranking can be more transferable than raw score regression. In this work, we target \uAP directly, replacing these row-, column-, or pair-wise losses with a single AP over all image--text or audio--text pairs under a shared threshold.

\paragraph{Self-supervised learning.}
Contrastive self-supervised learning treats augmented views of the same sample as positives and the other samples of the batch as negatives, and optimizes InfoNCE per anchor in vision~\citep{chen2020simple,he2020momentum,chen2021empirical} and audio~\citep{spijkervet2021contrastive}. Non-contrastive methods avoid negatives altogether, either by self-distillation between a student and a momentum teacher~\citep{grill2020bootstrap,caron2021emerging} or by masked image modeling, which reconstructs masked patches in pixel~\citep{he2022masked} or token space~\citep{bao2022beit}. iBOT~\citep{zhou2022ibot} combines the two by distilling masked patch tokens from the teacher, and AttMask~\citep{kakogeorgiou2022hide} masks the patches the teacher attends to most, making the distillation task harder. DINOv2~\citep{oquab2023dinov2} and DINOv3~\citep{simeoni2025dinov3} scale this recipe into general-purpose backbones; DINOv3 is the starting point of our metric learning and distractor-query experiments. \ours applies to the contrastive family: since the views define the relevance matrix, it replaces InfoNCE with no other change, whereas distillation and reconstruction losses have no similarity matrix over negatives to rank.

\paragraph{Surrogate ranking losses.} To better align training with evaluation, recent work proposes differentiable surrogates for ranking metrics. Early approaches smooth NDCG~\citep{taylor2008softrank} or learn from pairwise preferences~\citep{burges2005learning}. Several methods approximate AP for optimizing mAP~\citep{triantafillou2017few,rolinek2020optimizing,revaud2019learning,cakir2019deep,brown2020smooth}. BlackBox~\citep{rolinek2020optimizing} estimates gradients through sorting via black-box differentiation, while FastAP~\citep{cakir2019deep} improves efficiency by quantizing distances into differentiable histograms, reducing the computational cost of AP gradients. Smooth-AP~\citep{brown2020smooth} further scales AP optimization by applying a sigmoid to similarity differences. ROADMAP~\citep{ramzi2021robust} replaces the sigmoid with a different function on positives and negatives, propagating only through the latter. Beyond AP, RS@$k$~\citep{patel2022recall} addresses the non-differentiability of Recall@$k$ with a smooth surrogate. GlobalAP~\citep{liu2023globalap}, despite its name, extends each query's ranking with a memory bank and still averages AP over queries. Surrogates used in representation learning, however, remain query-centric and do not target global criteria such as \uAP, which our loss is designed to optimize.

\paragraph{Similarity consistency and thresholding.} Prior work aims to make similarities comparable across queries, classes, or domains, so that one threshold serves all of them; the retrieval literature calls this property cross-query calibration~\citep{pizzi2022self}, which is distinct from the calibration of scores as match probabilities~\citep{guo2017calibration}. Consistency can be encouraged during training via modified objectives that incorporate cross-example similarities~\citep{veit2020improving}, regularizers that constrain pairwise distances within bounds~\citep{ramzi2021robust,zhang2023threshold,liu2022oneface}, or dispersion losses that push nearest negatives apart~\citep{sablayrolles2018spreading}. Beyond training time, similarity scores can be mapped post hoc to a common scale, \eg by per-query score normalization against a background set~\citep{pizzi2022self}, Platt scaling or isotonic regression~\citep{guo2017calibration}, or learned or transductive mapping modules~\citep{zhang2024learning}. In contrast, \ours needs no auxiliary regularizer or post-hoc mapping. By directly optimizing \uAP, it encourages similarities that are consistent across queries as a direct effect of the loss, and adding a consistency regularizer on top brings no further gain.

\paragraph{Evaluation metrics.}
The predominant evaluation is query-centric; mAP~\citep{zhu2004recall} measures the ranking of the relevant samples according to the query, and Recall@$k$~\citep{jegou2010product} measures the proportion of queries for which at least one positive is ranked in the first $k$ positions. Other metrics are classification-based, \eg precision, recall, F1-score~\citep{sivic2003video}, and retrieval-based, \eg normalized discounted cumulative gain (NDCG)~\citep{jarvelin2002cumulated}, and mean reciprocal rank (MRR)~\citep{voorhees1999trec}. However, these metrics emphasize per-query top-rank performance and are insensitive to dataset-level retrieval quality. \uAP, also reported as micro-AP ($\mu$AP) in landmark recognition and copy detection~\citep{weyand2020google,pizzi2022self}, instead aggregates all positive and negative pairs across queries to produce a single global precision--recall curve, effectively casting retrieval as a detection problem under a shared similarity threshold. We therefore adopt \uAP as a primary metric and derive a surrogate loss to optimize it directly.

\newpage

\section{Additional results}
\label{sec:supp_results}

\subsection{Cross-modal learning}
\label{sec:supp_vlm}

\begin{wraptable}{r}{0.49\linewidth}
    \vspace{-12pt}
    \centering
    \caption{\textbf{Zero-shot retrieval and classification on audio datasets.} Recall@1 on Clotho and top-1 accuracy on ESC-50 and UrbanSound8K, comparison of \ours with the CLIP (InfoNCE) and SigLIP losses. All models fine-tune the pretrained LAION-CLAP HTS-AT base on AudioCaps.}
    \label{tab:sota_clap}
    \vspace{-8pt}
    \newcolumntype{C}{>{\centering\arraybackslash}p{20pt}}
\scriptsize
\setlength{\tabcolsep}{2.0pt}
\begin{tabular}{lCCCCCC}
\toprule
\multirow{2}{*}{\textbf{method}} &
\multicolumn{2}{c}{\textbf{Clotho val}} &
\multicolumn{2}{c}{\textbf{Clotho eval}} &
\multicolumn{2}{c}{\textbf{classification}} \\
\cmidrule(lr){2-3}\cmidrule(lr){4-5}\cmidrule(lr){6-7}
& a$\to$t & t$\to$a & a$\to$t & t$\to$a & ESC & US8K \\
\midrule
\textbf{pretrained} &
14.3 & 13.9 & 13.1 & 12.4 & 90.5 & 82.3 \\
\midrule
\textbf{$+$ CLIP} &
16.3 & 15.5 & 15.2 & 15.5 & 90.9 & 83.3 \\
\textbf{$+$ SigLIP} &
16.0 & 15.2 & 15.0 & 14.4 & 90.0 & 83.3 \\
\rowcolor{appleteal!60}
\textbf{$+$ \ours} &
\textbf{16.7} & \textbf{15.9} & \textbf{15.7} & \textbf{16.7} & \textbf{91.7} & \textbf{83.7} \\
\bottomrule
\end{tabular}

    \vspace{-8pt}
\end{wraptable}

\paragraph{Audio--language.} We assess whether the benefits of optimizing \uAP extend beyond vision by repeating the protocol in the audio--language domain. We initialize LAION-CLAP~\citep{wu2023large} with an HTS-AT~\citep{chen2022htsat} base audio encoder that is pretrained on music, speech, and AudioSet~\citep{gemmeke2017audioset} and fine-tune for 10 epochs on AudioCaps~\citep{kim2019audiocaps} with the same three losses as in the main paper. \Cref{tab:sota_clap} reports zero-shot audio--text retrieval on the Clotho~\citep{drossos2020clotho} validation and evaluation splits, together with zero-shot classification on ESC-50~\citep{piczak2015esc} and UrbanSound8K~\citep{salamon2014urban}. \ours outperforms both CLIP and SigLIP on every column, with a clear margin in text-to-audio retrieval on the Clotho evaluation split, while SigLIP falls below the pretrained model on ESC-50.

\paragraph{Vision--language.} We extend the vision--language results of \Cref{sec:experiments} with the same fine-tuning protocol. We first analyze the effect of batch size and of negative dropping with OpenAI CLIP ViT-B/32, then repeat the zero-shot retrieval and classification evaluations starting from two other models, OpenAI CLIP ViT-B/16 and LAION-400M~\citep{schuhmann2021laion} CLIP ViT-B/32, and finally evaluate on an additional task, zero-shot composed image retrieval. \ours is the best of the three losses in every setting.

\begin{figure}[t]
    \centering
    \pgfplotstableread{
batch  SGAP_gap  VAN_gap  SIG_gap  SGAP_map_ti  SGAP_map_it  VAN_map_ti  VAN_map_it  SIG_map_ti  SIG_map_it
512    41.18     37.16    38.49    71.62        63.54        69.90       61.33       70.63       61.98
1024    43.09     39.57    41.62    72.63        64.74        70.65       63.09       72.14       63.34
1536    44.42     40.27    42.40    72.96        65.04        71.67       63.12       72.30       64.00
2048   44.48     41.90    42.43    73.02        65.68        71.77       63.57       72.60       64.22
}\flickrEightK

\pgfplotstableread{
batch  SGAP_gap  VAN_gap  SIG_gap  SGAP_map_ti  SGAP_map_it  VAN_map_ti  VAN_map_it  SIG_map_ti  SIG_map_it
512    56.72     51.56    54.78    74.58        68.22        72.67       65.17       74.24       66.99
1024    58.40     54.48    56.67    75.60        69.42        73.77       67.27       75.52       68.60
1536    59.30     55.68    57.33    76.32        69.83        74.67       68.01       75.85       68.99
2048   59.50     56.17    58.01    76.44        70.30        75.26       68.27       76.23       69.37
}\flickrThirtyK

\pgfplotstableread{
batch  SGAP_gap  VAN_gap  SIG_gap  SGAP_map_ti  SGAP_map_it  VAN_map_ti  VAN_map_it  SIG_map_ti  SIG_map_it
512    19.88     16.73    16.77    49.31        40.90        47.59       38.99       49.26       39.51
1024    21.19     18.08    18.57    50.56        42.29        48.52       40.30       50.02       40.62
1536    21.26     18.46    19.33    50.91        42.58        49.40       40.68       50.50       41.27
2048   21.37     18.51    19.23    51.11        42.91        49.60       41.02       50.63       41.74
}\mscocoCaps

\pgfplotsset{every tick label/.append style={font=\footnotesize}}

\pgfplotsset{
    vanilla/.style={color=clipblue!70, mark=*, mark size=2.0, opacity=1.0,
        mark options={solid, draw=black, line width=0.35pt}, line width=1.2pt},
    siglip/.style={color=siglipviolet!70, mark=*, mark size=2.0, opacity=1.0,
        mark options={solid, draw=black, line width=0.35pt}, line width=1.2pt},
    sgap/.style={color=appleteal, mark=*, mark size=2.0, opacity=1.0,
        mark options={solid, draw=black, line width=0.35pt}, line width=1.2pt},
}

\begin{tikzpicture}
  \begin{axis}[
    hide axis,
    scale only axis,
    height=0cm,
    width=0cm,
    legend style={
      draw=black,
      cells={anchor=west},
      legend columns=5,
      column sep=4pt,
      font=\footnotesize,
      anchor=north,
    }
  ]

    \addlegendimage{dotted, thick, color=appleorange, line width=1.2pt};
    \addlegendentry{pretrained}

    \addlegendimage{solid, thick, vanilla}
    \addlegendentry{CLIP (InfoNCE)}
    
    \addlegendimage{solid, thick, siglip}
    \addlegendentry{SigLIP}
    
    \addlegendimage{solid, thick, sgap}
    \addlegendentry{\ours}

    \addplot[draw=none] coordinates {(0,0)};
    
  \end{axis}
\end{tikzpicture}

\vspace{2pt}

\begin{tabular}{@{}c@{\hspace{8pt}}c@{\hspace{8pt}}c@{}}

\begin{tikzpicture}
  \begin{axis}[
    width=0.36\linewidth, height=0.30\linewidth,
    title={\textbf{Flickr8k}}, title style={yshift=-4pt},
    xlabel={batch size}, ylabel={\uAP},
    grid=both, grid style={color=lightgray!60, dash pattern=on 2pt off 2pt},
    xtick={512,1024,1536,2048}, xmin=496, xmax=2062,
    ymin=30, ymax=46, xmin=470, xmax=2090,
    xlabel style={yshift=3pt}, ylabel style={yshift=-4pt},
  ]
    \addplot[sgap]   table[x=batch,y=SGAP_gap] {\flickrEightK};
    \addplot[vanilla] table[x=batch,y=VAN_gap]  {\flickrEightK};
    \addplot[siglip]    table[x=batch,y=SIG_gap]  {\flickrEightK};

    \addplot[dotted, thick, color=appleorange, line width=1.2pt] coordinates {(1, 31.0) (2048, 31.0)};
  \end{axis}
\end{tikzpicture}
&
\begin{tikzpicture}
  \begin{axis}[
    width=0.36\linewidth, height=0.30\linewidth,
    title={\textbf{Flickr30k}}, title style={yshift=-4pt},
    xlabel={batch size}, ylabel={},
    grid=both, grid style={color=lightgray!60, dash pattern=on 2pt off 2pt},
    xtick={512,1024,1536,2048}, xmin=496, xmax=2062,
    ymin=44.5, ymax=61, xmin=470, xmax=2090,
    xlabel style={yshift=3pt},
  ]
    \addplot[sgap]   table[x=batch,y=SGAP_gap] {\flickrThirtyK};
    \addplot[vanilla] table[x=batch,y=VAN_gap]  {\flickrThirtyK};
    \addplot[siglip]    table[x=batch,y=SIG_gap]  {\flickrThirtyK};
    \addplot[dotted, thick, color=appleorange, line width=1.2pt] coordinates {(1, 45.30) (2048, 45.30)};
  \end{axis}
\end{tikzpicture}
&
\begin{tikzpicture}
  \begin{axis}[
    width=0.36\linewidth, height=0.30\linewidth,
    title={\textbf{MS COCO}}, title style={yshift=-4pt},
    xlabel={batch size}, ylabel={},
    grid=both, grid style={color=lightgray!60, dash pattern=on 2pt off 2pt},
    xtick={512,1024,1536,2048}, xmin=496, xmax=2062,
    ymin=12.5, ymax=23, xmin=470, xmax=2090,
    xlabel style={yshift=3pt},
  ]
    \addplot[sgap]   table[x=batch,y=SGAP_gap] {\mscocoCaps};
    \addplot[vanilla] table[x=batch,y=VAN_gap]  {\mscocoCaps};
    \addplot[siglip]    table[x=batch,y=SIG_gap]  {\mscocoCaps};
    \addplot[dotted, thick, color=appleorange, line width=1.2pt] coordinates {(1, 13.10) (2048, 13.10)};
  \end{axis}
\end{tikzpicture}

\end{tabular}
    \vspace{-6pt}
    \caption{\textbf{Effect of batch size on zero-shot retrieval \uAP.} \uAP comparison of \ours with InfoNCE (CLIP) and SigLIP losses for varying batch sizes. All models fine-tune the pretrained OpenAI CLIP ViT-B/32 for 5 epochs on CC12M.
    \label{fig:vlms_multiple_batch_sizes}
    \vspace{-8pt}
    }
\end{figure}
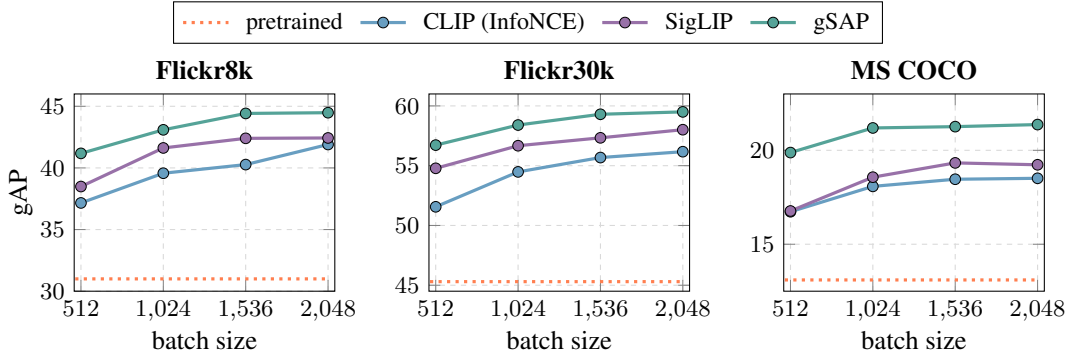

\noindent\textit{Batch size sensitivity.} We fine-tune OpenAI CLIP ViT-B/32 on CC12M for 5 epochs with varying batch sizes and compare InfoNCE and SigLIP with \ours on zero-shot image--text retrieval on Flickr8k, Flickr30k, and MS COCO (\Cref{fig:vlms_multiple_batch_sizes}). \ours achieves the highest \uAP on all three datasets at every batch size, with the largest margin at the smallest batch. At 512, it already exceeds InfoNCE trained with a four times larger batch on Flickr30k and MS COCO, so its gain does not rely on large batches.

\noindent\textit{Negative dropping.} \Cref{tab:sota_vlm_drop_bs} evaluates negative dropping on zero-shot retrieval on the three image datasets with OpenAI CLIP ViT-B/32. At the same batch size, \ours outperforms InfoNCE and SigLIP in every metric on all three datasets. Dropping $95\%$ of the negatives ($\dagger$) costs at most $0.9$ \uAP, and \ours still outperforms both losses in \uAP. The saved memory allows increasing the batch size from 2,304 to 4,096, which gives the best result in every metric. In contrast, training InfoNCE and SigLIP with the largest batch size that fits in GPU memory (3,072, $*$) improves InfoNCE, which still remains below \ours at 2,304, and even drops the \uAP of SigLIP.

\newpage

\begin{table}[t]
    \centering
    \caption{\textbf{Zero-shot retrieval with negative dropping on image datasets.} \uAP, mAP, and Recall@5 comparison of \ours with InfoNCE (CLIP) and SigLIP losses. All models fine-tune the pretrained OpenAI CLIP ViT-B/32 for 5 epochs on CC12M. $\dagger$: negative dropping with $\alpha=95\%$. $*$: largest batch size that fits in GPU memory.}
    \vspace{-8pt}
    \label{tab:sota_vlm_drop_bs}
    \scalebox{0.88}{
    \hspace{-10pt}
    \newcolumntype{C}{>{\centering\arraybackslash}p{21pt}}
\scriptsize
\setlength{\tabcolsep}{0.4pt}
\begin{tabular}{lcCCCCCCCCCCCCCCC}
\toprule
\multirow{3}{*}{\textbf{method}} &
\multirow{3}{*}{\textbf{batch}} &
\multicolumn{5}{c}{\textbf{Flickr8k}} &
\multicolumn{5}{c}{\textbf{Flickr30k}} &
\multicolumn{5}{c}{\textbf{MS COCO}} \\
\cmidrule(lr){3-7}\cmidrule(lr){8-12}\cmidrule(lr){13-17}
& 
& \multirow{2}{*}{\uAP} & mAP t$\rightarrow$i & mAP i$\rightarrow$t & R@5 t$\rightarrow$i & R@5 i$\rightarrow$t
& \multirow{2}{*}{\uAP} & mAP t$\rightarrow$i & mAP i$\rightarrow$t & R@5 t$\rightarrow$i & R@5 i$\rightarrow$t
& \multirow{2}{*}{\uAP} & mAP t$\rightarrow$i & mAP i$\rightarrow$t & R@5 t$\rightarrow$i & R@5 i$\rightarrow$t \\
\midrule
\textbf{pretrained} &
- &
31.0 & 65.0 & 57.7 & 79.2 & 92.0 &
45.3 & 68.7 & 62.7 & 82.9 & 94.0 &
13.1 & 42.1 & 36.7 & 55.0 & 73.4 \\
\midrule

\textbf{$+$ CLIP} &
2304 &
41.3 & 71.9 & 63.5 & 85.5 & 92.4 &
56.1 & 75.1 & 68.3 & 87.6 & 94.7 &
18.7 & 49.6 & 41.1 & 63.6 & 76.9 \\

\textbf{$+$ SigLIP} &
2304 &
43.2 & 72.5 & 64.6 & 86.0 & 92.7 &
58.5 & 76.3 & 69.5 & 88.5 & 95.7 &
19.4 & 50.6 & 41.8 & 64.7 & 77.5 \\

\textbf{$+$ \ours} &
2304 &
45.1 & 73.2 & 65.6 & 86.7 & 93.3 &
59.7 & 76.6 & 70.5 & 88.9 & 96.0 &
21.4 & 51.3 & 43.1 & 65.3 & 79.0 \\
\midrule

\textbf{$+$ CLIP$^{*}$} &
3072 &
42.3 & 72.1 & 63.9 & 85.8 & 92.7 &
57.0 & 75.6 & 68.7 & 88.1 & 95.0 &
19.9 & 50.0 & 41.7 & 64.0 & 77.7 \\

\textbf{$+$ SigLIP$^{*}$} &
3072 &
42.7 & 72.5 & 64.0 & 86.1 & 92.4 &
57.5 & 76.0 & 69.0 & 88.4 & 96.0 &
18.7 & 50.4 & 41.3 & 64.4 & 77.1 \\

\midrule

\textbf{$+$ $\text{gSAP}^\dagger$} &
2304 &
44.2 & 73.0 & 65.6 & 86.5 & 92.6 &
59.2 & 76.6 & 70.4 & 88.6 & 96.0 &
20.7 & 51.2 & 42.9 & 65.3 & 79.0 \\

\rowcolor{appleteal!60}
\textbf{$+$ $\text{gSAP}^{\dagger*}$} &
4096 &
\textbf{45.3} & \textbf{73.6} & \textbf{66.4} & \textbf{86.8} & \textbf{93.6} &
\textbf{59.9} & \textbf{76.9} & \textbf{71.0} & \textbf{89.1} & \textbf{96.5} &
\textbf{22.3} & \textbf{51.6} & \textbf{43.6} & \textbf{65.6} & \textbf{79.8} \\

\bottomrule
\end{tabular}

    }
\end{table}

\begin{table}[t]
    \centering
    \caption{\textbf{Zero-shot retrieval on image datasets with other backbones and pretraining data.} \uAP, mAP, and Recall@5 comparison of \ours with the CLIP (InfoNCE) and SigLIP losses. All models fine-tune the pretrained OpenAI CLIP ViT-B/16 or LAION-400M CLIP ViT-B/32 for 5 epochs on CC12M.}
    \label{tab:sota_vlm_image_retrieval_backbones}
    \vspace{-8pt}
    \scalebox{0.93}{
    \newcolumntype{C}{>{\centering\arraybackslash}p{21pt}}
\scriptsize
\setlength{\tabcolsep}{0.8pt}
\begin{tabular}{lCCCCCCCCCCCCCCC}
\toprule
\multirow{2}{*}{\textbf{method}} &
\multicolumn{5}{c}{\textbf{Flickr8k}} &
\multicolumn{5}{c}{\textbf{Flickr30k}} &
\multicolumn{5}{c}{\textbf{MS COCO}} \\
\cmidrule(lr){2-6}\cmidrule(lr){7-11}\cmidrule(lr){12-16}
& \multirow{2}{*}{\uAP} & mAP t$\rightarrow$i & mAP i$\rightarrow$t & R@5 t$\rightarrow$i & R@5 i$\rightarrow$t
& \multirow{2}{*}{\uAP} & mAP t$\rightarrow$i & mAP i$\rightarrow$t & R@5 t$\rightarrow$i & R@5 i$\rightarrow$t
& \multirow{2}{*}{\uAP} & mAP t$\rightarrow$i & mAP i$\rightarrow$t & R@5 t$\rightarrow$i & R@5 i$\rightarrow$t \\
\midrule
\multicolumn{16}{l}{\textit{OpenAI CLIP ViT-B/16}} \\
\textbf{pretrained} &
30.6 & 65.6 & 56.8 & 80.4 & 90.2 &
44.7 & 69.8 & 64.5 & 83.2 & 94.7 &
12.2 & 42.3 & 37.6 & 55.0 & 74.0 \\
\textbf{$+$ CLIP} &
42.7 & 75.3 & 66.5 & 87.9 & 93.8 &
61.0 & 79.4 & 72.9 & 90.7 & 96.7 &
21.7 & 52.9 & 44.0 & 66.4 & 79.8 \\
\textbf{$+$ SigLIP} &
46.4 & 76.2 & 67.7 & \textbf{88.7} & 94.4 &
64.0 & 80.7 & 74.5 & 91.5 & 97.4 &
22.7 & 53.9 & 44.9 & 67.4 & 80.0 \\
\rowcolor{appleteal!60}
\textbf{$+$ \ours} &
\textbf{47.3} & \textbf{76.6} & \textbf{68.8} & \textbf{88.7} & \textbf{94.5} &
\textbf{64.6} & \textbf{81.1} & \textbf{74.7} & \textbf{91.7} & \textbf{97.7} &
\textbf{24.6} & \textbf{54.7} & \textbf{46.0} & \textbf{68.0} & \textbf{81.2} \\
\midrule
\multicolumn{16}{l}{\textit{LAION-400M CLIP ViT-B/32}} \\
\textbf{pretrained} &
38.9 & 67.6 & 61.1 & 81.4 & 91.8 &
52.1 & 70.5 & 64.9 & 83.8 & 93.7 &
17.3 & 46.4 & 39.4 & 60.0 & 76.3 \\
\textbf{$+$ CLIP} &
38.5 & 68.0 & 60.1 & 82.7 & 91.3 &
53.7 & 72.6 & 65.4 & 85.4 & 93.6 &
16.6 & 47.7 & 39.1 & 61.3 & 75.0 \\
\textbf{$+$ SigLIP} &
39.9 & 68.9 & 61.4 & 83.3 & 90.8 &
54.5 & 73.2 & 66.5 & 85.9 & 93.8 &
17.2 & 48.5 & 39.9 & 62.1 & 76.1 \\
\rowcolor{appleteal!60}
\textbf{$+$ \ours} &
\textbf{41.5} & \textbf{69.7} & \textbf{62.6} & \textbf{84.0} & \textbf{92.2} &
\textbf{55.7} & \textbf{73.3} & \textbf{66.6} & \textbf{86.3} & \textbf{94.4} &
\textbf{18.5} & \textbf{49.0} & \textbf{40.9} & \textbf{62.6} & \textbf{77.2} \\
\bottomrule
\end{tabular}

    }
\end{table}

\begin{table}[h!]
    \centering
    \caption{\textbf{Zero-shot retrieval on video datasets with other backbones and pretraining data.} \uAP and Recall@$k$ comparison. Same setting as \Cref{tab:sota_vlm_image_retrieval_backbones}.}
    \vspace{-8pt}
    \label{tab:sota_vlm_video_retrieval_backbones}
    \resizebox{\linewidth}{!}{%
    \providecommand{\tbd}{\textcolor{red}{--}}
\newcolumntype{C}{>{\centering\arraybackslash}p{17pt}}
\scriptsize
\setlength{\tabcolsep}{0.8pt}
\begin{tabular}{lCCCCCCCCCCCCCCCCCCCCC}
\toprule
\multirow{3}{*}{\textbf{method}} &
\multicolumn{7}{c}{\textbf{ActivityNet}} &
\multicolumn{7}{c}{\textbf{MSR-VTT}} &
\multicolumn{7}{c}{\textbf{MSVD}} \\
\cmidrule(lr){2-8}\cmidrule(lr){9-15}\cmidrule(lr){16-22}
& \multirow{2}{*}{\uAP} & \multicolumn{3}{c}{text$\to$video} & \multicolumn{3}{c}{video$\to$text}
& \multirow{2}{*}{\uAP} & \multicolumn{3}{c}{text$\to$video} & \multicolumn{3}{c}{video$\to$text}
& \multirow{2}{*}{\uAP} & \multicolumn{3}{c}{text$\to$video} & \multicolumn{3}{c}{video$\to$text} \\
\cmidrule(lr){3-5}\cmidrule(lr){6-8}\cmidrule(lr){10-12}\cmidrule(lr){13-15}\cmidrule(lr){17-19}\cmidrule(lr){20-22}
& & R@1 & R@5 & R@10 & R@1 & R@5 & R@10
& & R@1 & R@5 & R@10 & R@1 & R@5 & R@10
& & R@1 & R@5 & R@10 & R@1 & R@5 & R@10 \\
\midrule
\multicolumn{22}{l}{\textit{OpenAI CLIP ViT-B/16}} \\
\textbf{pretrained} &
11.7 & 25.2 & 55.4 & 69.0 & 22.7 & 50.0 & 62.0 &
5.0 & 15.6 & 32.6 & 41.4 & 31.8 & 52.8 & 63.9 &
17.1 & 36.5 & 63.9 & 73.3 & 55.0 & 74.8 & 82.1 \\
\textbf{$+$ CLIP} &
14.7 & 29.9 & 63.6 & 76.9 & 29.2 & 61.4 & 73.7 &
4.5 & 18.2 & 36.9 & 45.9 & 30.2 & 52.4 & 62.2 &
26.3 & 42.5 & 70.9 & 79.8 & 54.9 & 76.3 & 83.6 \\
\textbf{$+$ SigLIP} &
15.3 & 31.3 & 65.2 & 77.6 & 29.9 & 60.8 & 74.0 &
4.6 & 18.7 & 37.5 & 46.6 & 30.9 & 53.6 & 63.5 &
26.8 & 43.2 & 71.2 & 80.0 & 55.0 & 77.1 & \textbf{85.2} \\
\rowcolor{appleteal!60}
\textbf{$+$ \ours} &
\textbf{15.6} & \textbf{31.9} & \textbf{66.2} & \textbf{77.7} & \textbf{30.9} & \textbf{61.8} & \textbf{74.1} &
\textbf{5.3} & \textbf{19.1} & \textbf{38.1} & \textbf{47.2} & \textbf{34.2} & \textbf{56.2} & \textbf{65.5} &
\textbf{27.5} & \textbf{43.5} & \textbf{71.7} & \textbf{80.4} & \textbf{56.7} & \textbf{78.3} & \textbf{85.2} \\
\midrule
\multicolumn{22}{l}{\textit{LAION-400M CLIP ViT-B/32}} \\
\textbf{pretrained} &
10.9 & 23.7 & 49.8 & 61.3 & 21.4 & 49.4 & 61.8 & 4.2 & 13.9 & 29.1 & 37.4 & \textbf{28.7} & \textbf{48.7} & \textbf{59.3} & 15.5 & 33.4 & 59.6 & 68.6 & 47.0 & 67.2 & 75.7 \\
\textbf{$+$ CLIP} &
10.3 & 24.2 & 54.1 & 68.2 & 21.7 & 51.5 & 64.7 & 3.1 & 14.6 & 31.0 & 40.1 & 22.5 & 43.1 & 53.1 & 19.1 & 36.3 & 65.8 & 75.1 & 47.5 & 69.4 & 78.7 \\
\textbf{$+$ SigLIP} &
10.5 & 24.0 & 54.2 & 68.1 & 22.1 & 51.7 & 65.0 & 3.4 & 14.9 & 31.7 & 40.6 & 23.8 & 45.8 & 55.8 & \textbf{19.7} & 37.3 & 66.2 & 75.7 & 49.0 & 69.8 & 78.1 \\
\rowcolor{appleteal!60}
\textbf{$+$ \ours} &
\textbf{11.4} & \textbf{25.1} & \textbf{55.2} & \textbf{68.5} & \textbf{23.2} & \textbf{51.9} & \textbf{65.1} & \textbf{4.8} & \textbf{15.4} & \textbf{32.2} & \textbf{41.2} & 26.3 & 47.9 & 57.5 & \textbf{19.7} & \textbf{37.6} & \textbf{66.7} & \textbf{76.0} & \textbf{50.1} & \textbf{72.2} & \textbf{80.0} \\
\bottomrule
\end{tabular}

    }
\end{table}

\begin{table}[h!]
\centering
\scriptsize
\caption{\textbf{Zero-shot composed image retrieval on ConCon-Chi.} \uAP, mAP, mAP@$k$, and Recall@$k$ comparison of \ours with InfoNCE (CLIP) and SigLIP losses. All models fine-tune the pretrained OpenAI CLIP ViT-B/32 (top) or ViT-B/16 (bottom) for 5 epochs on CC12M.}
  \vspace{-8pt}
\label{tab:sota_vlm_conconchi}
\begin{tabular}{l cc ccccc ccccc}
\toprule
\multirow{2}{*}{\textbf{method}} & & & \multicolumn{5}{c}{mAP@$k$} & \multicolumn{5}{c}{Recall@$k$} \\
\cmidrule(lr){4-8}\cmidrule(lr){9-13}
& gAP & mAP & @1 & @5 & @10 & @25 & @50 & @1 & @5 & @10 & @25 & @50 \\
\midrule
\multicolumn{13}{l}{\emph{OpenAI CLIP ViT-B/32}} \\
\textbf{pretrained}
  & 24.6 & 20.4 & 6.5 & 13.9 & 16.2 & 18.2 & 19.1
  & 6.5 & 20.5 & 28.5 & 42.0 & 53.3 \\
$+$ \textbf{CLIP}
  & 30.6 & 25.2 & 8.7 & 17.8 & 20.7 & 22.9 & 23.8
  & 8.7 & 24.4 & 34.1 & 46.7 & 57.4 \\
$+$ \textbf{SigLIP}
  & 31.5 & 26.4 & 9.1 & 18.9 & 21.8 & 24.1 & 25.0
  & 9.1 & 25.7 & 35.0 & 48.5 & 59.6 \\
  \rowcolor{appleteal!60}
$+$ \textbf{\ours}
  & \textbf{31.7} & \textbf{26.9} & \textbf{9.5} & \textbf{19.3} & \textbf{22.3} & \textbf{24.6} & \textbf{25.5}
  & \textbf{9.5} & \textbf{26.2} & \textbf{35.6} & \textbf{49.2} & \textbf{60.0} \\
\midrule
\multicolumn{13}{l}{\emph{OpenAI CLIP ViT-B/16}} \\
\textbf{pretrained}
  & 28.6 & 24.7 & 8.3 & 18.0 & 20.4 & 22.4 & 23.3
  & 8.3 & 24.5 & 32.6 & 45.3 & 55.4 \\
$+$ \textbf{CLIP}
  & 36.1 & 30.2 & 10.2 & 22.0 & 25.6 & 27.9 & 28.9
  & 10.2 & 29.0 & 39.8 & 52.7 & 62.8 \\
$+$ \textbf{SigLIP}
  & 37.4 & 32.5 & 11.0 & 24.3 & 27.8 & 30.2 & 31.1
  & 11.0 & 32.0 & 42.6 & 55.6 & 64.8 \\
  \rowcolor{appleteal!60}
$+$ \textbf{\ours}
  & \textbf{39.0} & \textbf{33.4} & \textbf{11.5} & \textbf{24.9} & \textbf{28.8} & \textbf{31.1} & \textbf{32.0}
  & \textbf{11.5} & \textbf{32.1} & \textbf{43.1} & \textbf{56.1} & \textbf{65.7} \\
\bottomrule
\end{tabular}%

\end{table}

\newpage

\noindent\textit{Zero-shot retrieval.} \Cref{tab:sota_vlm_image_retrieval_backbones,tab:sota_vlm_video_retrieval_backbones} repeat the image and video retrieval evaluation starting from OpenAI CLIP ViT-B/16 and LAION-400M CLIP ViT-B/32. The ViT-B/32 trends hold: for both models, \ours is the best or tied-best objective in every metric and direction on every dataset. With ViT-B/16, it doubles the \uAP of the pretrained model on MS COCO. The LAION-400M model is harder to improve by fine-tuning on CC12M. \uAP drops for InfoNCE on Flickr8k and MS COCO, and both InfoNCE and SigLIP degrade on ActivityNet and MSR-VTT, whereas \ours improves \uAP on all six datasets and is the only loss that improves every image retrieval metric. MSR-VTT is the hardest case. With ViT-B/16, InfoNCE and SigLIP fall below the pretrained model in \uAP and video$\to$text R@1, whereas \ours improves both, outperforming SigLIP by $3.3$ R@1. With LAION-400M, no loss recovers the pretrained video$\to$text recall, but \ours loses the least.

\noindent\textit{Zero-shot composed image retrieval.} \Cref{tab:sota_vlm_conconchi} reports results on ConCon-Chi~\citep{conconchi}, an instance-level composed image retrieval (CIR) benchmark where each query pairs a reference image of a specific object with a text modifier describing the target scene. Following the standard zero-shot CIR protocol, the retrieval score is the product of the image-to-image and text-to-image cosine similarities. All three losses improve over the pretrained model, and \ours is best in every metric with both backbones; SigLIP stays close with ViT-B/32, but the gap widens with ViT-B/16.

\noindent\textit{Zero-shot classification.}
\Cref{tab:sota_vlm_classification_backbones} repeats the zero-shot classification evaluation starting from the same two models, on the same four image and three video datasets and with the original prompt templates of each dataset. For both models, \ours is the best or tied-best loss in every metric on all seven datasets. \ours outperforms SigLIP by a small margin and InfoNCE by a clearer one. Starting from the LAION-400M checkpoint, InfoNCE even degrades top-1 accuracy of the pretrained model on ImageNet-1k and ImageNet-V2.

\begin{table}[t]
    \centering
    \caption{\textbf{Zero-shot classification on image and video datasets with other backbones and pretraining data.} Top-1, top-5, and mean per-class recall (mR) comparison of \ours with the CLIP (InfoNCE) and SigLIP losses. All models fine-tune the pretrained model in \textit{italics} for 5 epochs on CC12M.}
    \label{tab:sota_vlm_classification_backbones}
    \vspace{-8pt}
    \providecommand{\tbd}{\textcolor{red}{--}}
\newcolumntype{C}{>{\centering\arraybackslash}p{17pt}}
\scriptsize
\setlength{\tabcolsep}{1.5pt}
\begin{tabular}{lCCCCCCCCCCCCCCCCCC}
\toprule
\multirow{3}{*}{\textbf{method}} &
\multicolumn{12}{c}{\textbf{image datasets}} &
\multicolumn{6}{c}{\textbf{video datasets}} \\
\cmidrule(lr){2-13}\cmidrule(lr){14-19}
& \multicolumn{3}{c}{\textbf{IN-1k}} &
\multicolumn{3}{c}{\textbf{IN-V2}} &
\multicolumn{3}{c}{\textbf{IN-Rendition}} &
\multicolumn{3}{c}{\textbf{IN-Sketch}} &
\multicolumn{2}{c}{\textbf{K-400}} &
\multicolumn{2}{c}{\textbf{K-600}} &
\multicolumn{2}{c}{\textbf{K-700}} \\
\cmidrule(lr){2-4}\cmidrule(lr){5-7}\cmidrule(lr){8-10}\cmidrule(lr){11-13}
\cmidrule(lr){14-15}\cmidrule(lr){16-17}\cmidrule(lr){18-19}
& top-1 & top-5 & mR
& top-1 & top-5 & mR
& top-1 & top-5 & mR
& top-1 & top-5 & mR
& top-1 & top-5
& top-1 & top-5
& top-1 & top-5 \\
\midrule
\multicolumn{19}{l}{\textit{OpenAI CLIP ViT-B/16}} \\
\textbf{pretrained} &
64.4 & 89.7 & 64.4 &
57.8 & 85.0 & 57.8 &
73.5 & 91.0 & 71.7 &
44.2 & 72.1 & 44.3 &
43.4 & 69.3 &
20.0 & 35.8 &
32.2 & 56.5 \\
\textbf{$+$ CLIP} &
65.3 & 90.7 & 65.3 &
58.8 & 86.1 & 58.8 &
76.0 & 91.9 & 74.1 &
47.2 & 75.7 & 47.3 &
45.0 & 71.4 &
21.7 & 37.3 &
33.8 & 59.7 \\
\textbf{$+$ SigLIP} &
66.3 & 91.1 & 66.3 &
59.4 & 86.5 & 59.5 &
77.0 & \textbf{92.3} & \textbf{75.5} &
48.3 & 76.8 & 48.4 &
45.4 & 71.4 &
21.8 & 37.2 &
34.5 & 59.8 \\
\rowcolor{appleteal!60}
\textbf{$+$ \ours} &
\textbf{66.6} & \textbf{91.2} & \textbf{66.6} &
\textbf{60.0} & \textbf{86.6} & \textbf{60.1} &
\textbf{77.1} & \textbf{92.3} & \textbf{75.5} &
\textbf{49.0} & \textbf{77.0} & \textbf{49.0} &
\textbf{45.8} & \textbf{71.9} &
\textbf{22.0} & \textbf{37.5} &
\textbf{35.0} & \textbf{60.4} \\
\midrule
\multicolumn{19}{l}{\textit{LAION-400M CLIP ViT-B/32}} \\
\textbf{pretrained} &
60.2 & 85.9 & 60.2 & 52.3 & 79.4 & 52.3 & 70.8 & 88.6 & 69.1 & 46.5 & 72.6 & 46.5 & 35.5 & 60.0 & 32.7 & 56.7 & 25.3 & 47.2 \\
\textbf{$+$ CLIP} &
59.3 & 86.0 & 59.3 & 51.2 & 79.8 & 51.2 & 70.8 & 89.0 & 69.5 & 46.8 & 74.0 & 46.8 & 36.3 & 61.8 & 34.2 & 58.7 & 26.4 & 49.0 \\
\textbf{$+$ SigLIP} &
60.9 & 86.7 & 60.9 & 52.9 & 80.5 & 52.9 & 71.9 & \textbf{89.3} & 70.7 & 48.3 & 75.2 & 48.4 & 36.9 & 62.2 & 34.7 & 59.5 & 27.2 & 50.0 \\
\rowcolor{appleteal!60}
\textbf{$+$ \ours} &
\textbf{61.0} & \textbf{86.8} & \textbf{61.0} & \textbf{53.0} & \textbf{80.7} & \textbf{53.1} & \textbf{72.1} & \textbf{89.3} & \textbf{70.9} & \textbf{48.7} & \textbf{75.4} & \textbf{48.8} & \textbf{37.2} & \textbf{62.6} & \textbf{35.1} & \textbf{59.7} & \textbf{27.4} & \textbf{50.2} \\
\bottomrule
\end{tabular}

    \vspace{-10pt}
\end{table}

\subsection{Self-supervised learning}
\label{sec:supp_ssl}

\paragraph{In- and out-of-distribution performance.}
\Cref{tab:ssl_ood} breaks down the OOD mean. With ViT-S, \ours is ahead on every set under both probes. With ViT-B, the two models tie under LP, each leading on one shifted set, while \ours leads on every set under $k$NN. The OOD ordering follows the in-distribution ordering per probe: the better model in distribution is at least as good under shift, so neither loss trades in-distribution accuracy for robustness.

\paragraph{Transfer learning.}
\Cref{tab:ssl_transfer} shows that the transfer gain comes mostly from fine-grained datasets. On the six sub-category datasets, \ours is clearly ahead under both probes and at both scales, with the largest margins on Stanford Cars, FGVC Aircraft, and CUB-200, whereas on the four coarse object, scene, and texture sets the two losses are close, with InfoNCE slightly ahead at ViT-B. In fine-grained classes, hard negatives crowd the positives, which is exactly what a single shared threshold penalizes.

\paragraph{Robustness to corruptions.}
\Cref{tab:ssl_inc} reports ImageNet-C~\citep{hendrycks2019benchmarking} results by corruption family and severity. With ViT-S, \ours leads under both probes. With ViT-B, the two models tie under LP on every family, while under $k$NN \ours leads on every family and severity, by a margin that shrinks as corruptions grow stronger. \ours also retains a slightly larger share of its clean accuracy under both probes, so its robustness is not merely inherited from higher clean accuracy.

\begin{table}[t]
    \centering
    \caption{\textbf{In- and out-of-distribution classification with self-supervised pretraining.} Top-1 accuracy on ImageNet-1k (ID) and three OOD datasets (ImageNet-V2/-R/-Sketch) of MoCo~v3 models trained with InfoNCE or \ours. Linear probing (LP) and $k$NN on the frozen backbone; LP is trained on ImageNet-1k and applied to every test set without dataset-specific training.}
    \label{tab:ssl_ood}
    \newcolumntype{C}{>{\centering\arraybackslash}p{32pt}}
\scriptsize
\setlength{\tabcolsep}{3pt}
\begin{tabular}{l l l CCCCC}
    \toprule
    \multirow{2}{*}{\textbf{backbone}} & \multirow{2}{*}{\textbf{probe}} & \multirow{2}{*}{\textbf{loss}} & \multicolumn{1}{c}{\textbf{ID}} & \multicolumn{4}{c}{\textbf{OOD}} \\ \cmidrule(lr){4-4}\cmidrule(lr){5-8}
    & & & IN-1k & IN-V2 & IN-R & IN-Sketch & \textbf{avg} \\ \midrule
     &  & InfoNCE & 73.2 & 61.1 & 32.0 & 21.3 & 38.2 \\
    \rowcolor{appleteal!60} \cellcolor{white} & \cellcolor{white}\multirow{-2}{*}{LP} & \ours & \textbf{73.5} & \textbf{61.9} & \textbf{34.3} & \textbf{23.1} & \textbf{39.7} \\ \cmidrule(lr){2-8}
     &  & InfoNCE & 68.4 & 54.9 & 23.0 & 16.5 & 31.5 \\
    \rowcolor{appleteal!60} \cellcolor{white}\multirow{-4}{*}{\textbf{ViT-S}} & \cellcolor{white}\multirow{-2}{*}{$k$NN} & \ours & \textbf{70.9} & \textbf{58.4} & \textbf{27.4} & \textbf{20.5} & \textbf{35.4} \\ \midrule
     &  & InfoNCE & \textbf{76.7} & \textbf{65.2} & \textbf{35.9} & 24.9 & \textbf{42.0} \\
    \rowcolor{appleteal!60} \cellcolor{white} & \cellcolor{white}\multirow{-2}{*}{LP} & \ours & 75.6 & 63.9 & 35.8 & \textbf{26.1} & \textbf{42.0} \\ \cmidrule(lr){2-8}
     &  & InfoNCE & 71.4 & 58.6 & 29.6 & 21.8 & 36.6 \\
    \rowcolor{appleteal!60} \cellcolor{white}\multirow{-4}{*}{\textbf{ViT-B}} & \cellcolor{white}\multirow{-2}{*}{$k$NN} & \ours & \textbf{72.8} & \textbf{60.4} & \textbf{29.7} & \textbf{24.2} & \textbf{38.1} \\
    \bottomrule
\end{tabular}

\end{table}

\begin{table}[t]
    \centering
    \caption{\textbf{Transfer learning with self-supervised pretraining.} Top-1 accuracy on six fine-grained and four coarse-grained datasets of MoCo~v3 models trained with InfoNCE or \ours. Linear probing (LP) and $k$NN on the frozen backbone, using each dataset's own training split.}
    \label{tab:ssl_transfer}
    \resizebox{\linewidth}{!}{\newcolumntype{C}{>{\centering\arraybackslash}p{21pt}}
\scriptsize
\setlength{\tabcolsep}{3pt}
\begin{tabular}{l l l CCCCCCCCCCCCC}
    \toprule
    \multirow{2}{*}{\textbf{backbone}} & \multirow{2}{*}{\textbf{probe}} & \multirow{2}{*}{\textbf{loss}} & \multicolumn{7}{c}{\textbf{fine-grained (6)}} & \multicolumn{5}{c}{\textbf{coarse-grained (4)}} & \multicolumn{1}{c}{} \\ \cmidrule(lr){4-10}\cmidrule(lr){11-15}
    & & & Aircraft & CUB & Cars & Pets & Flowers & Food & \textbf{avg} & C-10 & C-100 & DTD & SUN & \textbf{avg} & \textbf{all (10)} \\ \midrule
     &  & InfoNCE & 37.6 & 64.4 & 33.3 & 88.3 & 88.1 & 75.8 & 64.6 & 93.9 & 79.9 & \textbf{73.3} & \textbf{63.1} & 77.6 & 69.8 \\
    \rowcolor{appleteal!60} \cellcolor{white} & \cellcolor{white}\multirow{-2}{*}{LP} & \ours & \textbf{45.2} & \textbf{76.4} & \textbf{48.3} & \textbf{91.7} & \textbf{89.6} & \textbf{76.2} & \textbf{71.2} & \textbf{94.9} & \textbf{81.0} & 73.1 & 62.6 & \textbf{77.9} & \textbf{73.9} \\ \cmidrule(lr){2-16}
     &  & InfoNCE & 29.3 & 47.6 & 20.6 & 82.3 & 78.2 & 63.6 & 53.6 & 92.0 & 73.6 & \textbf{69.3} & 53.3 & 72.1 & 61.0 \\
    \rowcolor{appleteal!60} \cellcolor{white}\multirow{-4}{*}{\textbf{ViT-S}} & \cellcolor{white}\multirow{-2}{*}{$k$NN} & \ours & \textbf{35.1} & \textbf{65.0} & \textbf{28.5} & \textbf{88.6} & \textbf{81.5} & \textbf{66.0} & \textbf{60.8} & \textbf{93.7} & \textbf{75.8} & 65.7 & \textbf{55.0} & \textbf{72.5} & \textbf{65.5} \\ \midrule
     &  & InfoNCE & 44.9 & 70.5 & 49.9 & 91.4 & 91.5 & \textbf{79.3} & 71.3 & 95.3 & \textbf{83.2} & \textbf{76.6} & \textbf{65.7} & \textbf{80.2} & 74.8 \\
    \rowcolor{appleteal!60} \cellcolor{white} & \cellcolor{white}\multirow{-2}{*}{LP} & \ours & \textbf{53.9} & \textbf{78.9} & \textbf{64.3} & \textbf{93.1} & \textbf{92.9} & 78.6 & \textbf{76.9} & \textbf{96.0} & 82.7 & 74.1 & 63.5 & 79.1 & \textbf{77.8} \\ \cmidrule(lr){2-16}
     &  & InfoNCE & 27.5 & 47.4 & 23.1 & 84.5 & 79.9 & 64.6 & 54.5 & 93.3 & 76.2 & \textbf{72.7} & \textbf{56.2} & \textbf{74.6} & 62.5 \\
    \rowcolor{appleteal!60} \cellcolor{white}\multirow{-4}{*}{\textbf{ViT-B}} & \cellcolor{white}\multirow{-2}{*}{$k$NN} & \ours & \textbf{35.6} & \textbf{64.2} & \textbf{33.7} & \textbf{89.9} & \textbf{83.6} & \textbf{66.3} & \textbf{62.2} & \textbf{94.8} & \textbf{77.4} & 67.0 & 55.8 & 73.8 & \textbf{66.8} \\
    \bottomrule
\end{tabular}
}
\end{table}

\begin{table}[h!]
    \centering
    \caption{\textbf{Robustness to corruptions with self-supervised pretraining.} Top-1 accuracy on ImageNet-C of MoCo~v3 models trained with InfoNCE or \ours, with linear probing (LP, trained on ImageNet-1k) and $k$NN on the frozen backbone. Corruption families average over their corruptions and the five severities; severities average over the 15 corruptions. rel.: ImageNet-C accuracy as a share of the ImageNet-1k (ID) accuracy.}
    \label{tab:ssl_inc}
    \newcolumntype{C}{>{\centering\arraybackslash}p{19pt}}
\scriptsize
\setlength{\tabcolsep}{3pt}
\begin{tabular}{l l l CCCCCCCCCCc}
    \toprule
    \multirow{2}{*}{\textbf{backbone}} & \multirow{2}{*}{\textbf{probe}} & \multirow{2}{*}{\textbf{loss}} & \multicolumn{4}{c}{\textbf{corruption family}} & \multicolumn{5}{c}{\textbf{severity}} & \multicolumn{2}{c}{} \\ \cmidrule(lr){4-7}\cmidrule(lr){8-12}
    & & & noise & blur & weather & digital & 1 & 2 & 3 & 4 & 5 & \textbf{all} & \textbf{rel.}\ (\%) \\ \midrule
     &  & InfoNCE & 41.4 & 36.2 & 46.0 & 53.9 & 62.1 & 53.5 & 46.7 & 36.2 & 24.5 & 44.6 & 60.9 \\
    \rowcolor{appleteal!60} \cellcolor{white} & \cellcolor{white}\multirow{-2}{*}{LP} & \ours & \textbf{41.5} & \textbf{38.5} & \textbf{48.1} & \textbf{56.0} & \textbf{63.9} & \textbf{55.6} & \textbf{48.7} & \textbf{37.9} & \textbf{25.4} & \textbf{46.3} & \textbf{63.0} \\ \cmidrule(lr){2-14}
     &  & InfoNCE & 35.3 & 31.2 & 40.8 & 49.4 & 58.2 & 49.0 & 41.1 & 29.9 & 18.9 & 39.4 & 57.7 \\
    \rowcolor{appleteal!60} \cellcolor{white}\multirow{-4}{*}{\textbf{ViT-S}} & \cellcolor{white}\multirow{-2}{*}{$k$NN} & \ours & \textbf{38.5} & \textbf{36.7} & \textbf{46.4} & \textbf{55.6} & \textbf{63.1} & \textbf{54.5} & \textbf{47.1} & \textbf{35.7} & \textbf{23.2} & \textbf{44.7} & \textbf{63.1} \\ \midrule
     &  & InfoNCE & 45.5 & 41.9 & \textbf{51.0} & \textbf{58.9} & 66.6 & 58.6 & 52.0 & 41.6 & \textbf{29.1} & 49.6 & 64.7 \\
    \rowcolor{appleteal!60} \cellcolor{white} & \cellcolor{white}\multirow{-2}{*}{LP} & \ours & \textbf{45.7} & \textbf{42.3} & 50.9 & 58.7 & \textbf{66.8} & \textbf{58.7} & \textbf{52.1} & \textbf{41.7} & \textbf{29.1} & \textbf{49.7} & \textbf{65.7} \\ \cmidrule(lr){2-14}
     &  & InfoNCE & 41.9 & 39.6 & 49.3 & 56.1 & 64.1 & 56.4 & 49.6 & 38.7 & 26.3 & 47.0 & 65.9 \\
    \rowcolor{appleteal!60} \cellcolor{white}\multirow{-4}{*}{\textbf{ViT-B}} & \cellcolor{white}\multirow{-2}{*}{$k$NN} & \ours & \textbf{43.9} & \textbf{40.7} & \textbf{49.8} & \textbf{58.9} & \textbf{66.0} & \textbf{58.1} & \textbf{51.3} & \textbf{40.3} & \textbf{27.5} & \textbf{48.6} & \textbf{66.8} \\
    \bottomrule
\end{tabular}

\end{table}

\begin{wraptable}{r}{0.48\linewidth}
    \centering
    \caption{\textbf{Music tagging with self-supervised pretraining.} ROC-AUC and PR-AUC on MagnaTagATune of CLMR models trained with InfoNCE or \ours, with linear probing on the frozen encoder.}
    \vspace{-8pt}
    \label{tab:sota_audio_ssl}
    \scalebox{1.}{\newcolumntype{C}{>{\centering\arraybackslash}p{25pt}}
\scriptsize
\setlength{\tabcolsep}{4pt}
\begin{tabular}{lcc}
    \toprule
    \textbf{method} & ROC-AUC & PR-AUC \\ \midrule
    \textbf{InfoNCE}          &  88.7 &  35.6 \\
    \rowcolor{appleteal!60}
    \textbf{\ours}     &  \textbf{89.3} &  \textbf{37.1} \\
    \bottomrule
\end{tabular}
}
    \vspace{-10pt}
\end{wraptable}

\paragraph{Audio-based SSL.}
We also experiment with audio contrastive learning using the CLMR~\citep{spijkervet2021contrastive} framework, with the settings of \Cref{sec:ssl_settings}. Substituting InfoNCE with the proposed loss improves music tagging on MagnaTagATune~\citep{law2009evaluation}, as shown in \Cref{tab:sota_audio_ssl}, with the larger gain in PR-AUC. PR-AUC, the area under the precision--recall curve, is the tagging analogue of \uAP, as \ours directly optimizes a closely related objective.

\begin{figure}[t]
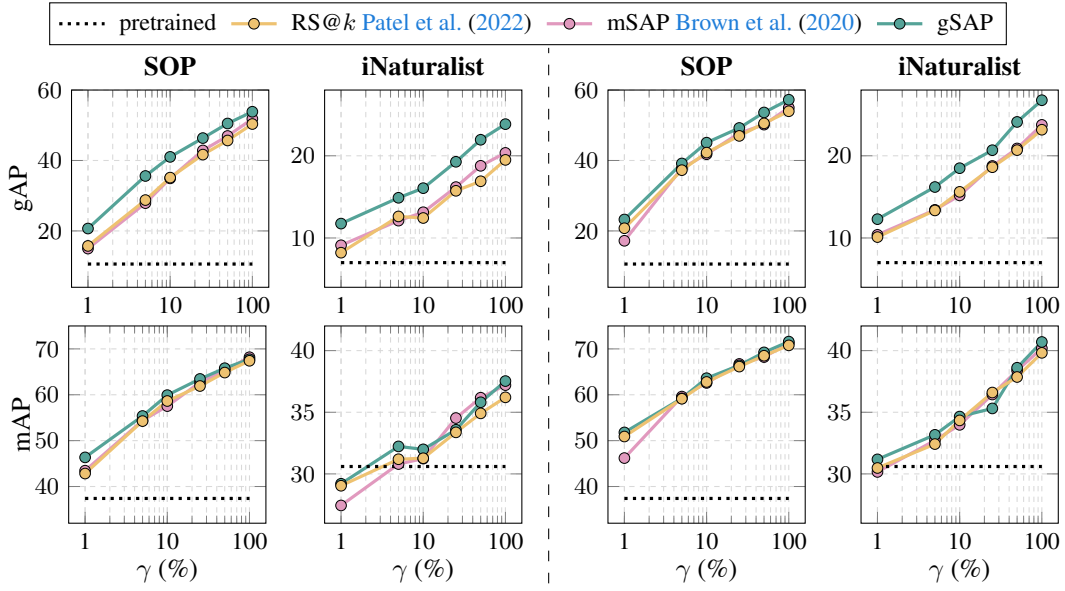

    \centering
    \pgfplotsset{every tick label/.append style={font=\footnotesize}}

\pgfplotsset{
    recall/.style={color=applemustard!80, mark=*, mark size=2.0, opacity=1.0, mark options={solid, draw=black, line width=0.35pt}, line width=1.2pt},
    sap/.style={color=applepink!80, mark=*, mark size=2., opacity=1.0, mark options={solid, draw=black, line width=0.35pt}, line width=1.2pt},
    sgap/.style={color=appleteal, mark=*, mark size=2., opacity=1.0, mark options={solid, draw=black, line width=0.35pt}, line width=1.2pt},
    spc2/.style={solid, thick},
    spc3/.style={solid, thick},
    spc4/.style={solid, thick},
    spc5/.style={solid, thick},
}

\begin{tikzpicture}
    \begin{axis}[
        hide axis,
        scale only axis,
        height=0cm,
        width=0cm,
        legend style={
            draw=black,
            cells={anchor=west},
            legend columns=7,
            column sep=3pt,
            font=\footnotesize,
            anchor=north,
        }
    ]
        \input{data/dino_uap}
        \addlegendimage{dotted, thick, color=black, line width=1.2pt};
        \addlegendentry{pretrained}
        
        \addlegendimage{solid, thick, recall}
        \addlegendentry{RS@$k$~\cite{patel2022recall}}
        
        \addlegendimage{solid, thick, sap}
        \addlegendentry{mSAP~\cite{brown2020smooth}}
        
        \addlegendimage{solid, thick, sgap}
        \addlegendentry{\ours}

        \addplot[draw=none] coordinates {(0,0)};
    \end{axis}
\end{tikzpicture}

\begin{tabular}{@{}c@{\hspace{1pt}}c@{\hspace{6pt}}:c@{\hspace{1pt}}c@{}}
\begin{tikzpicture}
    \begin{axis}[%
      width=0.3\linewidth,
      height=0.3\linewidth,
      ylabel={\uAP},
      xlabel={},
      grid=both,
      grid style={color=lightgray!60, dash pattern=on 2pt off 2pt},
      xlabel style={yshift=3pt},
      ylabel style={yshift=-5pt},
      xmode=log,
      ymax=60,
      ymin=4,
      xtick={1, 10, 100},
      xticklabels={1, 10, 100},
      title={\textbf{SOP}},
      title style={yshift=-4pt},
      ]
    \input{data/dino_uap}
    \addplot[spc2, thick, sap] table[x=percentage,y=SAP_2] {\sop};
    \addplot[spc2, thick, sgap] table[x=percentage,y=SGAP_2] {\sop};
    \addplot[spc2, thick, recall] table[x=percentage,y=RecK_2] {\sop};
    
    \addplot[dotted, thick, color=black, line width=1.2pt] coordinates {(1, 10.6) (100, 10.6)};

    \end{axis}
\end{tikzpicture}

&

\begin{tikzpicture}
    \begin{axis}[%
      width=0.3\linewidth,
      height=0.3\linewidth,
      xlabel={},
      grid=both,
      grid style={color=lightgray!60, dash pattern=on 2pt off 2pt},
      xlabel style={yshift=3pt},
      ylabel style={yshift=-4pt},
      xmode=log,
      ymax=28,
      ymin=4,
      xtick={1, 10, 100},
      xticklabels={1, 10, 100},
      title={\textbf{iNaturalist}},
      title style={yshift=-4pt},
      ]
    \input{data/dino_uap}
    \addplot[spc2, thick, sap] table[x=percentage,y=SAP_2] {\inat};
    \addplot[spc2, thick, sgap] table[x=percentage,y=SGAP_2] {\inat};
    \addplot[spc2, thick, recall] table[x=percentage,y=RecK_2] {\inat};
    
    \addplot[dotted, thick, color=black, line width=1.2pt] coordinates {(1, 7.0) (100, 7.0)};

    \end{axis}
\end{tikzpicture}

&

\begin{tikzpicture}
    \begin{axis}[%
      width=0.3\linewidth,
      height=0.3\linewidth,
      ylabel={},
      xlabel={},
      grid=both,
      grid style={color=lightgray!60, dash pattern=on 2pt off 2pt},
      xlabel style={yshift=3pt},
      ylabel style={yshift=-4pt},
      xmode=log,
      ymax=60,
      ymin=4,
      xtick={1, 10, 100},
      xticklabels={1, 10, 100},
      title={\textbf{SOP}},
      title style={yshift=-4pt},
      ]
    \input{data/dino_uap}
    \addplot[spc4, thick, sap] table[x=percentage,y=SAP_4] {\sop};
    \addplot[spc4, thick, sgap] table[x=percentage,y=SGAP_4] {\sop};
    \addplot[spc4, thick, recall] table[x=percentage,y=RecK_4] {\sop};

    \addplot[dotted, thick, color=black, line width=1.2pt] coordinates {(1, 10.6) (100, 10.6)};
    
    \end{axis}
\end{tikzpicture}

&

\begin{tikzpicture}
    \begin{axis}[%
      width=0.3\linewidth,
      height=0.3\linewidth,
      xlabel={},
      grid=both,
      grid style={color=lightgray!60, dash pattern=on 2pt off 2pt},
      xlabel style={yshift=3pt},
      ylabel style={yshift=-4pt},
      xmode=log,
      ymax=28,
      ymin=4,
      xtick={1, 10, 100},
      xticklabels={1, 10, 100},
      title={\textbf{iNaturalist}},
      title style={yshift=-4pt},
      ]
    \input{data/dino_uap}
    \addplot[spc4, thick, sap] table[x=percentage,y=SAP_4] {\inat};
    \addplot[spc4, thick, sgap] table[x=percentage,y=SGAP_4] {\inat};
    \addplot[spc4, thick, recall] table[x=percentage,y=RecK_4] {\inat};
    
    \addplot[dotted, thick, color=black, line width=1.2pt] coordinates {(1, 7.0) (100, 7.0)};

    \end{axis}
\end{tikzpicture}

\\
[-2pt] %

\begin{tikzpicture}
\begin{axis}[%
      width=0.3\linewidth,
      height=0.3\linewidth,
      ylabel={mAP},
      xlabel={$\gamma$ (\%)},
      grid=both,
      grid style={color=lightgray!60, dash pattern=on 2pt off 2pt},
      xlabel style={yshift=3pt},
      ylabel style={yshift=-5pt},
      xmode=log,
      ymax=75,
      ymin=32,
      xtick={1, 10, 100},
      xticklabels={1, 10, 100},
      title style={yshift=-4pt},
      ]
    
    \input{data/dino_map}
    \addplot[spc2, thick, sap] table[x=percentage,y=SAP_2] {\sop};
    \addplot[spc2, thick, sgap] table[x=percentage,y=SGAP_2] {\sop};
    \addplot[spc2, thick, recall] table[x=percentage,y=RecK_2] {\sop};
    
    \addplot[dotted, thick, color=black, line width=1.2pt] coordinates {(1, 37.4) (100, 37.4)};

    \end{axis}
\end{tikzpicture}

&

\begin{tikzpicture}
    \begin{axis}[%
      width=0.3\linewidth,
      height=0.3\linewidth,
      xlabel={$\gamma$ (\%)},
      grid=both,
      grid style={color=lightgray!60, dash pattern=on 2pt off 2pt},
      xlabel style={yshift=3pt},
      ylabel style={yshift=-4pt},
      xmode=log,
      ymax=42,
      ymin=26,
      xtick={1, 10, 100},
      xticklabels={1, 10, 100},
      title style={yshift=-4pt},
      ]
  
    \input{data/dino_map}
    \addplot[spc2, thick, sap] table[x=percentage,y=SAP_2] {\inat};
    \addplot[spc2, thick, sgap] table[x=percentage,y=SGAP_2] {\inat};
    \addplot[spc2, thick, recall] table[x=percentage,y=RecK_2] {\inat};
    
    \addplot[dotted, thick, color=black, line width=1.2pt] coordinates {(1, 30.6) (100, 30.6)};

    \end{axis}
\end{tikzpicture}

&

\begin{tikzpicture}
    \begin{axis}[%
      width=0.3\linewidth,
      height=0.3\linewidth,
      ylabel={},
      xlabel={$\gamma$ (\%)},
      grid=both,
      grid style={color=lightgray!60, dash pattern=on 2pt off 2pt},
      xlabel style={yshift=3pt},
      ylabel style={yshift=-4pt},
      xmode=log,
      ymax=75,
      ymin=32,
      xtick={1, 10, 100},
      xticklabels={1, 10, 100},
      title style={yshift=-4pt},
      ]

    \input{data/dino_map}
    \addplot[spc4, thick, sap] table[x=percentage,y=SAP_4] {\sop};
    \addplot[spc4, thick, sgap] table[x=percentage,y=SGAP_4] {\sop};
    \addplot[spc4, thick, recall] table[x=percentage,y=RecK_4] {\sop};

    \addplot[dotted, thick, color=black, line width=1.2pt] coordinates {(1, 37.4) (100, 37.4)};
    
    \end{axis}
\end{tikzpicture}

&

\begin{tikzpicture}
    \begin{axis}[%
      width=0.3\linewidth,
      height=0.3\linewidth,
      xlabel={$\gamma$ (\%)},
      grid=both,
      grid style={color=lightgray!60, dash pattern=on 2pt off 2pt},
      xlabel style={yshift=3pt},
      ylabel style={yshift=-4pt},
      xmode=log,
      ymax=42,
      ymin=26,
      xtick={1, 10, 100},
      xticklabels={1, 10, 100},
      title style={yshift=-4pt},
      ]

    \input{data/dino_map}
    \addplot[spc4, thick, sap] table[x=percentage,y=SAP_4] {\inat};
    \addplot[spc4, thick, sgap] table[x=percentage,y=SGAP_4] {\inat};
    \addplot[spc4, thick, recall] table[x=percentage,y=RecK_4] {\inat};
    
    \addplot[dotted, thick, color=black, line width=1.2pt] coordinates {(1, 30.6) (100, 30.6)};

    \end{axis}
\end{tikzpicture}

\end{tabular}
    \vspace{-8pt}
    \caption{\textbf{Training with scarce annotated data.} \uAP (top) and mAP (bottom) of RS@$k$, mSAP, and \ours on SOP and iNaturalist, fine-tuning DINOv3 under the scarce-data protocol of \Cref{sec:metric_learning_impl}, which varies the percentage of classes $\gamma$ and the number of samples per class $\rho$ ($\rho{=}2$: left pair, $\rho{=}4$: right pair). The full training sets have 11,318 (SOP) and 5,690 (iNaturalist) classes. Dotted: pretrained model.}
    \vspace{-8pt}
    \label{fig:scarce_data_uap}
\end{figure}

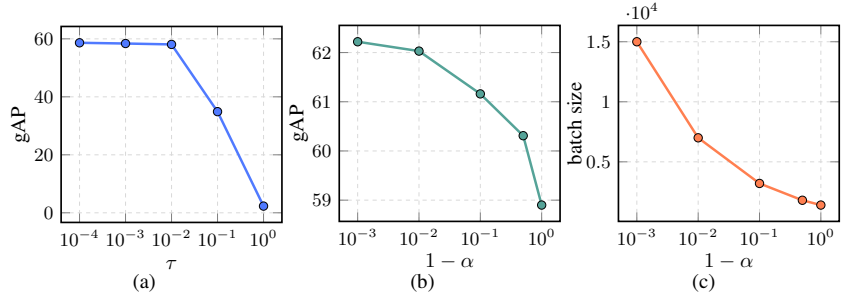
\begin{figure}[h!]
    \centering
    \resizebox{0.8\linewidth}{!}{\pgfplotsset{every tick label/.append style={font=\footnotesize}}

\begin{tabular}{@{\nssp}c@{\nxssp}c@{\nxssp}c@{\nxssp}}

\begin{tikzpicture}
\begin{axis}[
    width=.38\linewidth,
    height=.35\linewidth,
    xlabel={{\color{white} 1} $\tau$ {\color{white} 1}},
    ylabel={\uAP},
    xmode=log,
    xtick={1, 0.1, 0.01, 0.001, 0.0001},
    grid=both,
    grid style={color=lightgray!60, dash pattern=on 2pt off 2pt},
    ymajorgrids=true,
    xmajorgrids=true,
    xlabel style={yshift=3pt},
    ylabel style={yshift=-5pt},
    thick,
]
    
    \addplot[solid, color=appleblue, line width=1.2pt, mark=*, mark size=2., opacity=1.0, mark options={solid, draw=black, line width=0.35pt}] coordinates {
                                                    (1.0, 2.31)
                                                    (0.1, 34.89)
                                                    (0.01, 58.08)
                                                    (0.001, 58.39)
                                                    (0.0001, 58.64)};
    
\end{axis}
\end{tikzpicture}

     &

\begin{tikzpicture}
\begin{axis}[
    width=.38\linewidth,
    height=.35\linewidth,
    xlabel={$1-\alpha$},
    ylabel={gAP},
    xmode=log,
    xtick={1,0.1,0.01,0.001},
    grid=both,
    grid style={color=lightgray!60, dash pattern=on 2pt off 2pt},
    ymajorgrids=true,
    xmajorgrids=true,
    xlabel style={yshift=3pt},
    ylabel style={yshift=-5pt},
    thick,
]

\addplot[
    color=appleteal,
    line width=1.2pt,
    mark=*,
    mark size=2,
    mark options={solid, draw=black, line width=0.35pt}
] coordinates {
    (1.0,58.9)
    (0.5,60.31)
    (0.1,61.16)
    (0.01,62.03)
    (0.001,62.22)
};
\end{axis}
\end{tikzpicture}

&

\begin{tikzpicture}
\begin{axis}[
    width=.38\linewidth,
    height=.35\linewidth,
    xlabel={$1-\alpha$},
    ylabel={batch size},
    xmode=log,
    xtick={1,0.1,0.01,0.001},
    grid=both,
    grid style={color=lightgray!60, dash pattern=on 2pt off 2pt},
    ymajorgrids=true,
    xmajorgrids=true,
    xlabel style={yshift=3pt},
    ylabel style={yshift=-5pt},
    thick,
]

\addplot[
    color=appleorange,
    line width=1.2pt,
    mark=*,
    mark size=2,
    mark options={solid, draw=black, line width=0.35pt}
] coordinates {
    (1.0,1400)
    (0.5,1800)
    (0.1,3200)
    (0.01,7000)
    (0.001,15000)
};
\end{axis}
\end{tikzpicture}

\vspace{-5pt}
\\
(a) & (b) & (c)
\end{tabular}} 
    \vspace{-8pt}
    \caption{\textbf{Impact of hyper-parameters.} (a) \uAP for different temperatures $\tau$, without negative dropping. (b), (c) Impact of negative dropping on \uAP and on the batch size, set to the largest number of images that fits on one H200 GPU. All results on SOP, fine-tuning DINOv3 with full-scale training.}
    \vspace{-8pt}
    \label{fig:ablation}
\end{figure}

\subsection{Metric learning}
\label{sec:metric_learning_add}

\paragraph{Comparison on limited data settings.}
We further evaluate the methods in a scarce data regime controlled by two parameters: $\gamma$, the percentage of training classes, and $\rho$, the number of samples per class. Evaluation settings are described in \Cref{sec:metric_learning_impl}. \Cref{fig:scarce_data_uap} reports the performance of \ours compared with the other two ranking surrogates (mSAP and RS@$k$) under different settings. \ours is consistently the best in terms of \uAP, with the largest margins at $\rho{=}2$, reaching up to 6 \uAP over the second-best method. When training with a small $\gamma$, \ours often surpasses competitors trained with more than twice the data. We attribute this to its formulation over all pairwise similarities, whereas competing losses operate row-wise and observe a single positive per row in this setting. Similarly, for $\rho{=}4$, \ours remains the best, and trained models generally surpass the pretrained model, reflecting the benefit of additional annotations and multiple positives per row. The bottom row of the figure shows the mAP results, where all three losses perform similarly, achieving nearly identical mAP scores, in line with the results from the main paper. The pretrained model already has good mAP.

\paragraph{Impact of hyper-parameters.}
\Cref{fig:ablation} presents the \uAP performance of the model trained with \ours under different settings. Performance is very low for large $\tau$ values greater than 0.1; however, it peaks and plateaus as $\tau$ decreases below 0.01. Additionally, we assess the impact of the drop rate $\alpha$ on the batch size and, consequently, on performance. It is evident that the larger the batch size that fits in memory, which is determined by the drop rate, the better the performance.

\begin{table}[t]
    \centering
    \begin{minipage}[t]{0.49\linewidth}
        \centering
        \caption{\textbf{Combination with similarity-scale regularizers.} Recall@1 (R@1), mAP, and \uAP when TCM or KoLeo is added to each loss. DINOv3, full-scale training, no negative dropping.
        \label{tab:calibration}
        }
        \vspace{-6pt}
        \resizebox{\linewidth}{!}{\scriptsize
\setlength{\tabcolsep}{2.5pt}
\begin{tabular}{cl ccc ccc}
    \toprule
    \multirow{2}{*}{\textbf{regularizer}} & \multirow{2}{*}{\textbf{method}} & \multicolumn{3}{c}{\textbf{SOP}} & \multicolumn{3}{c}{\textbf{iNaturalist}} \\ \cmidrule(lr){3-5} \cmidrule(lr){6-8}
    & & R@1 & mAP & \uAP & R@1 & mAP & \uAP \\ \midrule
     & \textbf{RS@$k$} & 86.7 & 72.5 & 56.7 & 81.0 & 44.5 & 26.6 \\
     & \textbf{mSAP} & \textbf{86.9} & \textbf{72.9} & 57.0 & 81.3 & 45.4 & 27.8 \\
    \rowcolor{appleteal!60} \cellcolor{white}\multirow{-3}{*}{\textbf{--}} & \textbf{\ours} & 86.8 & \textbf{72.9} & \textbf{58.9} & \textbf{81.5} & \textbf{45.5} & \textbf{31.6} \\ \midrule
     & \textbf{RS@$k$} & 86.9 & 72.6 & 57.9 & 81.7 & 44.9 & 30.2 \\
     & \textbf{mSAP} & 86.5 & 72.0 & 56.7 & \textbf{81.9} & \textbf{45.9} & 31.0 \\
    \rowcolor{appleteal!60} \cellcolor{white}\multirow{-3}{*}{\textbf{TCM}} & \textbf{\ours} & \textbf{87.0} & \textbf{72.8} & \textbf{58.7} & 81.5 & 45.3 & \textbf{31.4} \\ \midrule
     & \textbf{RS@$k$} & 83.5 & 66.4 & 49.7 & 79.9 & 41.0 & 24.6 \\
     & \textbf{mSAP} & 85.3 & 69.9 & 53.6 & 80.5 & 42.5 & 25.9 \\
    \rowcolor{appleteal!60} \cellcolor{white}\multirow{-3}{*}{\textbf{KoLeo}} & \textbf{\ours} & \textbf{85.4} & \textbf{70.1} & \textbf{55.5} & \textbf{81.5} & \textbf{45.1} & \textbf{31.2} \\
    \bottomrule
\end{tabular}
}
    \end{minipage}\hfill
    \begin{minipage}[t]{0.49\linewidth}
        \centering
        \caption{\textbf{AP approximations as mAP or \uAP surrogates.} Recall@1 (R@1), mAP, and \uAP on SOP and iNaturalist. Each AP approximation is optimized as a surrogate of either mAP or \uAP; the \uAP surrogate with Smooth-AP is \ours.
        \label{tab:surrogates}
        }
        \vspace{-8pt}
        \resizebox{\linewidth}{!}{\scriptsize
\setlength{\tabcolsep}{2.5pt}
\begin{tabular}{lc ccc ccc}
    \toprule
    \multirow{2}{*}{\textbf{AP approx.}} & \multirow{2}{*}{\textbf{surrogate}} & \multicolumn{3}{c}{\textbf{SOP}} & \multicolumn{3}{c}{\textbf{iNaturalist}} \\ \cmidrule(lr){3-5} \cmidrule(lr){6-8}
    & & R@1 & mAP & \uAP & R@1 & mAP & \uAP \\ \midrule
     & mAP & \textbf{87.0} & \textbf{73.0} & 57.3 & 81.2 & 45.3 & 27.4 \\
    \rowcolor{appleteal!60} \cellcolor{white}\multirow{-2}{*}{\textbf{SoftBin}} & \uAP & 86.7 & 72.7 & 58.6 & 81.3 & 44.8 & 30.6 \\ \midrule
     & mAP & 86.5 & 72.3 & 56.6 & 80.7 & 44.5 & 26.2 \\
    \rowcolor{appleteal!60} \cellcolor{white}\multirow{-2}{*}{\textbf{BlackBox}} & \uAP & 86.4 & 72.4 & 58.2 & 80.7 & 44.6 & 29.5 \\ \midrule
     & mAP & 85.5 & 70.1 & 54.2 & 80.5 & 44.0 & 29.5 \\
    \rowcolor{appleteal!60} \cellcolor{white}\multirow{-2}{*}{\textbf{ROADMAP}} & \uAP & 86.8 & 72.6 & 58.8 & 81.2 & 44.9 & 31.1 \\ \midrule
     & mAP & 86.9 & 72.9 & 57.0 & 81.3 & 45.4 & 27.8 \\
    \rowcolor{appleteal!60} \cellcolor{white}\multirow{-2}{*}{\textbf{Smooth-AP}} & \uAP & 86.8 & 72.9 & \textbf{58.9} & \textbf{81.5} & \textbf{45.5} & \textbf{31.6} \\
    \bottomrule
\end{tabular}
}
    \end{minipage}
\end{table}

\paragraph{Combination with similarity-scale regularizers.} We add two regularizers proposed to constrain the similarity scale, TCM~\citep{zhang2023threshold} and KoLeo~\citep{sablayrolles2018spreading}, as auxiliary losses to each ranking loss (\Cref{tab:calibration}). TCM helps in general the per-query losses and brings them closer to \ours, but not consistently, \eg performance of mSAP on SOP drops. Also, it adds nothing on top of \ours, while KoLeo hurts \uAP everywhere. TCM constrains the similarity scale across queries, which \ours already achieves through its loss alone, whereas KoLeo spreads the embeddings apart and does not target this consistency. An explicit regularizer thus recovers part of what \ours provides, at the price of extra hyper-parameters, and without consistent effectiveness.

\paragraph{Performance with other AP approximations.} \Cref{tab:surrogates} replaces the default smooth approximation~\citep{brown2020smooth} with SoftBin~\citep{revaud2019learning}, BlackBox~\citep{rolinek2020optimizing} and ROADMAP~\citep{ramzi2021robust}, each implemented both as an mAP and as a \uAP surrogate, with its own hyper-parameters tuned based on the values recommended in the original work. Whatever the approximation, the \uAP surrogate beats its mAP counterpart in \uAP by a clear margin at similar mAP and R@1. The gain therefore comes from ranking the whole similarity matrix at once, not from the particular AP approximation.

\begin{wraptable}{r}{0.52\linewidth}
    \centering
    \vspace{-12pt}
    \caption{\textbf{Negative dropping and batch size.} \uAP, mAP, and R@1 of \ours on SOP for different drop rates $\alpha$, the fraction of the lowest negative similarities dropped during training. A higher drop rate allows a larger batch size on the same GPU, which improves \uAP.}
    \label{tab:drop}
    \vspace{-8pt}
    \scalebox{0.99}{
    \newcolumntype{C}{>{\centering\arraybackslash}p{20pt}}
\scriptsize
\setlength{\tabcolsep}{4pt}
\begin{tabular}{lcrCCC}
    \toprule
    $\boldsymbol{\alpha}$ & \textbf{batch size} & \textbf{neg sim.} & \textbf{\uAP} & \textbf{mAP} & \textbf{R@1}  \\ \midrule
    0\%       & 1,000 & 1M   & 58.9 & 72.9 & 86.8 \\
    90\%      & 1,000 & 100k & 58.9 & 72.9 & 86.7 \\
    99\%      & 1,000 & 10k  & 59.0 & 72.8 & 86.6 \\
    99.9\%    & 1,000 & 1k   & 58.0 & 71.9 & 86.1 \\
    99.99\%   & 1,000 & 100  & 48.0 & 62.5 & 81.0 \\
    99.999\%  & 1,000 & 10   & 40.4 & 56.1 & 77.5 \\
    99.9999\% & 1,000 & 1    & 27.5 & 45.2 & 69.7 \\ \midrule
    99\%      & 7,000 & 500k & 60.0 & 72.8 & 87.1  \\    
    \bottomrule
\end{tabular}

    }
    \vspace{-6pt}
\end{wraptable}

\paragraph{Impact of negative dropping.} 
\Cref{tab:drop} analyzes the effect of negative dropping by varying the drop percentage $\alpha$. Performance remains remarkably stable even when discarding a large fraction of negatives. In particular, removing up to $99.9\%$ of negatives, reducing their number by three orders of magnitude, has only a marginal impact. Performance degrades only when the remaining negatives become extremely few, suggesting that a minimal number is still necessary for effective training. Increasing the batch size while keeping $\alpha=99\%$ improves \uAP, showing that the proposed loss benefits from additional pairwise interactions even when most easy negatives are discarded.

\subsection{Tasks with distractor queries}
\label{sec:supp_distractors}

\begin{table}[t]
    \centering
    \caption{\textbf{Per-query metrics of the tasks with distractor queries.} Each benchmark's own metric, which does not depend on the distractor queries. ACC: top-1 accuracy. BAKS: balanced accuracy on known samples. \textbf{Best} in bold, second best \underline{underlined}.}
    \vspace{-8pt}
    \label{tab:rebuttal_tasks_pq}
    \scriptsize
\setlength{\tabcolsep}{6pt}
\begin{tabular}{lccccc}
    \toprule
    \multirow{2}{*}{\textbf{method}} & \multicolumn{1}{c}{\textbf{DISC21}} & \multicolumn{1}{c}{\textbf{Met}} & \multicolumn{1}{c}{\textbf{ILIAS}} & \multicolumn{2}{c}{\textbf{WildlifeReID-10k}} \\ \cmidrule(lr){2-2} \cmidrule(lr){3-3} \cmidrule(lr){4-4} \cmidrule(lr){5-6}
    & ACC & ACC & mAP & ACC & BAKS \\ \midrule
    \textbf{baseline} & 71.6 & 56.3 & 33.2 & 79.2 & 76.8 \\
    \textbf{InfoNCE}  & \underline{75.0} & \underline{74.6} & 33.3 & 84.0 & 82.3 \\
    \textbf{mSAP}     & 74.6 & \textbf{74.7} & \underline{33.4} & \underline{86.3} & \textbf{84.9} \\
    \rowcolor{appleteal!60}
    \textbf{\ours}    & \textbf{75.1} & 74.5 & \textbf{34.6} & \textbf{86.4} & \underline{84.8} \\
    \bottomrule
\end{tabular}

\end{table}

\paragraph{Per-query metrics.}
\Cref{tab:rebuttal_tasks_pq} reports the per-query metrics of the four tasks with distractor queries, \ie each benchmark's own metric, which by construction ignores those queries. All three fine-tuned losses improve over the baseline on every column, and \ours is best or on par with the best on every column. The three fine-tuned losses differ far less on these measures than in \uAP, which points to the consistency across queries as the source of the \uAP differences, as discussed in the main paper.

\begin{wraptable}{r}{0.44\linewidth}
    \centering
    \vspace{0pt}
    \caption{\textbf{Recall at $90\%$ precision on DISC21.} Share of the true copies retrieved under one global threshold set for $90\%$ precision, without and with score normalization (SN), in \%. Same models as \Cref{tab:rebuttal_tasks}. \textbf{Best} in bold, second best \underline{underlined}.}
    \vspace{-8pt}
    \label{tab:disc21_rp90}
    \scriptsize
\setlength{\tabcolsep}{6pt}
\begin{tabular}{lcc}
    \toprule
    \textbf{method} & R@P90 & R@P90$_{\text{SN}}$ \\ \midrule
    \textbf{baseline} & 23.7 & 55.8 \\
    \textbf{InfoNCE}  & 21.5 & 54.9 \\
    \textbf{mSAP}     & \underline{26.4} & \textbf{57.1} \\
    \rowcolor{appleteal!60}
    \textbf{\ours}    & \textbf{28.6} & \textbf{57.1} \\
    \bottomrule
\end{tabular}
\end{wraptable}

\paragraph{Operating point on DISC21.}
\Cref{tab:disc21_rp90} reports R@P90, the metric of the SSCD benchmark, which is the recall obtained when one threshold, shared by all queries, is set for $90\%$ precision over all query--reference pairs. It is a global rather than a per-query measure, which is why it is not in \Cref{tab:rebuttal_tasks_pq}. Under one threshold, \ours retrieves the largest share of the copies, ahead of mSAP and well ahead of InfoNCE. SN lifts every loss and closes the gap between the two AP surrogates in terms of R@P90, so a post-hoc correction recovers much of the operating-point gap.

\begin{wraptable}{l}{0.50\linewidth}
    \centering
    \vspace{-12pt}
    \caption{\textbf{Open-set recognition on WildlifeReID-10k.} A query of an unknown individual has to be rejected. AUROC separates known from unknown queries by their top-1 similarity. The other three columns use one threshold per source dataset, set for a $95\%$ true positive rate on its known queries, and report the true negative rate (TNR) on the unknown queries, and the balanced accuracy on the known (BAKS) and on the unknown (BAUS) queries at that same threshold. All in \%. Same models as \Cref{tab:rebuttal_tasks}. \textbf{Best} in bold, second best \underline{underlined}.}
    \vspace{-8pt}
    \label{tab:wildlife_openset}
    \scriptsize
\setlength{\tabcolsep}{6pt}
\begin{tabular}{lcccc}
    \toprule
    \multirow{2}{*}{\textbf{method}} & \multirow{2}{*}{AUROC} & \multicolumn{3}{c}{\textbf{at 95\% TPR}} \\ \cmidrule(lr){3-5}
    & & TNR & BAKS & BAUS \\ \midrule
    \textbf{baseline} & 78.9 & 28.2 & 74.9 & 29.0 \\
    \textbf{InfoNCE} & 84.8 & 46.8 & 79.8 & 47.1 \\
    \textbf{mSAP} & \underline{87.4} & \underline{53.5} & \textbf{82.2} & \underline{53.0} \\
    \rowcolor{appleteal!60}
    \textbf{\ours} & \textbf{88.2} & \textbf{55.8} & \underline{81.9} & \textbf{55.7} \\
    \bottomrule
\end{tabular}

    \vspace{-10pt}
\end{wraptable}

\paragraph{Open-set recognition on WildlifeReID-10k.}
The queries of unknown individuals in WildlifeReID-10k are the distractor queries of the task, and a deployed system rejects them with a threshold on the similarity to the nearest database image. \Cref{tab:wildlife_openset} evaluates this with one threshold per source dataset, set for a $95\%$ true positive rate on its known queries. \ours separates known from unknown queries best and, at the operating point, rejects the largest share of the unknown queries, well ahead of InfoNCE, while the two AP surrogates are on par on the known queries at the same threshold. \ours thus gains on the side of the decision that depends on the similarity scale, while matching mSAP on the side that depends on the per-query ranking.

\paragraph{Backbones on ILIAS.}
\Cref{tab:ilias_backbones} repeats the ILIAS experiment, run with PE-L@336~\citep{bolya2025perception}, with two further frozen backbones, DINOv3 ViT-L/16~\citep{simeoni2025dinov3} and SigLIP2-L/16@512~\citep{tschannen2025siglip}, under the same protocol. The conclusions from the main paper hold for every backbone. \ours has the best \uAP with and without distractors and also the best mAP, so the gain is tied neither to one backbone nor to \uAP alone. With SigLIP2, InfoNCE and \ours are on par without the distractor queries and \ours is clearly ahead with them, the same cross-query pattern as on Met. \uAP falls sharply once distractor queries are added, for every loss and backbone. Score normalization raises $\uAP^{-}$ for two of the three backbones but never raises \uAP with distractors, for any loss or backbone. On ILIAS a post-hoc correction thus helps only in the absence of distractor queries. Therefore, score normalization is not a straightforward solution for every task and setting.

\newpage

\begin{table}[t]
    \centering
    \vspace{-4pt}
    \caption{\textbf{ILIAS with different backbones.} Linear adaptation on frozen features with the protocol of \Cref{sec:distractor_settings} for every backbone. mAP: per-query metric of the benchmark. \uAP: all queries. $\uAP^{-}$: distractor queries removed. SN: after score normalization. \textbf{Best} per backbone in bold, second best \underline{underlined}.}
    \vspace{-8pt}
    \label{tab:ilias_backbones}
    \newcolumntype{C}{>{\centering\arraybackslash}p{22pt}}
\newcolumntype{D}{>{\leavevmode\color{gray!65!black}\centering\arraybackslash}p{22pt}}
\scriptsize
\setlength{\tabcolsep}{4pt}
\begin{tabular}{l l C CC CC}
    \toprule
    \textbf{backbone} & \textbf{method} & mAP & \uAP & $\uAP^{-}$ & $\uAP_{\text{SN}}$ & $\uAP^{-}_{\text{SN}}$ \\ \midrule
     & \textbf{baseline} & 33.2 & 3.5 & 17.1 & 2.7 & 20.3 \\
     & \textbf{InfoNCE} & 33.3 & \underline{7.4} & \underline{22.0} & \textbf{5.4} & \underline{22.7} \\
     & \textbf{mSAP} & \underline{33.4} & 5.6 & 19.6 & 3.9 & 21.2 \\
    \rowcolor{appleteal!60} \cellcolor{white}\multirow{-4}{*}{PE-L@336} & \textbf{\ours} & \textbf{34.6} & \textbf{8.7} & \textbf{22.7} & \underline{4.3} & \textbf{24.4} \\ \midrule
     & \textbf{baseline} & \underline{28.3} & 3.7 & 15.0 & 0.8 & \underline{15.6} \\
     & \textbf{InfoNCE} & 28.0 & \underline{4.1} & \underline{16.5} & 1.2 & 15.0 \\
     & \textbf{mSAP} & 27.0 & 3.2 & 14.1 & \underline{1.3} & 13.7 \\
    \rowcolor{appleteal!60} \cellcolor{white}\multirow{-4}{*}{DINOv3} & \textbf{\ours} & \textbf{29.5} & \textbf{4.5} & \textbf{17.3} & \textbf{1.5} & \textbf{16.4} \\ \midrule
     & \textbf{baseline} & 31.3 & 3.8 & 17.6 & 3.8 & 20.9 \\
     & \textbf{InfoNCE} & \underline{31.8} & \underline{6.1} & \underline{21.0} & \underline{5.0} & \underline{25.7} \\
     & \textbf{mSAP} & 31.6 & 4.7 & 19.3 & 4.6 & 23.6 \\
    \rowcolor{appleteal!60} \cellcolor{white}\multirow{-4}{*}{SigLIP2-L@512} & \textbf{\ours} & \textbf{32.8} & \textbf{7.4} & \textbf{21.1} & \textbf{5.2} & \textbf{26.3} \\
    \bottomrule
\end{tabular}

    \vspace{-5pt}
\end{table}

\section{Evaluation settings}
\label{sec:evaluation}

\subsection{Cross-modal learning}
\label{sec:vlm_settings}

\paragraph{Datasets.}
\textit{Vision--language.} We follow the standard CLIP benchmark suite~\citep{cherti_clip_bench}, covering both cross-modal retrieval and zero-shot recognition across a broad range of domains. For zero-shot retrieval, we evaluate on three image datasets, Flickr8k~\citep{flickr}, Flickr30k~\citep{flickr30k}, and MS COCO Captions~\citep{cococaptions}, and three video datasets, ActivityNet~\citep{activitynet}, MSR-VTT~\citep{msrvtt}, and MSVD~\citep{msvd} (\Cref{tab:sota_vlm_video_retrieval_vitb32}). For zero-shot classification, the benchmark includes generic object and scene recognition datasets such as ImageNet-1k~\citep{deng2009imagenet} and its OOD variants ImageNet-V2~\citep{recht2019imagenetv2}, ImageNet-Sketch~\citep{wang2019learning}, ImageNet-A and ImageNet-O~\citep{hendrycks2021natural}, and ImageNet-R~\citep{hendrycks2021many}; Caltech101~\citep{fei2004learning}; CIFAR-10 and CIFAR-100~\citep{krizhevsky2009learning}; SUN397~\citep{xiao2010sun}; STL10~\citep{coates2011analysis}; DTD~\citep{cimpoi2014describing}; FGVC Aircraft~\citep{maji2013fine}; Flowers-102~\citep{nilsback2008automated}; Oxford-IIIT Pets~\citep{parkhi2012cats}; Stanford Cars~\citep{krause20133d}; GTSRB~\citep{stallkamp2011gtsrb}; SVHN~\citep{netzer2011reading}; VOC2007~\citep{everingham2010pascal} (single- and multi-label); and FER2013~\citep{goodfellow2013challenges}. It further includes robustness, bias, and geographic benchmarks such as ObjectNet~\citep{barbu2019objectnet} and Country211~\citep{country211_openai}; remote-sensing datasets such as EuroSAT~\citep{helber2019eurosat} and RESISC45~\citep{cheng2017remote}; medical datasets such as Diabetic Retinopathy~\citep{kaggle2015dr} and PatchCamelyon~\citep{veeling2018rotation}; and synthetic or structured reasoning benchmarks such as CLEVR~\citep{johnson2017clevr}, dSprites~\citep{matthey2017dsprites}, DeepMind Lab~\citep{beattie2016deepmind}, SmallNORB~\citep{lecun2004learning}, KITTI distance~\citep{geiger2012we}, MNIST~\citep{lecun1998gradient}, and Rendered SST2~\citep{socher2013recursive,radford2021learning}. For benchmarks derived from a common source dataset, such as ImageNet-A/O/R/Sketch or the different attribute splits of CLEVR, dSprites, and SmallNORB, we use the standard task definitions from the CLIP benchmark. Additionally, we evaluate zero-shot classification on three video datasets, Kinetics-400~\citep{kay2017kinetics}, -600~\citep{carreira2018short}, and -700~\citep{carreira2019short}.
\textit{Audio--language.} We fine-tune on the AudioCaps~\citep{kim2019audiocaps} training split (45{,}178 audio--caption pairs) and evaluate zero-shot on datasets not used for fine-tuning: audio--text retrieval on the Clotho~\citep{drossos2020clotho} validation (1{,}045 clips) and evaluation split (1{,}045 clips), and classification on ESC-50~\citep{piczak2015esc} (2{,}000 clips, 50 classes) and UrbanSound8K~\citep{salamon2014urban} (8{,}732 clips, 10 classes).

\paragraph{Implementation details.} \textit{Vision--language.} All experiments are implemented in OpenCLIP~\citep{ilharco_gabriel_2021_5143773,cherti2023reproducible}\footnote{\url{https://github.com/mlfoundations/open_clip}}. Our main experiments fine-tune the released OpenAI CLIP ViT-B/32 checkpoint~\citep{radford2021learning}; in the appendix we additionally fine-tune the OpenAI ViT-B/16~\citep{radford2021learning} and the OpenCLIP ViT-B/32 trained on LAION-400M~\citep{schuhmann2021laion,cherti2023reproducible}. Fine-tuning is conducted on CC12M~\citep{changpinyo2021conceptual} (10.97M image--text pairs, one pass per epoch without resampling), for 5 epochs with learning rate $10^{-5}$, 2{,}000 warm-up steps followed by cosine decay, on 2 H200 GPUs (141\,GB each). The global batch size is 1{,}792 for ViT-B/16 and 2{,}304 for ViT-B/32. We adopt the default OpenCLIP optimization setup: AdamW~\citep{loshchilov2017decoupled} with weight decay 0.2, $(\beta_1,\beta_2)=(0.9,0.98)$ and $\epsilon=10^{-6}$, fp16 automatic mixed precision, no gradient clipping, and no gradient accumulation, with the default OpenCLIP image preprocessing at $224\times224$. Both image and text encoders are updated jointly, without tower freezing, locking, or architectural changes, and we evaluate the final checkpoint without checkpoint selection. We compare three losses: the original CLIP InfoNCE loss~\citep{oord2018representation,radford2021learning}, the SigLIP sigmoid loss~\citep{zhai2023sigmoid}, and our \ours loss. In all three, the logit scale is learnable and initialized from the pretrained checkpoint ($t=100$); for SigLIP, the learnable bias is initialized to $0$, since CLIP checkpoints do not provide one. For \ours, we apply our differentiable surrogate to the same gathered, scaled similarity matrix, with $\tau=10^{-3}$ and no negative dropping unless stated otherwise.
\textit{Audio--language.} We fine-tune the released LAION-CLAP~\citep{wu2023large}\footnote{\url{https://github.com/LAION-AI/CLAP}} checkpoint, which pairs an HTS-AT base audio encoder~\citep{chen2022htsat} with a RoBERTa text encoder~\citep{liu2019roberta} and is pretrained on music, speech, and AudioSet~\citep{gemmeke2017audioset}. Both towers are fine-tuned jointly for 10 epochs. Audio is resampled to 48\,kHz and cropped or zero-padded to 10\,s, without feature fusion. We use AdamW with learning rate $10^{-5}$, weight decay 0.2, $(\beta_1,\beta_2)=(0.9,0.98)$, and $\epsilon=10^{-6}$, 60 warm-up steps followed by cosine decay, gradient clipping at 1.0, bf16 mixed precision, and a global batch size of 1{,}024. The logit scale is learnable and initialized to $1/0.07$; for SigLIP, the bias is initialized to $-10$.
\textit{Both domains.} In all cases, we use distributed gathering so that the loss is computed from similarities over all samples in the global batch. Reported numbers are the mean over three seeds. 

\paragraph{Evaluation protocol.} \textit{Vision--language.} Image benchmarks follow the CLIP benchmark~\citep{cherti_clip_bench}\footnote{\url{https://github.com/LAION-AI/CLIP_benchmark}} with its default class names and prompt templates. For video, we encode a single frame per video: the frame at 3\,s for MSR-VTT, MSVD, and Kinetics, and the temporal midpoint for the much longer ActivityNet videos. Kinetics class embeddings average four action prompts (``a photo of a person \{\}.'', ``a video of a person \{\}.'', ``a photo of people \{\}.'', ``a video of \{\}.''). We use the full MSR-VTT test split (2{,}990 videos, 59{,}800 captions), the 436 available MSVD test videos (18{,}098 captions), and paragraph-to-video retrieval on ActivityNet val\_1, where each of the 1{,}063 available videos is paired with the concatenation of its captions. Kinetics-400, -600, and -700 are evaluated on their validation sets (19{,}881, 29{,}739, and 33{,}966 clips; 400, 599, and 700 classes). For retrieval we report gAP, i.e., average precision over the flattened query--item similarity matrix, together with Recall@$k$; for classification, top-1/top-5 accuracy and mean per-class recall. \textit{Audio--language.} For retrieval, each Clotho clip is paired with its first reference caption, and we report R@1 in both directions against the full gallery. For classification, class embeddings average four prompts (``the sound of \{\}'', ``a sound of \{\}'', ``an audio recording of \{\}'', ``\{\} sounds''), and we report accuracy over all clips of each dataset.

\subsection{Self-supervised learning}
\label{sec:ssl_settings}

\paragraph{Datasets.}
Pretraining uses ImageNet-1k~\citep{deng2009imagenet} without labels. In-distribution accuracy is measured on the ImageNet-1k validation set; distribution shift on ImageNet-V2~\citep{recht2019imagenetv2}, ImageNet-R~\citep{hendrycks2021many} and ImageNet-Sketch~\citep{wang2019learning}, reported as their mean; corruption robustness on ImageNet-C~\citep{hendrycks2019benchmarking}, 15 corruptions at 5 severities. Transfer uses ten datasets: six fine-grained ones---FGVC Aircraft~\citep{maji2013fine}, CUB-200~\citep{wah2011caltech}, Stanford Cars~\citep{krause20133d}, Oxford-IIIT Pets~\citep{parkhi2012cats}, Flowers-102~\citep{nilsback2008automated} and Food-101~\citep{bossard2014food}---and four coarse-grained object, scene and texture sets---CIFAR-10 and CIFAR-100~\citep{krizhevsky2009learning}, DTD~\citep{cimpoi2014describing} and SUN397~\citep{xiao2010sun}.

\paragraph{Implementation details.}
We take the MoCo~v3~\citep{chen2021empirical} recipe unchanged\footnote{\url{https://github.com/facebookresearch/moco-v3}}. We train a ViT-S/16 and a ViT-B/16 on ImageNet-1k for 300 epochs with batch size 4{,}096 and learning rate $4\cdot10^{-3}$. We replace the InfoNCE loss by \ours computed over the same gathered query--key similarity matrix, with the positives on the diagonal and temperature $\tau=0.2$. Nothing else changes: augmentations, momentum encoder, projection and prediction heads, optimizer and schedule are those of the official code. The baselines are the released 300-epoch MoCo~v3 checkpoints with their published linear heads. Re-training the official recipe reproduces them to within seed variance.

\paragraph{Evaluation protocol.}
All numbers are top-1 accuracy on frozen features. \emph{Linear probing} (LP) on ImageNet-1k, its OOD variants and ImageNet-C uses the official 90-epoch linear head on the CLS features of the frozen backbone (the published MoCo~v3 heads for the baselines and heads trained with the identical recipe for \ours). Every OOD set is classified with the ImageNet-1k head and no dataset-specific training. ImageNet-R covers 200 classes and the softmax is restricted to them. Transfer datasets have no published head, so LP there follows prior work~\citep{kornblith2019better,ericsson2021how}: a linear classifier per dataset on frozen features, with the learning rate and weight decay chosen on a held-out split and the classifier refit on train and validation. Mean of three seeds is reported, with seed standard deviation $\le0.11$ on every set. \emph{$k$-Nearest Neighbors} ($k$NN) is a weighted vote over the $k{=}20$ nearest training features, with $k$ fixed and not tuned (cosine similarity, weights $\exp(s/0.07)$) on the full 1.28M-image ImageNet-1k train set, or on each transfer set's own training split. Neither the choice of $k$ nor of the linear head affects any comparison: $k{=}20$ is within $0.15$ of every model's best $k\in\{5,\dots,200\}$ with the same model ranking at every $k$, and the official head is $0.2$--$0.7$ points above a short probe on the same features, again with an identical ranking.

\paragraph{Audio.}
For audio we follow CLMR~\citep{spijkervet2021contrastive} unchanged\footnote{\url{https://github.com/Spijkervet/CLMR}}: a SampleCNN encoder~\citep{lee2017sample} is pretrained on the raw waveforms of MagnaTagATune~\citep{law2009evaluation} with the augmentation chain, batch size, optimizer and schedule of the official code. We replace InfoNCE by \ours on the same similarity matrix between the two augmented views of a batch, with temperature $\tau=0.05$, and keep every other setting of the InfoNCE run. Evaluation is linear probing on the frozen encoder for the 50 tags of MagnaTagATune, scored by ROC-AUC and PR-AUC. We pretrain three seeds per loss, each with its own probe, and report the mean.

\subsection{Metric learning}
\label{sec:metric_learning_impl}

\paragraph{Datasets.}
We conduct experiments on two widely used datasets: Stanford Online Products (SOP)~\citep{oh2016deep} and iNaturalist~\citep{van2018inaturalist}. We follow the dataset split used in prior work~\citep{brown2020smooth,patel2022recall}. SOP consists of products. The training split has 59,551 images from 11,318 classes, and the test split contains 60,502 images from 11,316 classes. iNaturalist consists of natural images of wildlife and plants. The training split has 325,846 images across 5,690 classes, while the test split contains 136,093 images across 2,452 classes. For SOP, all test images serve as queries. Due to the large size of the iNaturalist test set, for mAP and \uAP computation, we sample 4 images per test class as queries, resulting in 9,808 queries.

\paragraph{Evaluation protocol.} For all runs, we tune hyper-parameters with Optuna~\citep{akiba2019optuna}, optimizing \uAP. Tuning on a different metric, \ie mAP, yields the same results in several verification runs. Specifically, we split the training set into 70\% for training and 30\% for validation and run 10 trials to search for the learning rate $\eta\in[10^{-5}, 5\cdot10^{-4}]$ and the temperature $\tau\in[10^{-4}, 5\cdot10^{-2}]$. For the other metric learning losses, we tune their main hyper-parameter and keep the rest fixed to the recommended settings of the original papers, and for the regularizers of \Cref{tab:calibration} we also tune their weight. After tuning, the final model is trained with the selected hyper-parameters on the whole training set and evaluated on the test set. The full process, tuning, training and evaluation, is run for three random seeds.

\paragraph{Implementation details.} For the comparison of metric learning losses, we use the ViT-S/16 variant of the DINOv3~\citep{simeoni2025dinov3} vision encoders. Our implementation is in PyTorch~\citep{paszke2019pytorch} and builds on the open-source repository of Recall@$k$\footnote{\url{https://github.com/yashvarpatel/RecallatK_surrogate}}. Unless specified otherwise, we use a batch size of 1{,}000 and a chunk size of 200 for the multi-stage backpropagation of \citet{revaud2019learning,patel2022recall}. We train for 100 epochs with the AdamW~\citep{loshchilov2017decoupled} optimizer and fix the embedding dimension $d$ to 512 in all experiments. For the comparison with the state of the art, we experiment with ResNet50~\citep{he2016deep} and ViT-B/16~\citep{dosovitskiy2020image}, trained for 200 epochs following the implementation details of~\citet{patel2022recall}. Experiments are conducted on Nvidia A100 and H200 GPUs with 40\,GB and 141\,GB of memory, respectively. Each run is executed on a single GPU.

\newpage

\paragraph{Thresholding protocol.}
For \Cref{tab:threshold} and \Cref{fig:tauq}, the test classes are split in half, identically for every model, into \emph{fitting} classes, used only to select a threshold, and \emph{evaluation} classes, on which it is applied. The evaluation queries are the test images of the evaluation classes, and the gallery is the whole test split minus the query. On the fitting classes we select one threshold for maximal F1 and one as the lowest threshold with precision $0.90$. F1$_g$ is the mean F1 of the evaluation queries at the first, and R@P90 the recall over all evaluation pairs at the second. F1$_q$ uses the F1-optimal threshold of each evaluation query instead. Thresholds are searched with a step of $5\times10^{-4}$.

\paragraph{Scarce-data protocol.}
For \Cref{fig:scarce_data_uap}, we simulate limited annotations. From the training classes with at least $4$ samples, we keep a random $\gamma\%$ of the classes and draw $\rho$ images per class, which gives a reduced dataset $\cD^{\gamma,\rho}$. Its classes are split into $\cD^{\gamma,\rho}_{\text{train}}$ and $\cD^{\gamma,\rho}_{\text{val}}$ for hyper-parameter tuning, the model with the best validation performance at the end of training is selected, and the final model is trained with these hyper-parameters on all classes of $\cD^{\gamma,\rho}$ and evaluated on the full test set. We use $\gamma \in \{1, 5, 10, 25, 50, 100\}$ and $\rho \in \{2, 4\}$, and run the whole process with three random seeds for each pair.

\subsection{Tasks with distractor queries}
\label{sec:distractor_settings}

\paragraph{Datasets.}
DISC21~\citep{douze2021image} is an image copy detection benchmark: queries are edited copies of reference images, and the query set is dominated by distractor queries that have no source in the reference set. The Met dataset~\citep{ypsilantis2021met} is instance-level recognition of artworks, where a query is either a photograph of an exhibit in the database or a distractor photograph of no exhibit. ILIAS~\citep{kordopatis2025ilias} is instance-level image retrieval at scale, retrieving images of particular object instances from a database dominated by distractor images. WildlifeReID-10k~\citep{adam2024wildlifereid} is re-identification of individual animals across many species. Each task keeps the training and evaluation code of its official repository, with only the loss replaced.

\paragraph{Implementation details.}
Unless tuned on the validation split of a task, \ours uses $\tau=0.001$ and negative dropping with $\alpha=90\%$, and every task trains with batches of $4{,}096$ samples. All tuned hyper-parameters are selected by grid search. InfoNCE and mSAP are trained with exactly the same recipes and their hyper-parameters are tuned in the same way. All results are averaged over three random seeds. On DISC21 we follow the SSCD repository\footnote{\url{https://github.com/facebookresearch/sscd-copy-detection}} and fine-tune a DINOv3 ViT-B/16 at $224$~px for $30$ epochs with AdamW. The absolute learning rate, the temperature $\tau$ and the weight $\lambda$ of the entropy regularizer of SSCD are tuned on the dev queries. On Met we follow the official repository\footnote{\url{https://github.com/nikosips/met}} and fine-tune a DINOv3 ViT-B/16 at $512$~px for $10$ epochs with AdamW, one warm-up epoch and cosine decay. The learning rate and the temperature $\tau$ are tuned on the validation set, as are the $k$ and the temperature of the $k$NN classifier, in accordance with the original work. On ILIAS we follow the official repository\footnote{\url{https://github.com/ilias-vrg/ilias}} and train a linear adaptation layer on UnED~\citep{ypsilantis2023towards}, on top of publicly available extracted features. We report PE-L@336~\citep{bolya2025perception} for the main comparison as it achieves the best performance. We follow the default training settings from the original work, with the only difference that we draw four samples per class in every batch. The DINOv3-L and SigLIP2-L@512 runs of \Cref{tab:ilias_backbones} use exactly the same protocol. On WildlifeReID-10k we follow the WildlifeDatasets toolkit~\citep{cermak2024wildlifedatasets}\footnote{\url{https://github.com/WildlifeDatasets/wildlife-tools}} and fine-tune a DINOv3 ViT-B/16 at $224$~px for $100$ epochs with AdamW, learning rate $5\cdot10^{-5}$ and $10$ warm-up epochs, again with four samples per class in every batch.  Score normalization (SN)~\citep{pizzi2022self} reduces every similarity of a query by $\beta$ times the mean similarity of the query to its $k$ nearest neighbors in a background database. On DISC21 we use the setting of the original work, $\beta=1$ and $k=3$ with the DISC21 training set as background, and on ILIAS $\beta=0.4$ and $k=1$ with UnED as background.

\paragraph{Baselines.}
On DISC21 the baseline is the SSCD-large model~\citep{pizzi2022self}, on Met the R50-SWSL backbone with a $k$NN classifier, the best result reported in the original work~\citep{ypsilantis2021met}, on ILIAS the method of the original work~\citep{kordopatis2025ilias}, and on WildlifeReID-10k MiewID-msv3~\citep{otarashvili2024multispecies}, the reference model of the benchmark.

\paragraph{Metrics.}
\uAP is computed over all query--database pairs, including those of the distractor queries, and $\uAP^{-}$ after removing the distractor queries. The relative drop $(\uAP^{-}-\uAP)/\uAP^{-}$ is the share of \uAP lost to their negatives. Per-query metrics (accuracy, mAP, and the balanced accuracy on known samples, BAKS) are those of each benchmark and ignore the distractor queries. ILIAS has no distractor queries of its own, so we simulate them. Every query retrieves its top-$1{,}000$ images, and queries are processed in chunks of $100$. For a chunk, $\uAP^{-}$ ranks the lists of its queries, and \uAP the lists of all queries restricted to the positives of the chunk's instances and the YFCC100M distractors, so that every query outside the chunk acts as a distractor query. Both are averaged over the chunks. On WildlifeReID-10k the distractor queries are those of individuals with no image in the database, and we also report open-set metrics, where such a query has to be rejected by a threshold on its top-1 similarity. On DISC21 we also report R@P90, the recall at the global threshold that gives $90\%$ precision.

\section{Detailed results}
\label{sec:det_results}

\paragraph{Detailed results.} \Cref{tab:full_benchmark_vitb32,tab:full_benchmark_vitb16} report per-dataset zero-shot classification results on the full CLIP benchmark~\citep{cherti_clip_bench} for ViT-B/32 and ViT-B/16, respectively: top-1, top-5, and mean per-class recall (mR) on 38 tasks (37 classification datasets, with VOC2007 evaluated as single- and multi-label), each averaged over three runs. Results on the three image--text retrieval benchmarks are reported in \Cref{tab:sota_vlm_image_retrieval_backbones,tab:sota_vlm_image_retrieval_vitb32}. Fine-tuning with \ours yields the strongest overall performance on both backbones. Averaged over all 108 classification scores, it improves over the pretrained model, the CLIP InfoNCE loss, and SigLIP by 1.16, 0.49, and 0.64 for ViT-B/32, and by 1.43, 0.65, and 0.77 points for ViT-B/16. On retrieval, the task that \ours directly optimizes, the gains over the pretrained model and InfoNCE are considerably larger. Where \ours is not the best, two patterns account for most cases. First, on some datasets fine-tuning on CC12M brings little or no benefit regardless of the loss, and the pretrained model remains competitive (\eg ObjectNet, MNIST); the structured dSprites and SmallNORB tasks also stay close to chance level for all models. Second, CLIP or SigLIP occasionally perform slightly better on a specific metric (\eg CLEVR distance, Flowers, or Pets), but the differences are typically small. In contrast, cases where \ours underperforms by a clear margin are rare.

\begin{table}[t]
\centering
\footnotesize
\setlength{\tabcolsep}{1.4pt}
\newcolumntype{O}{>{\columncolor{appleteal!60}}c}
\begin{tabular}{l cccO cccO cccO}
\toprule
& \multicolumn{4}{c}{\textbf{top-1}} & \multicolumn{4}{c}{\textbf{top-5}} & \multicolumn{4}{c}{\textbf{mean per-class recall}} \\
\cmidrule(lr){2-5} \cmidrule(lr){6-9} \cmidrule(lr){10-13}
\textbf{dataset} & \textbf{Pre.} & \textbf{CLIP} & \textbf{SigLIP} & \textbf{\ours} & \textbf{Pre.} & \textbf{CLIP} & \textbf{SigLIP} & \textbf{\ours} & \textbf{Pre.} & \textbf{CLIP} & \textbf{SigLIP} & \textbf{\ours} \\
\midrule
\multicolumn{13}{l}{\textit{ImageNet and robustness}} \\
\quad ImageNet-1k & 59.6 & 60.7 & 61.3 & \textbf{61.8} & 86.4 & 87.4 & 87.8 & \textbf{87.9} & 59.6 & 60.7 & 61.3 & \textbf{61.8} \\
\quad ImageNet-V2 & 52.9 & 53.6 & 54.1 & \textbf{54.3} & 80.8 & 82.0 & 82.1 & \textbf{82.5} & 52.9 & 53.6 & 54.2 & \textbf{54.3} \\
\quad ImageNet-Sketch & 38.9 & 42.3 & 43.2 & \textbf{43.7} & 66.9 & 70.4 & 71.6 & \textbf{71.7} & 39.0 & 42.3 & 43.3 & \textbf{43.7} \\
\quad ImageNet-A & 28.1 & 27.8 & \textbf{28.4} & 27.9 & 60.3 & 60.0 & \textbf{60.7} & 60.1 & \textbf{29.4} & 29.1 & \textbf{29.4} & 29.0 \\
\quad ImageNet-O & 44.9 & \textbf{48.1} & 47.8 & \textbf{48.1} & 75.0 & 77.6 & \textbf{78.3} & 78.2 & 46.1 & \textbf{49.5} & 48.8 & 49.3 \\
\quad ImageNet-R & 67.0 & 68.3 & 69.1 & \textbf{69.4} & 87.2 & 87.9 & 88.2 & \textbf{88.3} & 64.9 & 66.8 & 67.8 & \textbf{68.1} \\
\quad ObjectNet & \textbf{43.4} & 42.5 & 41.8 & 42.2 & \textbf{69.4} & 68.3 & 68.3 & 68.5 & \textbf{41.7} & 41.6 & 41.0 & 41.3 \\
\quad Country211 & \textbf{15.7} & 15.0 & 15.3 & \textbf{15.7} & \textbf{37.7} & 36.2 & 37.0 & 37.4 & \textbf{15.7} & 15.0 & 15.3 & \textbf{15.7} \\
\midrule
\multicolumn{13}{l}{\textit{Generic objects and scenes}} \\
\quad Caltech101 & 80.5 & 82.5 & 83.0 & \textbf{83.2} & 94.0 & 96.1 & 96.1 & \textbf{96.7} & 85.7 & 86.6 & 87.8 & \textbf{88.0} \\
\quad CIFAR-10 & 86.5 & 89.3 & 89.4 & \textbf{90.0} & 99.2 & 99.6 & 99.5 & \textbf{99.7} & 86.5 & 89.4 & 89.4 & \textbf{90.0} \\
\quad CIFAR-100 & 62.4 & 64.8 & 65.8 & \textbf{66.0} & 87.1 & 88.3 & \textbf{89.2} & \textbf{89.2} & 62.4 & 64.8 & 65.8 & \textbf{66.0} \\
\quad SUN397 & 61.4 & 64.7 & 64.9 & \textbf{65.9} & 89.9 & 92.1 & 92.3 & \textbf{92.7} & 61.3 & 64.6 & 65.1 & \textbf{65.5} \\
\quad STL10 & 95.7 & 96.6 & 96.5 & \textbf{96.9} & 99.8 & \textbf{100.0} & \textbf{100.0} & \textbf{100.0} & 95.7 & 96.6 & 96.5 & \textbf{96.9} \\
\quad DTD & 41.5 & 44.8 & 44.6 & \textbf{46.0} & 73.0 & 76.6 & 77.3 & \textbf{77.4} & 41.5 & 44.7 & 44.6 & \textbf{46.0} \\
\quad FGVC Aircraft & \textbf{16.0} & 13.5 & 14.2 & 13.6 & 42.3 & 41.3 & \textbf{43.2} & 41.8 & \textbf{16.0} & 13.4 & 14.1 & 13.6 \\
\quad Flowers-102 & 60.2 & 59.2 & 59.4 & \textbf{61.2} & 82.7 & 83.7 & \textbf{84.0} & \textbf{84.0} & \textbf{60.4} & 57.8 & 58.4 & 60.1 \\
\quad Oxford-IIIT Pets & 84.0 & \textbf{85.5} & 85.0 & 85.1 & 98.0 & \textbf{99.3} & \textbf{99.3} & \textbf{99.3} & 83.7 & \textbf{85.4} & 84.9 & 84.9 \\
\quad Stanford Cars & 51.5 & 50.8 & 52.8 & \textbf{53.1} & 86.0 & 87.3 & 88.6 & \textbf{89.1} & 51.8 & 50.7 & 52.6 & \textbf{53.0} \\
\quad GTSRB & 26.6 & 30.8 & 31.9 & \textbf{33.6} & 65.1 & 63.8 & 65.1 & \textbf{67.5} & 25.9 & 27.7 & 26.8 & \textbf{30.1} \\
\quad SVHN & 9.4 & 23.6 & 19.7 & \textbf{24.6} & 61.8 & 69.7 & 70.7 & \textbf{72.6} & 13.2 & 18.7 & 20.3 & \textbf{20.6} \\
\quad VOC2007 & 74.1 & 75.4 & 75.1 & \textbf{76.6} & \textbf{95.1} & 94.2 & 93.7 & 94.8 & 78.2 & \textbf{79.3} & 78.9 & 79.0 \\
\quad VOC2007 (multi-label)$^\dagger$ & 75.1 & 81.6 & \textbf{82.1} & 80.6 & -- & -- & -- & -- & -- & -- & -- & -- \\
\quad FER2013 & 38.9 & \textbf{41.7} & 39.6 & 40.6 & 88.9 & 93.3 & 91.7 & \textbf{94.6} & 35.9 & 38.4 & \textbf{38.8} & 37.5 \\
\midrule
\multicolumn{13}{l}{\textit{Remote sensing and medical}} \\
\quad EuroSAT & 45.3 & 52.4 & 49.8 & \textbf{52.7} & 90.1 & 88.5 & 90.1 & \textbf{90.6} & 44.1 & 52.3 & 48.9 & \textbf{52.6} \\
\quad RESISC45 & 50.9 & 50.7 & \textbf{51.0} & 50.0 & 84.5 & 84.5 & \textbf{85.5} & 84.8 & 51.5 & 51.4 & \textbf{51.7} & 50.8 \\
\quad Diabetic Retinopathy & 10.2 & \textbf{20.5} & 18.4 & 8.0 & -- & -- & -- & -- & 21.8 & \textbf{23.9} & 21.4 & 22.4 \\
\quad PatchCamelyon & \textbf{58.3} & 55.5 & 53.9 & 56.3 & -- & -- & -- & -- & \textbf{58.3} & 55.6 & 53.9 & 56.3 \\
\midrule
\multicolumn{13}{l}{\textit{Synthetic and structured}} \\
\quad CLEVR (count) & \textbf{25.2} & 19.3 & 18.6 & 19.2 & \textbf{80.7} & 70.7 & 70.5 & 73.0 & \textbf{24.8} & 19.1 & 18.3 & 18.8 \\
\quad CLEVR (distance) & 15.7 & \textbf{16.5} & 15.8 & 16.3 & 92.2 & 89.8 & 87.5 & \textbf{94.2} & 16.3 & 16.1 & \textbf{16.7} & 16.6 \\
\quad dSprites (orientation) & \textbf{2.4} & 1.9 & 1.9 & 2.1 & 10.6 & 11.1 & 10.5 & \textbf{11.2} & \textbf{2.4} & 1.9 & 1.9 & 2.1 \\
\quad dSprites ($x$ position) & \textbf{3.4} & 3.0 & 3.3 & 3.1 & 15.7 & 15.4 & \textbf{16.2} & 14.9 & \textbf{3.4} & 3.0 & 3.2 & 3.1 \\
\quad dSprites ($y$ position) & \textbf{3.2} & \textbf{3.2} & \textbf{3.2} & \textbf{3.2} & \textbf{15.6} & \textbf{15.6} & 15.5 & 15.4 & \textbf{3.1} & \textbf{3.1} & \textbf{3.1} & \textbf{3.1} \\
\quad DMLab & \textbf{18.5} & 17.9 & 16.2 & 17.5 & 84.0 & 84.0 & 82.3 & \textbf{84.5} & 17.0 & \textbf{17.7} & 16.6 & \textbf{17.7} \\
\quad KITTI (distance) & 22.8 & \textbf{27.5} & 24.1 & 26.3 & -- & -- & -- & -- & 35.7 & 34.4 & \textbf{35.8} & 34.1 \\
\quad SmallNORB (azimuth) & \textbf{6.6} & 5.7 & 5.6 & 5.8 & \textbf{28.4} & 27.8 & 28.2 & 27.6 & \textbf{6.7} & 5.8 & 5.7 & 5.9 \\
\quad SmallNORB (elevation) & 11.1 & 12.9 & 11.8 & \textbf{13.1} & 54.1 & \textbf{58.1} & 57.4 & 57.2 & 11.0 & 12.8 & 11.7 & \textbf{12.9} \\
\quad MNIST & \textbf{50.9} & 30.7 & 30.7 & 33.0 & 85.9 & 83.0 & 78.6 & \textbf{87.0} & \textbf{50.6} & 31.4 & 31.2 & 34.2 \\
\quad Rendered SST2 & 59.3 & \textbf{61.1} & 56.6 & 59.3 & -- & -- & -- & -- & 59.2 & \textbf{61.1} & 56.6 & 59.2 \\
\midrule
Average & 42.1 & 43.2 & 42.8 & \textbf{43.3} & 71.8 & 72.2 & 72.3 & \textbf{73.2} & 42.0 & 42.3 & 42.2 & \textbf{42.8} \\
\bottomrule
\end{tabular}

\vspace{-8pt}
\caption{\textbf{Zero-shot classification on the full CLIP benchmark with ViT-B/32.} Top-1, top-5, and mean per-class recall of the pretrained OpenAI CLIP ViT-B/32 (Pre.) and of its fine-tuning on CC12M with the CLIP (InfoNCE), SigLIP, and \ours losses; mean over three seeds. $^\dagger$Mean average precision in place of top-1. Top-5 is omitted (--) for tasks with five or fewer classes. Averages are taken over the tasks with a value in each group (38, 33, and 37).}
\label{tab:full_benchmark_vitb32}
\end{table}

\begin{table}[t]
\centering
\footnotesize
\setlength{\tabcolsep}{1.4pt}
\newcolumntype{O}{>{\columncolor{appleteal!60}}c}
\begin{tabular}{l cccO cccO cccO}
\toprule
& \multicolumn{4}{c}{\textbf{top-1}} & \multicolumn{4}{c}{\textbf{top-5}} & \multicolumn{4}{c}{\textbf{mean per-class recall}} \\
\cmidrule(lr){2-5} \cmidrule(lr){6-9} \cmidrule(lr){10-13}
\textbf{dataset} & \textbf{Pre.} & \textbf{CLIP} & \textbf{SigLIP} & \textbf{\ours} & \textbf{Pre.} & \textbf{CLIP} & \textbf{SigLIP} & \textbf{\ours} & \textbf{Pre.} & \textbf{CLIP} & \textbf{SigLIP} & \textbf{\ours} \\
\midrule
\multicolumn{13}{l}{\textit{ImageNet and robustness}} \\
\quad ImageNet-1k & 64.4 & 65.3 & 66.3 & \textbf{66.6} & 89.7 & 90.7 & 91.1 & \textbf{91.2} & 64.4 & 65.3 & 66.3 & \textbf{66.6} \\
\quad ImageNet-V2 & 57.8 & 58.8 & 59.4 & \textbf{60.0} & 85.0 & 86.1 & 86.5 & \textbf{86.6} & 57.8 & 58.8 & 59.5 & \textbf{60.1} \\
\quad ImageNet-Sketch & 44.2 & 47.2 & 48.3 & \textbf{49.0} & 72.1 & 75.7 & 76.8 & \textbf{77.0} & 44.3 & 47.3 & 48.4 & \textbf{49.0} \\
\quad ImageNet-A & 44.5 & 44.3 & \textbf{44.8} & 44.7 & 75.3 & 76.1 & \textbf{76.9} & \textbf{76.9} & \textbf{43.4} & 42.7 & 43.2 & 42.5 \\
\quad ImageNet-O & 41.0 & 41.9 & 42.3 & \textbf{42.6} & 67.8 & 72.7 & 72.8 & \textbf{73.5} & 42.9 & 43.1 & \textbf{43.9} & \textbf{43.9} \\
\quad ImageNet-R & 73.5 & 76.0 & 77.0 & \textbf{77.1} & 91.0 & 91.9 & \textbf{92.3} & \textbf{92.3} & 71.7 & 74.1 & \textbf{75.4} & \textbf{75.4} \\
\quad ObjectNet & \textbf{54.7} & 51.6 & 51.3 & 51.0 & \textbf{78.3} & 76.2 & 76.1 & 76.3 & \textbf{52.3} & 50.6 & 50.5 & 50.1 \\
\quad Country211 & 20.2 & 19.5 & 20.1 & \textbf{20.3} & 44.5 & 44.1 & 44.8 & \textbf{45.0} & 20.2 & 19.5 & 20.1 & \textbf{20.3} \\
\midrule
\multicolumn{13}{l}{\textit{Generic objects and scenes}} \\
\quad Caltech101 & 81.7 & 82.8 & 83.5 & \textbf{84.0} & 95.4 & 95.7 & 96.6 & \textbf{97.0} & 86.9 & 88.3 & \textbf{88.8} & 88.5 \\
\quad CIFAR-10 & 88.2 & 91.7 & 91.2 & \textbf{92.0} & 98.9 & \textbf{99.7} & 99.5 & \textbf{99.7} & 88.2 & 91.7 & 91.2 & \textbf{92.0} \\
\quad CIFAR-100 & 62.0 & 68.4 & 68.5 & \textbf{70.6} & 85.4 & 90.2 & 90.5 & \textbf{91.4} & 62.0 & 68.4 & 68.5 & \textbf{70.6} \\
\quad SUN397 & 63.2 & 64.5 & 65.9 & \textbf{66.4} & 91.3 & 92.7 & 93.5 & \textbf{93.6} & 62.5 & 65.8 & 66.7 & \textbf{67.5} \\
\quad STL10 & 97.6 & 97.7 & 97.6 & \textbf{97.9} & \textbf{100.0} & \textbf{100.0} & 99.9 & 99.9 & 97.6 & 97.7 & 97.6 & \textbf{97.9} \\
\quad DTD & 40.4 & 46.6 & 46.3 & \textbf{47.3} & 71.6 & 75.5 & 76.0 & \textbf{76.2} & 40.4 & 46.6 & 46.3 & \textbf{47.3} \\
\quad FGVC Aircraft & \textbf{20.0} & 17.5 & 18.5 & 19.3 & 49.0 & 47.3 & 49.9 & \textbf{50.7} & \textbf{20.0} & 17.5 & 18.5 & 19.2 \\
\quad Flowers-102 & 65.3 & 62.0 & 64.8 & \textbf{65.5} & 82.2 & 83.5 & \textbf{84.9} & 84.1 & 63.9 & 61.1 & 62.7 & \textbf{64.1} \\
\quad Oxford-IIIT Pets & 85.2 & 88.2 & \textbf{89.7} & 89.4 & 98.8 & \textbf{99.5} & \textbf{99.5} & 99.4 & 85.0 & 88.1 & \textbf{89.7} & 89.4 \\
\quad Stanford Cars & 57.5 & 57.7 & 59.1 & \textbf{59.9} & 91.2 & 92.0 & 92.2 & \textbf{92.6} & 57.5 & 57.4 & 58.9 & \textbf{59.7} \\
\quad GTSRB & 41.1 & 45.8 & 43.5 & \textbf{46.7} & 65.9 & 71.7 & 72.8 & \textbf{74.1} & 33.8 & 35.8 & 36.8 & \textbf{38.7} \\
\quad SVHN & \textbf{38.2} & 29.1 & 22.2 & 30.4 & \textbf{77.6} & 69.1 & 65.6 & 73.5 & \textbf{37.8} & 29.3 & 25.5 & 30.0 \\
\quad VOC2007 & \textbf{78.0} & 76.0 & 75.7 & 75.3 & \textbf{95.5} & 91.4 & 92.7 & 92.3 & \textbf{81.0} & 79.5 & 79.1 & 80.5 \\
\quad VOC2007 (multi-label)$^\dagger$ & 78.5 & 83.7 & \textbf{83.9} & \textbf{83.9} & -- & -- & -- & -- & -- & -- & -- & -- \\
\quad FER2013 & 43.6 & 47.1 & \textbf{47.7} & 46.9 & \textbf{96.0} & 93.8 & 92.0 & 92.9 & 40.0 & 41.5 & \textbf{43.8} & 42.9 \\
\midrule
\multicolumn{13}{l}{\textit{Remote sensing and medical}} \\
\quad EuroSAT & 43.0 & 50.8 & \textbf{55.5} & 54.9 & 86.5 & 93.4 & \textbf{94.2} & 93.6 & 43.3 & 49.8 & \textbf{54.8} & 53.5 \\
\quad RESISC45 & 53.5 & 56.3 & 56.2 & \textbf{57.7} & 85.9 & 90.2 & 89.9 & \textbf{90.9} & 54.1 & 56.7 & 56.6 & \textbf{58.3} \\
\quad Diabetic Retinopathy & 3.8 & \textbf{52.7} & 28.1 & 32.5 & -- & -- & -- & -- & 20.2 & \textbf{28.7} & 26.1 & 26.1 \\
\quad PatchCamelyon & 60.2 & \textbf{61.1} & 52.3 & 58.3 & -- & -- & -- & -- & 60.2 & \textbf{61.1} & 52.3 & 58.3 \\
\midrule
\multicolumn{13}{l}{\textit{Synthetic and structured}} \\
\quad CLEVR (count) & 20.8 & 19.9 & 19.5 & \textbf{21.3} & \textbf{79.8} & 71.5 & 69.4 & 76.0 & 20.4 & 19.6 & 19.1 & \textbf{20.9} \\
\quad CLEVR (distance) & 15.9 & \textbf{16.1} & 15.7 & 16.0 & \textbf{80.1} & 80.0 & \textbf{80.1} & 80.0 & 16.5 & \textbf{17.7} & 16.7 & 17.3 \\
\quad dSprites (orientation) & 2.1 & \textbf{2.7} & 2.4 & 2.4 & 10.8 & 11.8 & \textbf{13.1} & 11.9 & 2.0 & \textbf{2.7} & 2.4 & 2.5 \\
\quad dSprites ($x$ position) & \textbf{3.4} & 3.1 & 3.1 & 3.1 & \textbf{16.4} & 15.4 & 16.1 & 15.4 & \textbf{3.3} & 3.0 & 3.1 & 3.1 \\
\quad dSprites ($y$ position) & \textbf{3.1} & \textbf{3.1} & \textbf{3.1} & \textbf{3.1} & 15.5 & 15.6 & \textbf{16.0} & 15.6 & \textbf{3.1} & \textbf{3.1} & \textbf{3.1} & \textbf{3.1} \\
\quad DMLab & \textbf{17.3} & 14.6 & 14.8 & 14.5 & \textbf{83.7} & 82.7 & 82.6 & 83.1 & \textbf{17.4} & 16.1 & 16.4 & 15.9 \\
\quad KITTI (distance) & \textbf{31.6} & 15.0 & 15.2 & 14.3 & -- & -- & -- & -- & \textbf{37.6} & 26.8 & 25.2 & 28.4 \\
\quad SmallNORB (azimuth) & 4.6 & \textbf{5.6} & 5.3 & 5.3 & 26.3 & 27.4 & \textbf{27.7} & 27.5 & 4.7 & \textbf{5.7} & 5.4 & 5.4 \\
\quad SmallNORB (elevation) & 11.4 & 11.3 & \textbf{11.6} & 11.2 & \textbf{55.8} & 53.9 & 54.7 & 55.2 & 11.4 & 11.2 & \textbf{11.5} & 11.1 \\
\quad MNIST & \textbf{69.6} & 59.2 & 59.9 & 63.1 & 92.5 & 90.3 & \textbf{93.0} & 91.4 & \textbf{69.4} & 59.3 & 60.4 & 63.5 \\
\quad Rendered SST2 & 60.7 & \textbf{63.0} & 60.9 & 62.5 & -- & -- & -- & -- & 60.8 & \textbf{63.0} & 60.9 & 62.5 \\
\midrule
Average & 45.8 & 47.3 & 46.6 & \textbf{47.6} & 73.8 & 74.2 & 74.5 & \textbf{75.0} & 45.4 & 45.8 & 45.8 & \textbf{46.7} \\
\bottomrule
\end{tabular}

\vspace{-8pt}
\caption{\textbf{Zero-shot classification on the full CLIP benchmark with ViT-B/16.} Top-1, top-5, and mean per-class recall of the pretrained OpenAI CLIP ViT-B/16 (Pre.) and of its fine-tuning on CC12M with the CLIP (InfoNCE), SigLIP, and \ours losses; mean over three seeds. $^\dagger$Mean average precision in place of top-1. Top-5 is omitted (--) for tasks with five or fewer classes. Averages are taken over the tasks with a value in each group (38, 33, and 37).}
\label{tab:full_benchmark_vitb16}
\end{table}

\clearpage

\section{Theoretical analysis}
\label{sec:analysis}
\label{sec:invariance}

The per-query rankings of a model fix mAP and Recall@$k$, but not \uAP, which also depends on how the similarities of different queries are interleaved in the global ranking. Our goal is to find the interleaving that gives the \emph{theoretical best} \uAP given the fixed per-query rankings, and to measure how close each loss comes to it. To make this precise, let $G$ be the group of transformations that apply a \emph{different} strictly increasing bijection of $\real$ to every row of $S$,
\begin{equation}
  (\psi\cdot S)_{i,j} = \psi_i(s_{i,j}), \qquad \psi=(\psi_1,\dots,\psi_M),\quad \psi_i:\real\to\real \text{ strictly increasing bijection},
  \label{eq:group}
\end{equation}
\ie a per-query transformation of the similarities that leaves each query's own ranking untouched. We call the set of all matrices $\psi\cdot S$ with $\psi\in G$ the \emph{orbit} of $S$. Its members have the same per-query rankings as $S$ and differ only in how the rows are interleaved.
\begin{lemma}
\label{lem:invariance}
Any function of $S$ that depends only on the ranking within each row takes the same value on every member of the orbit of $S$. mAP and Recall@$k$ are such functions, whereas \uAP is not.
\end{lemma}
A loss that reads only within-row rankings is thus constant on every orbit and cannot tell one interleaving of the rows from another, which is exactly what \uAP varies over. mSAP is such a loss (in the limit $\tau\to0$), since it reads only differences of similarities within a row. \ours, on the other hand, by flattening $S$, is the minimal change to mSAP that makes the loss depend on the interleaving.

\paragraph{Decomposing \uAP into rankings and consistency.} By the lemma, \uAP depends on the per-query rankings, which fix the orbit of $S$, and on the consistency of the similarities across queries, which fixes where on the orbit $S$ lies. The \emph{ceiling} $C(S,Y)$ is the largest \uAP on the orbit of $S$, \ie the \uAP the model would reach if its similarities were made perfectly consistent across queries with its per-query rankings unchanged. The \emph{efficiency} $\mathrm{eff}=\uAP/C$ is the share of the ceiling the model actually realizes, and measures how consistent its similarities are across queries. Since $\uAP=\mathrm{eff}\cdot C$, two models can differ in \uAP because their rankings allow different ceilings or because they realize different shares of them. Differencing the product separates the two exactly,
\begin{equation}
  \Delta\uAP \;=\; \underbrace{\overline{\mathrm{eff}}\,\Delta C}_{\text{ranking}} \;+\; \underbrace{\overline{C}\,\Delta\mathrm{eff}}_{\text{consistency}},
  \label{eq:split}
\end{equation}
where $\Delta$ is the difference between the two models and bars denote the mean of their values. The \emph{ranking} term is the part of the difference due to better per-query rankings, \ie a higher ceiling, at the average efficiency, and the \emph{consistency} term the part due to more consistent similarities across queries, \ie a higher efficiency, at the average ceiling. The proof of the lemma and the computation of $C$ follow the results.
\label{sec:supp_ceiling}

\begin{table}[b]
    \centering
    \vspace{-9pt}
    \caption{\textbf{Rankings \vs consistency across queries.} \uAP, ceiling $C$ and efficiency $\mathrm{eff}$ of the losses of \Cref{tab:sota} on two datasets. {\color{gray!65!black} rank} and {\color{gray!65!black} cons} are the terms of \cref{eq:split} relative to \ours on each row, in \uAP points.
    \label{tab:ceiling}
    \vspace{-9pt}
    }
    \scriptsize
\setlength{\tabcolsep}{2.5pt}
\begin{tabular}{l ccc >{\color{gray!65!black}}c >{\color{gray!65!black}}c ccc >{\color{gray!65!black}}c >{\color{gray!65!black}}c}
    \toprule
    \multirow{2}{*}{\textbf{method}} & \multicolumn{5}{c}{\textbf{SOP}} & \multicolumn{5}{c}{\textbf{iNaturalist}} \\ \cmidrule(lr){2-6} \cmidrule(lr){7-11}
    & \uAP & $C$ & eff & rank & cons & \uAP & $C$ & eff & rank & cons \\ \midrule
    \textbf{Contrastive} & 53.7 & 76.3 & 70.4 & +4.1 & +1.1 & 28.0 & 48.2 & 58.1 & +3.3 & +0.3 \\
    \textbf{Triplet} & 50.0 & 78.5 & 63.7 & +2.4 & +6.5 & 18.2 & 47.3 & 38.5 & +3.2 & +10.2 \\
    \textbf{ArcFace} & 50.3 & 72.8 & 69.1 & +6.5 & +2.1 & 23.0 & 48.9 & 47.0 & +2.6 & +6.0 \\
    \textbf{Proxy Anchor} & 51.9 & 75.9 & 68.4 & +4.3 & +2.7 & 21.8 & 48.9 & 44.6 & +2.5 & +7.3 \\
    \textbf{Multi-Similarity} & 57.3 & 80.3 & 71.4 & +1.2 & +0.4 & 29.9 & 51.8 & 57.7 & +1.2 & +0.5 \\
    \textbf{RS@$k$} & 56.7 & 81.7 & 69.4 & +0.2 & +2.0 & 26.6 & 53.2 & 50.0 & +0.3 & +4.7 \\
    \textbf{mSAP} & 57.0 & 82.0 & 69.5 & +0.0 & +1.9 & 27.8 & 54.1 & 51.4 & -0.2 & +4.0 \\
    \rowcolor{appleteal!60} \textbf{\ours} & 58.9 & 82.0 & 71.8 & \multicolumn{1}{c}{--} & \multicolumn{1}{c}{--} & 31.6 & 53.8 & 58.7 & \multicolumn{1}{c}{--} & \multicolumn{1}{c}{--} \\
    \bottomrule
\end{tabular}

\end{table}

\paragraph{Results.} In \Cref{tab:ceiling}, where $\Delta$ is \ours minus the loss of the row, the ceiling $C$ of \ours is within $0.6$ points of those of mSAP and RS@$k$, so the three losses learn rankings that allow the same \uAP, and what \ours adds is a higher efficiency on both datasets. Accordingly, \cref{eq:split} attributes its whole gain over mSAP to the consistency term. Among the pairwise and proxy losses, Multi-Similarity has the highest ceiling and, with Contrastive, an efficiency close to that of \ours, whereas the others have ceilings well below it. Contrastive loses mainly on rankings, ArcFace and Proxy Anchor mainly on rankings on SOP but on consistency on iNaturalist, and Triplet has by far the lowest efficiency. Optimizing the global ranking thus buys consistency at no cost in per-query ranking. Nonetheless, even \ours leaves much of the achievable consistency unrealized, a gap that per-query losses cannot act on, since they are blind to the interleaving of the queries.

\newpage
\paragraph{Proof of \Cref{lem:invariance}.}
Let $\psi\in G$ act on row $i$ by $\psi_i$. Since $\psi_i$ is strictly increasing,
\[
  \psi_i(s_{i,l})\ge \psi_i(s_{i,j}) \iff s_{i,l}\ge s_{i,j},
  \qquad\text{so}\qquad
  h\bigl(\psi_i(s_{i,l})-\psi_i(s_{i,j})\bigr)=h(s_{i,l}-s_{i,j})
\]
for all $j,l\in\Omega$, with $h$ the step function of \cref{eq:heaviside}. The precision of \cref{eq:precision} and the AP of a row are built from such within-row comparisons only, so they take the same value at $S$ and $\psi\cdot S$, and so does mAP, their average over rows. Recall@$k$ is the fraction of queries with a positive among their $k$ highest similarities,
\begin{equation}
  \mathrm{Recall@}k(S,Y)=\frac{1}{M}\sum_{i=1}^{M}h\Bigl(k-\min_{j\in\cP(Y_i)}\sum_{l\in\Omega} h(s_{i,l}-s_{i,j})\Bigr),
  \label{eq:recall}
\end{equation}
and is again invariant for the same reason. So is any function that reads $S$ only through within-row comparisons, hence every per-query ranking metric and every loss built from within-row rank estimates.

\uAP is different. It computes AP on $\vect(S)$, so its sums also contain comparisons $h(s_{i',j'}-s_{i,j})$ between similarities of different queries, $i'\neq i$. Since $\psi_i$ and $\psi_{i'}$ are chosen independently, there is no guarantee that $h\bigl(\psi_{i'}(s_{i',j'})-\psi_i(s_{i,j})\bigr)$ is equal to $h(s_{i',j'}-s_{i,j})$. Two examples show that \uAP indeed varies on an orbit.

\emph{Example 1.} Two queries $a$ and $b$ with one positive and one negative each, $a^{+}=0.8$, $a^{-}=0.3$, $b^{+}=0.6$, $b^{-}=0.2$. Keeping row $a$ and shifting row $b$ by $\psi_b(s)=s-0.5$ changes the global ranking from
\[
  a^{+},\,b^{+},\,a^{-},\,b^{-}\quad(\uAP=1)
  \qquad\text{to}\qquad
  a^{+},\,a^{-},\,b^{+},\,b^{-}\quad\bigl(\uAP=\tfrac12\bigl(1+\tfrac23\bigr)=0.833\bigr),
\]
while both per-query rankings, and with them mAP $=1$ and Recall@$1=1$, are unchanged.

\emph{Example 2.} $M$ queries with $N$ items each, every query having a single positive ranked first in its row, so mAP $=1$ on the whole orbit. If every positive is ranked above every negative, $\uAP=1$. Now stack the rows one below the other, \eg by $\psi_i(s)=s-2(i-1)$ for similarities in $[-1,1]$, so that the whole of row $1$ comes first in the global ranking, then the whole of row $2$, and so on. Row $i$ occupies global positions $(i-1)N+1$ to $iN$, and its positive, the first item of the row, sits at position $(i-1)N+1$. The positives above it are those of rows $1$ to $i-1$, so the precision at the positive of row $i$ is $i/((i-1)N+1)$, and averaging over the $M$ positives gives
\[
  \uAP=\frac{1}{M}\sum_{i=1}^{M}\frac{i}{(i-1)N+1}.
\]
The first term is $1$. For $i\ge2$, dropping the $+1$ and using $i/(i-1)\le2$ gives $i/((i-1)N+1)\le 2/N$, so the remaining $M-1$ terms sum to at most $2(M-1)/N$ and
\[
  \uAP\;\le\;\frac{1}{M}\Bigl(1+\frac{2(M-1)}{N}\Bigr),
\]
which approaches $1/M$ as $N\to\infty$. On this one orbit, \uAP thus ranges from $1$ down to about $1/M$ while mAP stays at $1$.\hfill$\square$

\paragraph{Computing $C$.}
The maps $\psi_i$ of \cref{eq:group} keep the order within each row and are otherwise arbitrary, so the global rankings of the members of the orbit of $S$ are exactly the interleavings of the $M$ per-query ranked lists that keep the order within each list, and $C(S,Y)$ is the largest \uAP over these interleavings. Let query $i$ have $K_i=\card{\cP(Y_i)}$ positives and let $K=\sum_i K_i$. As in the denominator of \cref{eq:precision}, the rank of a positive $j\in\cP(Y_i)$ within its own list is
\begin{equation}
  \rho_{i,j}=\sum_{l\in\Omega} h(s_{i,l}-s_{i,j}),
\end{equation}
and the position of a positive entry $t\in\cP(\vect(Y))$ in the global list is
\begin{equation}
  \pi_{t}=\sum_{t'=1}^{MN} h\bigl(\vect(S)_{t'}-\vect(S)_{t}\bigr).
  \label{eq:rank}
\end{equation}
The ranks $\rho$ are fixed on the orbit, whereas the positions $\pi$ change with the interleaving. Index the positives of query $i$ so that $\rho_{i,1}<\dots<\rho_{i,K_i}$, and all $K$ positives of the matrix so that $\pi_1<\dots<\pi_K$. Exactly $k$ positives are at or above position $\pi_k$, so the precision at the $k$-th positive of the global ranking can also be computed as $k/\pi_k$, and $\uAP=(1/K)\sum\nolimits_{k=1}^{K} k/\pi_k$. $C$ is the maximum of this mean over the interleavings, and we compute it in three steps. An optimal interleaving is built from blocks of each list, the blocks are ordered by a scheduling rule, and $C$ is bounded from both sides, as the rule need not be optimal for \uAP.

\emph{Blocks.} Consider the consecutive negatives immediately before a positive. If some of them come from other queries, moving them to just after the positive keeps the order within every list and raises \uAP, since that positive moves up and no other positive moves. In an optimal interleaving, the consecutive negatives immediately before a positive all come from the query of that positive, so each list enters as a chain of blocks, where block $j$ of query $i$ consists of the negatives between positives $j-1$ and $j$ followed by positive $j$. Its length is $\rho_{i,j}-\rho_{i,j-1}$, with $\rho_{i,0}=0$, and its density, positives per item, is the reciprocal of its length, since each block holds exactly one positive. The negatives after the last positive of a query go to the end of the global list, below every positive, where they cost nothing.

\emph{Ordering the blocks.} The blocks of a query must keep their order, so the task is to interleave $M$ chains of blocks, and we use the ratio rule of \citet{sidney1975decomposition} for it. Along each chain, consecutive blocks are merged into segments until the density is non-increasing along the chain, and the segments of all queries are then ordered by decreasing density, ties kept in chain order. The merging step is the pool-adjacent-violators algorithm~\citep{provost2001robust,fawcett2007pav}, and the ordering step is the probability ranking principle~\citep{robertson1977probability} applied to all pairs jointly.

\emph{Bounds.} Every feasible interleaving has \uAP at most $C$. The rule gives one, and a second is built greedily, without merging, by appending at each step the densest of the blocks that come next in their chains. We report the larger of their two \uAP values as $C$, which can only be too low, and bound how far it is from the true maximum as follows. Take the segments in the order of the rule and pretend that the positives of each segment are spread evenly over its items. Its \uAP is an upper bound on $C$, since at every depth it holds at least as many positives as any feasible interleaving and moving positives up can only raise \uAP. The true $C$ thus lies between the reported value and this upper bound, whether or not the rule is optimal, and on the full-size runs of \Cref{tab:ceiling} the two differ by at most $7\times10^{-7}$ (SOP) and $6\times10^{-6}$ (iNaturalist), far below the $0.1$ points at which $C$ is reported.

\emph{Checks.} We also compared the rule with an exhaustive search over all order-respecting interleavings on instances small enough for it. On $400$ random synthetic instances of two to four lists, the rule found the optimum in $382$ and the reported value, the better of the rule and the greedy schedule, in $393$, with mean deviations from the optimum of $4\times10^{-4}$ and $1\times10^{-4}$. On $100$ instances per dataset of three real query rows truncated to their top $40$ items, the rule found the optimum in $99$ (SOP) and $82$ (iNaturalist) and the reported value in $100$ and $86$, with a mean deviation of about $10^{-4}$. On the full-size runs the rule never fell below the greedy schedule.

\emph{Example.} Two queries $a$ and $b$ with six items each. Query $a$ has positives at ranks $1,2,6$ and $b$ at ranks $2,3$, so their APs are $0.833$ and $0.583$ and mAP is $0.708$ on the whole orbit. The blocks of $a$ have lengths $1,1,4$ and densities $1,1,\tfrac14$, already non-increasing, so none are merged. The blocks of $b$ have lengths $2,1$ and densities $\tfrac12,1$. The denser block comes second, so the two merge into one segment of density $\tfrac23$. Ordering by density gives the first two blocks of $a$, the segment of $b$, then the last block of $a$, \ie the global list ${+}{+}\,|\,{-}{+}{+}\,|\,{-}{-}{-}{+}$ followed by the trailing negatives of $b$, with positives at positions $1,2,4,5,9$ and
\[
  \uAP=\tfrac15\left(\tfrac11+\tfrac22+\tfrac34+\tfrac45+\tfrac59\right)=0.821,
\]
which an exhaustive search over all $924$ order-respecting interleavings confirms to be $C$. Putting the whole list of $a$ before that of $b$ gives positions $1,2,6,8,9$ and $\uAP=0.711$, and the list of $b$ first gives $2,3,7,8,12$ and $\uAP=0.502$. All three lie on one orbit, so mAP stays at $0.708$ while \uAP ranges from $0.50$ to $0.82$.

\paragraph{Remark.} The maps in $G$ act on the rows independently, so when the queries are themselves database items, as in \Cref{tab:ceiling}, the orbit contains asymmetric similarity matrices, which are infeasible with $\ell_2$-normalized embeddings. $C$ is therefore an upper bound on what an embedding could reach, by post-hoc score normalization~\citep{auckenthaler2000score,jegou2007contextual,pizzi2022self} or by a loss on all similarities jointly such as \ours, and it bounds every loss alike.

\end{document}